\def\year{2020}\relax
\documentclass[letterpaper]{article} 
\usepackage{aaai20}  
\usepackage{times}  
\usepackage{helvet} 
\usepackage{courier}  
\usepackage[hyphens]{url}  
\usepackage{graphicx} 
\usepackage{graphicx}  
\usepackage{booktabs}       
\usepackage{amsfonts}       
\usepackage{nicefrac}       
\usepackage{microtype}      

\usepackage{amsmath,amsthm,amssymb}
\usepackage{color}
\usepackage{subcaption}
\usepackage{pifont}
\usepackage{diagbox}
\usepackage{multirow}
\usepackage{enumitem}
\usepackage{algorithm}
\usepackage{algorithmic}

\newcommand{\beq}{\begin{equation}}
\newcommand{\eeq}{\end{equation}}
\newcommand{\beqs}{\begin{eqnarray}}
\newcommand{\eeqs}{\end{eqnarray}}
\newcommand{\barr}{\begin{array}}
	\newcommand{\earr}{\end{array}}
\newcommand{\bali}{\begin{aligned}}
	\newcommand{\eali}{\end{aligned}}

\newcommand{\Nc}[0]{\ensuremath{\mathcal{N}} }

\newcommand{\Dc}[0]{\ensuremath{\mathcal{D}} }

\newcommand{\Ebb}[0]{\ensuremath{\mathbb{E}} }

\newcommand{\Gbb}[0]{\ensuremath{\mathbb{G}} }

\newcommand{\ie}[0]{\emph{i.e., }}

\newcommand{\etc}[0]{\emph{etc. }}

\newcommand{\wrt}[0]{\emph{w.r.t. }}

\newcommand{\Imat}[0]{\ensuremath{{\bf I}} }

\newcommand{\bds}[1]{\boldsymbol{#1}}

\newcommand{\av}[0]{\ensuremath{\boldsymbol{a}} }

\newcommand{\xv}[0]{\ensuremath{\boldsymbol{x}} }

\newcommand{\zv}[0]{\ensuremath{\boldsymbol{z}} }

\newcommand{\Sigmamat}[0]{\ensuremath{\boldsymbol{\Sigma}} }

\newcommand{\alphav}[0]{\ensuremath{\boldsymbol{\alpha}} }
\newcommand{\betav}[0]{\ensuremath{\boldsymbol{\beta}} }
\newcommand{\gammav}[0]{\ensuremath{\boldsymbol{\gamma}} }

\newcommand{\etav}[0]{\ensuremath{\boldsymbol{\eta}} }
\newcommand{\thetav}[0]{\ensuremath{\boldsymbol{\theta}} }

\newcommand{\muv}[0]{\ensuremath{\boldsymbol{\mu}} }

\newcommand{\sigmav}[0]{\ensuremath{\boldsymbol{\sigma}} }

\newcommand{\phiv}[0]{\ensuremath{\boldsymbol{\phi}} }

\newcommand{\KL}[0]{\ensuremath{\mathrm{KL}} }

\newcommand{\argmax}{\operatornamewithlimits{argmax}}
\newcommand{\argmin}{\operatornamewithlimits{argmin}}

\newcommand{\cmark}{\ding{51}}
\newcommand{\xmark}{\ding{55}}

\usepackage{scalerel}
\newcommand\mathbox[1]{\mathord{\ThisStyle{%
			\fboxsep3\LMpt\relax\kern1\LMpt\fbox{$\SavedStyle#1$}\kern1\LMpt}}}

\title{Bridging Maximum Likelihood and Adversarial Learning via $\alpha$-Divergence}
\author{
	Miaoyun Zhao,\thanks{Correspondence to: Miaoyun Zhao {miaoyun9zhao@gmail.com} and Yulai Cong {yulaicong@gmail.com}.}
	Yulai Cong,\textsuperscript{*}
	Shuyang Dai, 
	Lawrence Carin
	\\ 
	Department of Electrical and Computer Engineering, Duke University\\
}

\begin{document}

\maketitle

\begin{abstract}

Maximum likelihood (ML) and adversarial learning are two popular approaches for training generative models, and from many perspectives these techniques are complementary. ML learning encourages the capture of all data modes, and it is typically characterized by stable training. However, ML learning tends to distribute probability mass diffusely over the data space, $e.g.$, yielding blurry synthetic images. Adversarial learning is well known to synthesize highly realistic natural images, despite practical challenges like mode dropping and delicate training. We propose an $\alpha$-Bridge to unify the advantages of ML and adversarial learning, enabling the smooth transfer from one to the other via the $\alpha$-divergence. We reveal that generalizations of the $\alpha$-Bridge are closely related to approaches developed recently to regularize adversarial learning,  providing insights into that prior work, and further understanding of why the $\alpha$-Bridge performs well in practice.

\end{abstract}

\section{Introduction}
\label{sec:Intro}

Given observed data samples, a well-known task concerns fitting a generative model to the unknown underlying data distribution. Two popular approaches for that task are classical maximum likelihood (ML) learning and recently-developed adversarial learning. Both of these approaches are equivalent to minimizing a corresponding divergence between the model distribution and the data distribution \cite{bishop_2006_PRML,goodfellow2014generative}.

ML learning seeks to find model parameters that maximize the log-likelihood of the model over the observed data samples, which is equivalent to minimizing the forward Kullback-Leibler (KL) divergence between the model distribution and the data distribution \cite{mclachlan2007algorithm}. 
When considering models with latent variables, variational inference (VI) \cite{jordan1999introduction,blei2006variational} is an important class of approximate ML learning, in which a variational expression constitutes a lower bound on the log-likelihood, and learning proceeds by seeking to maximize this bound. 
There has been significant recent work on utilizing neural networks within VI \cite{kingma2014auto,dai2018diagnosing}.

Because of the properties of the forward KL, ML learning tends to associate positive mass with each data sample, forming a zero-avoiding phenomenon \cite{minka2005divergence}. Accordingly, all data modes are ``covered'' by the model distribution; by contrast, adversarial learning is often characterized by a mode-dropping phenomenon \cite{srivastava2017veegan}.
Another advantage of ML learning (forward KL) is that its training procedure is typically much more stable than that of adversarial learning. The instability of adversarial learning is in part because the mode dropping may vary as a function of learning iteration. 
On the other hand, the zero-avoiding phenomenon of ML learning may loosely distribute probability mass among data modes.
An example consequence is that generative models trained with ML tend to generate blurry images \cite{goodfellow2014generative,larsen2015autoencoding}.
Adversarially-learned models, by contrast, are capable of synthesizing highly realistic natural images \cite{goodfellow2014generative,nowozin2016f,zhang2018self,gulrajani2017improved,brock2018large}.

The original generative adversarial network (GAN) minimizes the Jensen-Shannon (JS) divergence between the model distribution and that of the data \cite{goodfellow2014generative}. In \cite{nowozin2016f} it was shown that learning based on minimizing any $f$-divergence can be formulated as an adversarial learning objective (with the JS divergence as a special case).
In this paper, we focus on the reverse KL divergence as in \cite{li2019adversarial} because 
($i$) it naturally relates to ML learning (by reversing the KL); 
($ii$) \cite{GoogleCompareGAN} showed that most GANs with the same budget can reach similar performance with enough hyperparameter optimization and random restarts; ($iii$) adversarial learning with the $f$-divergence \cite{nowozin2016f} reduces to estimating a log-likelihood ratio between the true and model distributions, and the reverse-KL is as good as any other $f$-divergence choice for this purpose \cite{li2019adversarial};
and ($iv$) forward and reverse KL divergences are two ends of the $\alpha$-Bridge developed in this paper.

\begin{table}[]
	\caption{Comparing maximum likelihood and adversarial learning. \label{tab:forward_reverse_KL_pro_con}}	
	\centering
	\resizebox{0.9\hsize}{!}{
		\begin{tabular}{c|c|c}
			\hline\hline
			\multirow{2}{*}{{\diagbox{Property}{Method}}} 
			& Maximum Likelihood & Adversarial\\
			& (Forward KL) & (Reverse KL)\\
			\hline
			Mode covering (zero-avoiding) & \cmark & \xmark \\
			Stable training & \cmark & \xmark \\
			Inference  & \cmark & \xmark \\
			Realistic generated samples & \xmark & \cmark \\
			\hline\hline
		\end{tabular}
	}
\end{table}

\begin{figure}[tb]
	\centering
		\includegraphics[height=0.23 \columnwidth]{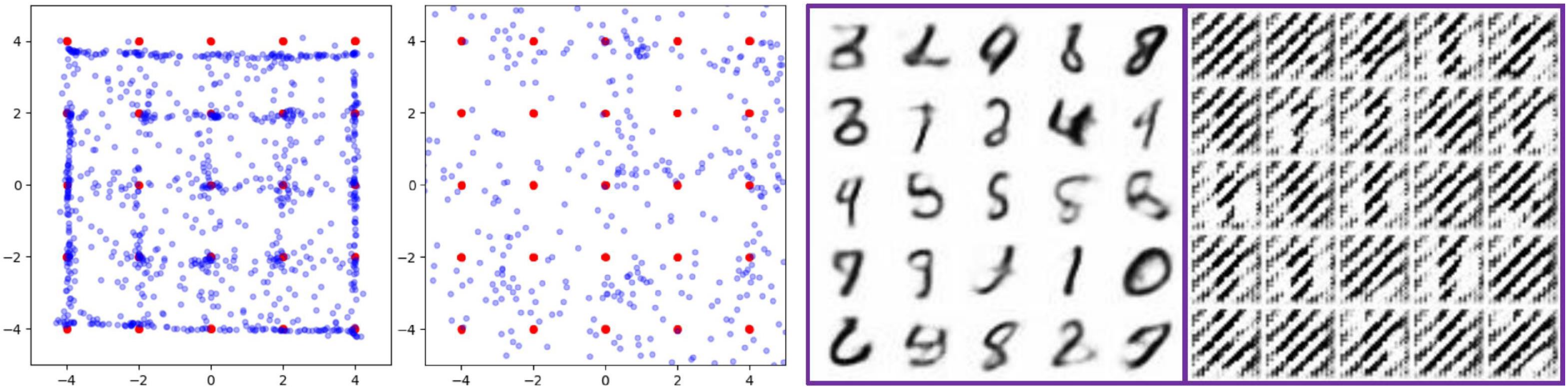}
		\caption{
			Demonstration of adversarial learning forgetting the information learned/initialized by ML learning on 25-Gaussians (the first two) and MNIST (the last two). From left to right are the snapshots of ML initialization and $20$ following iterations of adversarial learning, respectively. See 
			Appendix \ref{sec:ML_AL_explode} 
			for details.
		}
		\label{fig:explodes}
\end{figure}

It is interesting to note that ML learning (based on the forward KL) and adversarial learning (with the reverse KL as an important example) seem to have complementary advantages and disadvantages \cite{nguyen2017dual}, as shown in Table \ref{tab:forward_reverse_KL_pro_con}.
To unify their advantages, an intuitive approach would directly combine them. However, as stated in \cite{larsen2015autoencoding,mathieu2016disentangling} and empirically shown in \cite{zhang2019training}, such a naive method does not work well.
Another intuitive approach would, for example, directly initialize the reverse-KL-based adversarial learning with the parameters learned from ML learning. 
For the second approach, empirical results in Figure \ref{fig:explodes} indicate that catastrophic forgetting \cite{kirkpatrick2017overcoming,liang2018generative} happens when adversarially finetuning the ML-learned parameters.
Appendix \ref{sec:vs_FKLpRKL} 
discusses/compares other potential approaches to combine adversarial and ML learning. 
To unify the advantages from ML and adversarial learning in a principled way, we propose a novel $\alpha$-Bridge, via the $\alpha$-divergence, to smoothly connect the forward and reverse KL, through which one can transfer the advantages from one to the other. 
In addition to the practical value of the $\alpha$-Bridge, our subsequent analysis on the $\alpha$-divergence is deemed an important methodological perspective on how ML and adversarial learning are related and may be linked.

The main contributions of this paper are as follows.
($i$) An $\alpha$-Bridge is proposed to connect the forward and reverse KL in a principled manner, which can be interpreted as a novel way to ``bridge'' the two research fields of ML and adversarial learning.
($ii$) The gradient of the $\alpha$-divergence is shown to have two equivalent expressions, one that utilizes the gradient information from ML learning (forward KL), while the other uses the gradient information from adversarial learning (reverse KL). 
($iii$) The twin gradients of $\alpha$-divergence have complimentary variance properties, $\alpha$-Bridge elegantly combines the advantages of both and manages a low Monte Carlo (MC) variance along the varying of $\alpha$.
($iv$) Two generalizations of our $\alpha$-Bridge are revealed, that are closely related to CycleGAN \cite{zhu2017unpaired} and ALICE \cite{li2017alice}, two methods for regularizing (stabilizing) adversarial learning.
($v$) It is demonstrated empirically that the proposed $\alpha$-Bridge is capable of benefiting from the advantages of ML learning, 
transferring information from ML to adversarial learning, and is capable of transplanting the variational posterior in ML learning into an inference arm for adversarial learning.

\section{Preliminaries}
\label{sec:Preliminary}

Given observed data $\xv$, drawn from unknown underlying data distribution $q(\xv)$, and a parameterized model distribution $p_{\thetav}(\xv)$ with parameters $\thetav$, the task is to learn $\thetav^{*}$ so that $p_{\thetav^{*}}(\xv)$ best fits the observed data, or identically $p_{\thetav^{*}}(\xv)$ is closest to $q(\xv)$.
For that task, two popular research fields include ML learning (with ``closeness'' of $p_{\thetav^{*}}(\xv)$ and $q(\xv)$ quantified via the forward KL) and adversarial learning (with the reverse KL as an important example of how ``closeness'' is measured).

\subsection{Maximum Likelihood Learning (Forward KL)}

A classic method to match a model $p_{\thetav}(\xv)$ to the data distribution $q(\xv)$ is ML learning (or maximum likelihood estimation), namely,
\begin{equation}
\label{eq:ML_loss}
\resizebox{0.98\hsize}{!}{$
\begin{aligned}
    \thetav^{*} & = \argmax_{\thetav} \Ebb_{q(\xv)} [\log p_{\thetav}(\xv)]
    = \argmin_{\thetav} \Dc_{\KL} [q(\xv) \| p_{\thetav}(\xv)],
\end{aligned}
$}
\end{equation}
where $\Dc_{\KL} [q(\xv) \| p_{\thetav}(\xv)] \! = \! \Ebb_{q(\xv)}[\log q(\xv)\! -\! \log p_{\thetav}(\xv)]$ is the forward KL. The gradient wrt $\thetav$ is 
\begin{equation}\label{eq:ML_grad}
    {\nabla_{\thetav}} \Dc_{\KL} [q(\xv) \| p_{\thetav}(\xv)] 
    = \Ebb_{q(\xv)} [- \nabla_{\thetav} \log p_{\thetav} (x)].
\end{equation}
For more modeling capacity, it is often convenient to define $p_{\thetav}(\xv)$ as the marginal of some parameterized joint distribution $p_{\thetav}(\xv, \zv)$, with latent variable $\zv$. Although $\log p_{\thetav}(\xv)$ is usually intractable, variational inference \cite{jordan1999introduction,kingma2014auto,blei2017variational} seeks to solve the ML learning in \eqref{eq:ML_loss} via maximizing the evidence lower bound (ELBO) 
\beq\label{eq:elbo_ML}
\bali
\text{ELBO}(\thetav,\phiv) 
= \Ebb_{q(\xv) q_{\phiv}(\zv|\xv)} \big[
\log p_{\thetav}(\xv, \zv) - \log q_{\phiv}(\zv|\xv)
\big],
\eali
\eeq
where $q_{\phiv}(\zv|\xv)$ is the variational approximation with parameters $\phiv$, and the bound is tight when $q_{\phiv}(\zv|\xv) = p_{\thetav}(\zv|\xv)$.
The gradient wrt $\thetav$ becomes 
$$
\bali
    {\nabla_{\thetav}} 
    & \text{ELBO}(\thetav,\phiv) 
    = \Ebb_{q(\xv) q_{\phiv}(\zv|\xv)} [- \nabla_{\thetav} \log p_{\thetav} (\xv, \zv)].
\eali
$$
In practice, $\mathbb{E}_{q(\xv)}[\cdot]$ are approximated as averages over a finite set of observed samples.

\subsection{Adversarial Learning (Reverse KL)}

Recent progress has resulted in many techniques for adversarial training of generative models \cite{goodfellow2014generative,gulrajani2017improved,nowozin2016f,brock2018large}.
The original GAN \cite{goodfellow2014generative} seeks to solve
\begin{equation}\label{eq:GAN_lossd}
\resizebox{0.95\hsize}{!}{$
\begin{aligned}
    \min_{\thetav} \max_{\betav} \Ebb_{q(\xv)} [\log \sigma(f_{\betav}(\xv))]
    + \Ebb_{p_{\thetav} (\xv)} [\log (1-\sigma(f_{\betav}(\xv))) ],
\end{aligned}
$}
\end{equation}
where $\sigma(f_{\betav}(\xv)) \triangleq D_{\betav}(\xv)$ is called the discriminator, $\sigma(a)=1/[1+\exp(-a)]$, and samples are drawn from $p_{\thetav}(\xv)$ by the generative process 
\beq\label{eq:p_theta_x_gan}
\bali
\xv \sim \delta(\xv | G_{\thetav}(\zv)), \zv \sim p(\zv),
\eali
\eeq
where $\delta(\xv | \av)$ is the Dirac delta function located at $\av$, $G_{\thetav}(\zv)$ is called the generator, and $p(\zv)$ is an easy-to-sample distribution. It is shown \cite{goodfellow2014generative} that the optimal $\betav^{*}$ for \eqref{eq:GAN_lossd} satisfies
\begin{equation}\label{eq:f_log_ratio}
\begin{aligned}
    f_{\betav^{*}}(\xv) = \log q(\xv) - \log p_{\thetav} (\xv).
\end{aligned}
\end{equation}
Accordingly, \eqref{eq:GAN_lossd} seeks to minimize the Jensen-Shannon (JS) divergence for parameters $\thetav$ \cite{goodfellow2014generative}.

Alternatively, one could also consider a similar GAN objective based on the reverse KL divergence $\Dc_{\KL} [p_{\thetav}(\xv) \| q(\xv)] $ \cite{nowozin2016f,li2019adversarial}, on which we focus in this paper; as discussed in the Introduction, many GANs are closely related \cite{nowozin2016f} and can reach similar performance with the same budget \cite{GoogleCompareGAN}, and therefore a focus on the reverse KL is not deemed particularly limiting. The log-ratio estimate in \eqref{eq:f_log_ratio} is exploited both in the reverse KL and its gradient as
\begin{equation}\label{eq:RKL_loss_grad}
\begin{aligned}
	\bali
    & \Dc_{\KL} [p_{\thetav}(\xv) \| q(\xv)] 
    = \Ebb_{p_{\thetav}(\xv)} [-f_{\betav^{*}}(\xv)]
    \\
    & \nabla_{\thetav} \Dc_{\KL} [p_{\thetav}(\xv) \| q(\xv)] =
    \\
    & \qquad \Ebb_{p(\zv)\delta(\xv | G_{\thetav}(\zv))} \big[ -[\nabla_{\thetav} G_{\thetav}(\zv)]
    [\nabla_{\xv} f_{\betav^{*}}(\xv) ] \big].
\eali
\end{aligned}
\end{equation}

\section{Connecting Maximum Likelihood and Adversarial Learning via $\alpha$-Bridge}
\label{sec:Trans}

Maximum likelihood and adversarial learning have many {\em complementary} strengths and weaknesses, motivating development of a method that achieves their principled integration. Toward that end, we propose what we term an $\alpha$-Bridge, designed using the $\alpha$-divergence \cite{cichocki2010families}. The $\alpha$-Bridge smoothly connects the forward and reverse KL divergences, making it possible to transfer advantages from one to the other.

Given model distribution $p_{\thetav}(\xv)$ and the underlying data distribution $q(\xv)$, the $\alpha$-divergence measuring the dissimilarity between these two distributions is defined as 
\begin{equation}\label{eq:alpha-div}
    \Dc_{\alpha} [p_{\thetav}(\xv) \| q(\xv)]
    = \frac{1}{\alpha (1 - \alpha)} \Big[
    1 - \int p_{\thetav}(\xv)^{\alpha} q(\xv)^{1-\alpha} d\xv \Big].
\end{equation}
The $\alpha$-divergence has many attractive properties \cite{cichocki2010families}, for example, 
($i$) it is unique \cite{amari2009alpha};
($ii$) $\lim_{\alpha \to 0} \Dc_{\alpha} [p_{\thetav} (\xv) \| q(\xv) ] = \Dc_{\KL} [q(\xv) \| p_{\thetav} (\xv) ]$; 
($iii$) $\lim_{\alpha \to 1} \Dc_{\alpha} [p_{\thetav}(\xv) \| q(\xv) ] = \Dc_{\KL} [p_{\thetav}(\xv) \| q(\xv) ]$; and 
($iv$) the $\alpha$-divergence is a continuous function of $\alpha$. 
These properties motivate development of a smooth ``bridge'' via the $\alpha$-divergence, named the $\alpha$-Bridge, to continuously transfer between forward and reverse KL. Before discussing the proposed $\alpha$-Bridge in detail, below we first reveal its key foundation, in the context of this paper: the $\alpha$-divergence has two equivalent expressions for its gradient, which utilize the gradient information either from the forward or reverse KL.

\subsection{Twin Gradients of $\alpha$-Divergence}
\label{sec:Grad}

Given the $\alpha$-divergence defined in \eqref{eq:alpha-div}, with straightforward derivation, we have
\begin{equation}\label{eq:alpha_grad}
\begin{aligned}
    & \nabla_{\thetav} \Dc_{\alpha} [p_{\thetav}(\xv) \| q(\xv)] = 
    \\
    & 
    \quad \frac{1}{1 - \alpha} \Big[
    - \int p_{\thetav}(\xv)^{\alpha - 1} q(\xv)^{1-\alpha} \nabla_{\thetav} p_{\thetav}(\xv) d\xv \Big].
\end{aligned}
\end{equation}
An interesting fact of \eqref{eq:alpha_grad} is that one can turn it into an expectation-based expression wrt either the data distribution $q(\xv)$ or the model one $p_{\thetav}(\xv)$, resulting in two different expressions for the same gradient  
(see 
Appendix \ref{sec:derive_twin_grad} 
for details).
By forming expectations wrt $q(\xv)$, we have
\begin{equation}\label{eq:Forward_grad}
\begin{aligned}
    & \nabla_{\thetav} \Dc_{\alpha} [p_{\thetav}(\xv) \| q(\xv)] = 
    \\
    & 
    \frac{1}{1-\alpha} \Ebb_{q(\xv)} \left[
    -\Big[ \frac{p_{\thetav}(\xv)}{q(\xv)} \Big]^{\alpha}
    \nabla_{\thetav} \log p_{\thetav}(\xv)
    \right]
    \triangleq \nabla_{\thetav} \Dc_{\alpha}^F ,
\end{aligned}
\end{equation}
where $\nabla_{\thetav} \Dc_{\alpha}^F$ is used for brevity. The gradient information from the forward KL in \eqref{eq:ML_grad} serves as a building block for (\ref{eq:Forward_grad}).
However, it is {\em different} from the direct gradient of the forward KL in that $\nabla_{\thetav} \Dc_{\alpha}^F$ has an adaptive ratio-related weight term $\frac{1}{1-\alpha} \big[ \frac{p_{\thetav}(\xv)}{q(\xv)} \big]^{\alpha}$ within the expectation (when $\alpha\rightarrow 0^+$ this term vanishes, leading to the gradient of the forward KL in that limit). For the gradient expression related to $p_{\thetav}(\xv)$ modeled in \eqref{eq:p_theta_x_gan}, we have
\begin{equation}\label{eq:Reverse_grad}
\resizebox{0.95\hsize}{!}{$
\bali
    & \nabla_{\thetav} \Dc_{\alpha} [p_{\thetav}(\xv) \| q(\xv)]
    \triangleq \nabla_{\thetav} \Dc_{\alpha}^R = 
    \\
    &
    \Ebb_{p(\zv) \delta(\xv | G_{\thetav}(\zv))} \left[
    [\nabla_{\thetav} G_{\thetav}(\zv)]
    \Big[\frac{q(\xv)}{p_{\thetav}(\xv)} \Big]^{1-\alpha}
    \Big[ \nabla_{\xv} \log \frac{p_{\thetav}(\xv)}{q(\xv)} \Big]
    \right].
\eali
$}
\end{equation}
Similarly we use $\nabla_{\thetav} \Dc_{\alpha}^R$ for brevity. Compared to \eqref{eq:RKL_loss_grad}, $\nabla_{\thetav} \Dc_{\alpha}^R$ utilizes the gradient information from the reverse KL, with another adaptive weighting term $\big[\frac{q(\xv)}{p_{\thetav}(\xv)} \big]^{1-\alpha}$ (which vanishes in the limit $\alpha\rightarrow 1^{-}$, yielding the gradient of the reverse KL in that limit).
For more general model $p_{\thetav}(\xv)$ beyond \eqref{eq:p_theta_x_gan}, the GO gradient \cite{cong2019go} can be utilized to calculate $\nabla_{\thetav} \Dc_{\alpha}^R$.

It is important to note that $\nabla_{\thetav} \Dc_{\alpha}^F$ and $\nabla_{\thetav} \Dc_{\alpha}^R$ are two {\em equivalent} gradient expressions for the same objective $\Dc_{\alpha} [p_{\thetav}(\xv) \| q(\xv)]$, even though they utilize different gradient information (accordingly different MC variance properties as detailed below) from the forward and reverse KL, respectively. Thus, we call them the twin gradients of $\alpha$-divergence. In the limits on $\alpha$, the former is associated with the forward KL and the latter with the reverse KL, but for $\alpha\in (0,1)$ the twin gradients are {\em not} associated with either; this explains why the proposed $\alpha$-Bridge in Sec. \ref{sec:alpha_bridge} is different from a (possibly convex) combination of the forward and reverse KL.

\begin{figure}[tb]
	\centering	
	\subcaptionbox{$\mu=3, \sigma=1$ \label{fig:GamAlphaLargeVarMain}}{
		\includegraphics[width= 0.98\columnwidth]{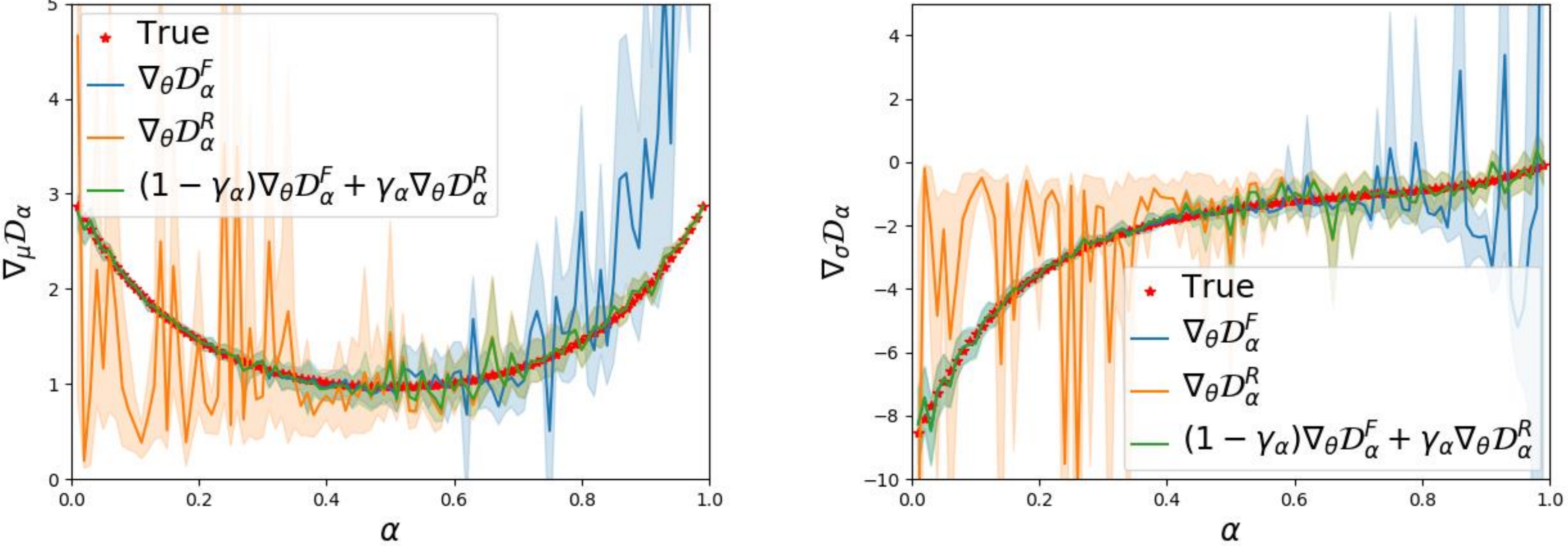}
	}
	\subcaptionbox{$\mu=1, \sigma=1$ \label{fig:GamELBOMain}}{
		\includegraphics[width= 0.98\columnwidth]{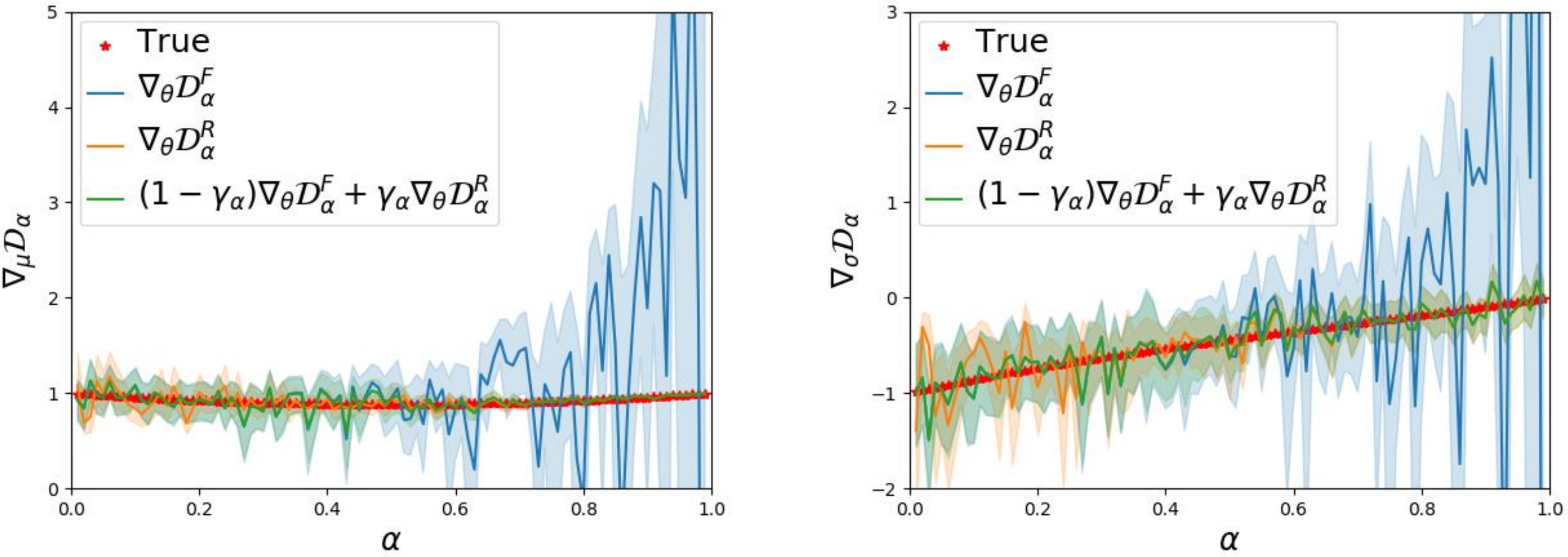}
	}
	\caption{Illustration of different MC variance properties of different gradient estimators of $\nabla_{\{\mu,\sigma\}}\Dc_{\alpha}[\Nc(x;\mu,\sigma^2)||\Nc(x;0,1)]$ for $\alpha \in (0,1)$. 
		$1$ MC sample is used to estimate the gradient. The results are based on $100$ random trials. } \label{fig:alpha_grad_var}
\end{figure}

Since $\nabla_{\thetav} \Dc_{\alpha}^F$ and $\nabla_{\thetav} \Dc_{\alpha}^R$ are equivalent expressions for $\nabla_{\thetav} \Dc_{\alpha} [p_{\thetav}(\xv) \| q(\xv)]$, any convex combination of them remains an unbiased gradient estimator, which may be interpreted as exploiting the information from one side to regularize the other side. 
We propose to use an $\alpha$-related dynamic combination as 
\begin{equation}\label{eq:combine_grad}
\begin{aligned}
\nabla_{\thetav} \Dc_{\alpha} [p_{\thetav}(\xv) \| q(\xv)]
& = (1 - \gamma_{\alpha}) \nabla_{\thetav} \Dc_{\alpha}^F
+ \gamma_{\alpha} \nabla_{\thetav} \Dc_{\alpha}^R,
\end{aligned}
\end{equation}
where $\gamma_{\alpha}$ is specified as a smooth increasing function\footnote{It is consistent with the instinct that, as smoothly transferring from the forward to reverse KL, the used information from the forward/reverse KL should smoothly decrease/increase correspondingly. 
	Appendix \ref{sec:alpha_w_alpha} 
	shows a series of experiments demonstrating several intuitive choices for $\gamma_{\alpha}$. We empirically find that the sigmoid-like function $\gamma_{\alpha} = \frac{\sigma(c\alpha+d) - \sigma(d)}{\sigma(c+d)-\sigma(d)}$ (with hyperparameters $c,d$) works well.
	Accordingly, we use such $\gamma_{\alpha}$ in our experiments and leave as future research how to optimally choose $\gamma_{\alpha}$.  
} of $\alpha$ satisfying $\gamma_{0} = 0$, $\gamma_{1}=1$, ensuring equation \eqref{eq:combine_grad} \emph{exactly} recovers the gradient of the forward$/$reverse KL when $\alpha=0 / \alpha=1$.
Such a $\gamma_{\alpha}$ is motivated by the smoothness of the $\alpha$-divergence.
When $\alpha\rightarrow0$ the $\alpha$-divergence smoothly approaches the forward KL with increasingly-similar gradients; intuitively to calculate the gradient $\nabla_{\thetav} \Dc_{\alpha} [p_{\thetav}(\xv) \| q(\xv)]$, one should prefer $\nabla_{\thetav} \Dc_{\alpha}^F$ more as it uses the ML gradient information. 
Similarly, $\nabla_{\thetav} \Dc_{\alpha}^R$ is preferred when $\alpha \rightarrow 1$ as it uses the adversarial gradient information and the $\alpha$-divergence now smoothly approaches the reverse KL.
With a simple example, Figure \ref{fig:alpha_grad_var} confirms that intuition by showing that the twin gradients $\nabla_{\thetav} \Dc_{\alpha}^F$ and $\nabla_{\thetav} \Dc_{\alpha}^R$ have complementary variance properties, the former$/$latter having lower MC variance when $\alpha \rightarrow 0 / \alpha \rightarrow 1$.
Figure \ref{fig:alpha_grad_var} also shows that combining the twin gradients as in \eqref{eq:combine_grad} unifies the advantages from both sides and presents a better gradient estimator with lower MC variance for $\alpha \in (0, 1)$. 
The twin gradients can be interpreted as control variants to each other. 
This is the foundation of our paper, which is further exploited in the following to develop our $\alpha$-Bridge. We are taking the convex combination of two different forms of the {\em same} gradient, which is {\em distinct} from just taking a convex combination of the {\em different} gradients from the forward and reverse KL. 

\begin{algorithm}[tb]
  \caption{$\alpha$-Bridge (from forward to reverse KL)}
  \label{alg:Alpha-divergence}
    \begin{algorithmic}[1]
    
    \item[\textbf{Input:} Data samples $\xv_i \sim q(\xv)$, an implicit model $p_{\thetav}(\xv)$]
    
    \item[\textbf{Output:} $\thetav^{*}$ such that $p_{\thetav^{*}}(\xv)$ is closest to the underlying ]
    \item[\qquad\qquad data distribution $q(\xv)$]

      \item[\,\,\, \textit{\textbf{\# Step I: ML learning (forward KL, $\bf \alphav=0$)}}]
      
      \STATE ML learning for the generator parameter $\thetav$ with the gradient in \eqref{eq:approx_grad_loglike}. Maximizing the ELBO in \eqref{eq:elbo_ML} for training the variational parameters $\phiv$. Pretrain the discriminator parameters $\betav$ with the objective in \eqref{eq:GAN_lossd}.
       
      \item[\,\,\, \textit{\textbf{\# Step II: Transferring from $\bf \alphav \to 0^{+}$ to $\bf \alphav \to 1^{-}$}}]
       
      \FOR{$\alpha$ gradually increasing from $0^{+}$ to $1^{-}$}

		\STATE Train $\thetav$ by minimizing the $\alpha$-divergence $\Dc_{\alpha} [p_{\thetav}(\xv) \| q(\xv)]$ with the gradient in \eqref{eq:combine_grad_practical};
		
		\STATE Train $\phiv$ by maximizing the ELBO in \eqref{eq:elbo_ML};

      	\STATE Train $\betav$ with the objective in \eqref{eq:GAN_lossd};      
      
      \ENDFOR
    
      \item[\,\,\, \textit{\textbf{\# Step III: Adversarial learning (reverse KL, $\bf \alphav=1$)}}]
    
      \STATE Refine $\thetav$, $\phiv$, and $\betav$ with the objectives in \eqref{eq:RKL_loss_grad}, \eqref{eq:elbo_ML}, and \eqref{eq:GAN_lossd}, respectively.
      
    \end{algorithmic}
\end{algorithm}

\subsection{$\alpha$-Bridge via Twin Gradients}
\label{sec:alpha_bridge}

Based on the twin gradients discussed above, we propose a novel $\alpha$-Bridge to dynamically transfer between forward KL (ML learning) and reverse KL (adversarial learning), so as to unify the advantages from both ends\footnote{See 
	Appendix \ref{sec:other_poten_gener}
	for detailed discussions on other potential generalizations of $\alpha$-Bridge.}.
In this paper, we are motivated by applications associated with GAN, with a goal of generating realistic samples from our model. Accordingly, we set our $\alpha$-Bridge to transfer from the forward KL to the reverse KL, in order to gradually transfer the advantages (see Table \ref{tab:forward_reverse_KL_pro_con}) of ML learning to adversarial learning. 
Specifically, we propose to train $p_{\thetav}(\xv)$ via the $\alpha$-Bridge with the following three successive steps. 

In Step $I$, we adopt ML learning (forward KL, $\alpha=0$) for efficient initialization thanks to its mode-covering and stable-training properties. One can skip this step if pretrained models from ML learning are available. 
From the perspective of practical implementation, one often need to approximately calculate the gradient in \eqref{eq:ML_grad}, as $p_{\thetav}(\xv)$ may be intractable for example for the implicit model in \eqref{eq:p_theta_x_gan}.
For this issue, we first add small Gaussian noise\footnote{We need not to add the noise if $p_{\thetav}(\xv)$ is modeled as \eqref{eq:tilde_p_theta_x_joint} in the first place, for example to take into consideration the widely-existing observation noise of data. See 
Appendix \ref{sec:app_semi_implicit} 
for details. } 
on top of the generative process of $p_{\thetav}(\xv)$ to form a semi-implicit surrogate model \cite{yin2018semi}
$
\tilde p_{\thetav}(\xv): \xv \sim \Nc(\xv | \xv', \sigma^2 \Imat), \xv' \sim p_{\thetav}(\xv'),
$
for which we have $\nabla_{\thetav} \log p_{\thetav}(\xv) = \lim_{\sigma^2 \to 0} \nabla_{\thetav} \log \tilde p_{\thetav}(\xv)$.
Observing that $\tilde p_{\thetav}(\xv)$ is equivalent to 
\beq\label{eq:tilde_p_theta_x_joint}
\tilde p_{\thetav}(\xv): \xv \sim \Nc(\xv | G_{\thetav}(\zv), \sigma^2 \Imat), \zv \sim p(\zv), 
\eeq
which has computable joint distribution $\tilde p_{\thetav}(\xv , \zv)$, 
we then use the ELBO technique to get 
\beq\label{eq:approx_grad_loglike}
\bali
\nabla_{\thetav} \log p_{\thetav}(\xv)
& \approx \nabla_{\thetav} \log \tilde p_{\thetav}(\xv)
\\ &
= \Ebb_{q(\xv) q_{\phiv^{*}}(\zv|\xv)} \big[
\nabla_{\thetav} \log \tilde p_{\thetav}(\xv, \zv)
\big],
\eali
\eeq
with an additional variational inference arm.

In the middle Step $I\!I$, we continue the training of $p_{\thetav}(\xv)$ by gradually changing $\alpha$ from $0^{+}$ to $1^{-}$, so as to transfer what's learned during Step $I$ to the next Step $I\!I\!I$ (reverse-KL-based adversarial learning, $\alpha=1$). 
The gradient of the $\alpha$-divergence in \eqref{eq:combine_grad} is used during training. The same techniques discussed above is adopted to calculate the $\nabla_{\thetav} \log p_{\thetav}(\xv)$ term within $\nabla_{\thetav} \Dc_{\alpha}^F$ in \eqref{eq:Forward_grad}.
To calculate the density-ratio-related terms in both $\nabla_{\thetav} \Dc_{\alpha}^F$ and $\nabla_{\thetav} \Dc_{\alpha}^R$ (see \eqref{eq:Forward_grad} and \eqref{eq:Reverse_grad}), we follow the common practice in the GAN literature to solve \eqref{eq:GAN_lossd} for $f_{\betav^{*}}(\xv)$ in \eqref{eq:f_log_ratio}. 
Accordingly, we have $\frac{p_{\thetav}(\xv)}{q(\xv)} = e^{-f_{\betav^{*}}(\xv)}$, $\nabla_{\xv} \log \frac{p_{\thetav}(\xv)}{q(\xv)} = -\nabla_{\xv} f_{\betav^{*}}(\xv)$, and 
\begin{equation}\label{eq:combine_grad_practical}
\resizebox{1\hsize}{!}{$
	\begin{aligned}
	& \nabla_{\thetav} \Dc_{\alpha} [p_{\thetav}(\xv) \| q(\xv)] \approx
	\\&  \frac{1 - \gamma_{\alpha}}{1 - \alpha} \Ebb_{q(\xv) q_{\phiv^{*}}(\zv|\xv)} \big[ -
	e^{- \alpha f_{\betav^{*}}(\xv)}
	\nabla_{\thetav} \log \tilde p_{\thetav}(\xv, \zv)
	\big] + 
	\\
	& \gamma_{\alpha} \Ebb_{p(\zv) \delta(\xv | G_{\thetav}(\zv))} \Big[ - 
	[\nabla_{\thetav} G_{\thetav}(\zv)]
	e^{(1 - \alpha) f_{\betav^{*}}(\xv)}
	\big[ \nabla_{\xv} f_{\betav^{*}}(\xv) \big]
	\Big],
	\end{aligned}
	$}
\end{equation}
which combines the gradient information from ML and adversarial learning with automatic weights related to both $\alpha$ and the GAN discriminator.

Finally in Step $I\!I\!I$, we use the zero-forcing reverse-KL-based adversarial learning ($\alpha=1$) to continually refine the generator parameters $\thetav$ and the discriminator parameters $\phiv$ using \eqref{eq:RKL_loss_grad} and \eqref{eq:GAN_lossd}, respectively. The corresponding training process is summarized in Algorithm \ref{alg:Alpha-divergence}.

\subsection{Connections to Prior GAN-Learning Regularization} \label{sec:insight}

Considering the aforementioned twin gradients and the $\alpha$-Bridge, we next present an interpretation of the gradient in \eqref{eq:combine_grad_practical}, with which we reveal two generalizations that are highly related to CycleGAN \cite{zhu2017unpaired} and ALICE \cite{li2017alice}. Details are given in 
Appendix \ref{sec:Two_connections_cycle}.
With $\mathbox{\xv}$ denoting the stop-gradient operator\footnote{{tf.stop\_gradient}$/${torch.no\_grad} in TensorFlow/PyTorch.}, the gradient in \eqref{eq:combine_grad_practical} can be reformulated as 
\begin{equation}\label{eq:cycle_reverseKL}
\resizebox{1\hsize}{!}{$
\begin{aligned}
    & \nabla_{\thetav} \Dc_{\alpha} [p_{\thetav}(\xv) \| q(\xv)] \approx
    \\
    & 
    \nabla_{\thetav} \left[
    \bali
        & \frac{1 - \gamma_{\alpha}}{1 - \alpha}
        \Ebb_{q(\xv) q_{\phiv^{*}}(\zv|\xv)} \Big[ 
        \frac{e^{- \alpha f_{\betav^{*}}(\xv)}}{ 2\sigma^2}
        \| \xv - G_{\thetav}(\zv) \|_2^2
        \Big] 
        \\
        & + \gamma_{\alpha} \Ebb_{p_{\thetav}(\xv)} \big[
        -e^{(1 - \alpha) f_{\betav^{*}}(\mathbox{\xv})}
        f_{\betav^{*}}(\xv) 
        \big]
    \eali
    \right],
\end{aligned}
$}
\end{equation}
where the first term can be interpreted as weighted half cycle-consistency \cite{li2017alice,zhu2017unpaired,kim2017learning}, 
and the second one is related to the reverse-KL-based adversarial learning.
Based on the interpretation in \eqref{eq:cycle_reverseKL}, one can readily verify (see 
Appendix \ref{sec:Two_connections_cycle}) 
that by generalizing the $\alpha$-Bridge derivations as in \eqref{eq:cycle_reverseKL} to consider both marginals
$$
\resizebox{0.95\hsize}{!}{$
\bali
& \Dc_{\alpha} [p_{\thetav}(\xv) \| q(\xv)] + \Dc_{\alpha} [q_{\phiv}(\zv) \| p(\zv)]
\\
& 
\overset{\nabla}{\approx} 
\left[
\bali
& \frac{1 - \gamma_{\alpha}}{1 - \alpha} \Ebb_{q(\xv) q_{\mathbox{\phiv}}(\zv|\xv)} \Big[ 
\frac{e^{- \alpha f_{\betav^{*}}(\xv)}}{2\sigma^2}
\| \xv - G_{\thetav}(\zv) \|_2^2
\Big] 
\\
& + \frac{1 - \gamma_{\alpha}}{1 - \alpha} \Ebb_{p(\zv) p_{\mathbox{\thetav}}(\xv|\zv)} \Big[ 
\frac{e^{- \alpha g_{\gammav^{*}}(\zv)}}{2\sigma^2}
\| \zv - E_{\phiv}(\xv) \|_2^2
\Big] 
\\
& + \gamma_{\alpha} \Ebb_{p_{\thetav}(\xv)} \Big[
-e^{(1 - \alpha) f_{\betav^{*}}(\mathbox{\xv})}
f_{\betav^{*}}(\xv)
\Big]
\\
& + \gamma_{\alpha} \Ebb_{q_{\phiv}(\zv)} \Big[
-e^{(1 - \alpha) g_{\gamma^{*}}(\mathbox{\zv})}
g_{\gammav^{*}}(\zv)
\Big]
\eali
\right],
\eali
$}
$$
where $\overset{\nabla}{\approx}$ means both sides have approximately equal gradients mimicking \eqref{eq:cycle_reverseKL}, and $g_{\gammav^{*}}(\zv)= \log p(\zv) - \log q_{\phiv} (\zv)$ corresponds to the optimal discriminator in the $\zv$ space.
Similarly, by considering both joint distributions
$$
\resizebox{0.95\hsize}{!}{$
\bali
& \Dc_{\alpha} [p_{\thetav}(\xv, \zv) \| q_{\mathbox{\phiv}}(\xv,\zv)]] + \Dc_{\alpha} [q_{\phiv}(\xv,\zv) \| p_{\mathbox{\thetav}}(\xv, \zv)]]
\\
& 
\overset{\nabla}{\approx} 
\left[
\bali
& \frac{1 - \gamma_{\alpha}}{1 - \alpha} \Ebb_{q_{\mathbox{\phiv}}(\xv,\zv)} \big[ -
e^{- \alpha h_{\etav^{*}}(\xv, \zv)}
\log p_{\thetav}(\xv| \zv)
\big] 
\\
& + \frac{1 - \gamma_{\alpha}}{1 - \alpha} \Ebb_{p_{\mathbox{\thetav}}(\xv, \zv)} \Big[ 
-e^{ \alpha h_{\etav^{*}}(\xv, \zv)}
\log q_{\phiv}(\zv | \xv) \Big]
\\
& + \gamma_{\alpha} \Ebb_{p_{\thetav}(\xv, \zv)} \big[-
e^{(1 - \alpha) h_{\etav^{*}}(\mathbox{\xv}, \zv)}
h_{\etav^{*}}(\xv, \zv) 
\big]
\\
& + \gamma_{\alpha} \Ebb_{q_{\phiv}(\xv,\zv)} \big[
e^{-(1 - \alpha) h_{\etav^{*}}(\xv, \mathbox{\zv})}
h_{\etav^{*}}(\xv, \zv)
\big]
\eali
\right],
\eali
$}
$$
one can generalize the $\alpha$-Bridge to a model very much resembling ALICE \cite{li2017alice}. We believe that the connections revealed above, and the techniques developed earlier, may be helpful for constituting a foundation that unifies ML learning, adversarial learning, and intuitive (regularization) properties like cycle-consistency.

\section{Related Work}

Motivated by the complementary properties of ML and adversarial learning, many methods have been considered for combining these two popular research fields, to unify their advantages. 
A direct combination of the VAE with GAN objectives was considered in \cite{larsen2015autoencoding}, only to ``observe the devil in the details'' during model development and training. Accordingly gradients were heuristically controlled in back-propagation.
It is also stated in \cite{mathieu2016disentangling} that naively combining 
those two objectives unstabilizes the system and does not lead to perceptually better generation, which is consistent with the empirical results from \cite{zhang2019training}.
The principle combination of ML and adversarial learning deserves a thorough exploration.
Instead of directly combining their objectives, the $\alpha$-Bridge dynamically transfers (information) between both sides to bypass the unstable problem.
Many other works combining ML and adversarial learning were motivated differently. On the one hand, with the target of ML learning unchanged, \cite{makhzani2015adversarial,mescheder2017adversarial} exploited GAN techniques to better handle the KL term between the prior and posterior of the latent variables, within the ELBO. On the other hand, keeping the target of adversarial learning, a variational auto-encoder/autoencoder was used as a building block within GAN discriminators, mainly for stabilizing training
\cite{berthelot2017began,ulyanov2018takes}.
A symmetric KL divergence was exploited to build objectives \cite{pu2017adversarial,chen2018symmetric}. Since those methods employed discriminators to estimate/replace the ratios within both the forward and reverse KL, the likelihood (gradient) information from forward KL was ignored.
By comparison, the $\alpha$-Bridge has the advantage of benefiting from the gradient information from ML learning.
Although combining ML and adversarial learning is enticing, no previous work has achieved this in a principled manner. 
The proposed $\alpha$-Bridge seeks to fill this gap.

\begin{figure}[tb]
	\centering
	\includegraphics[width=0.85\columnwidth]{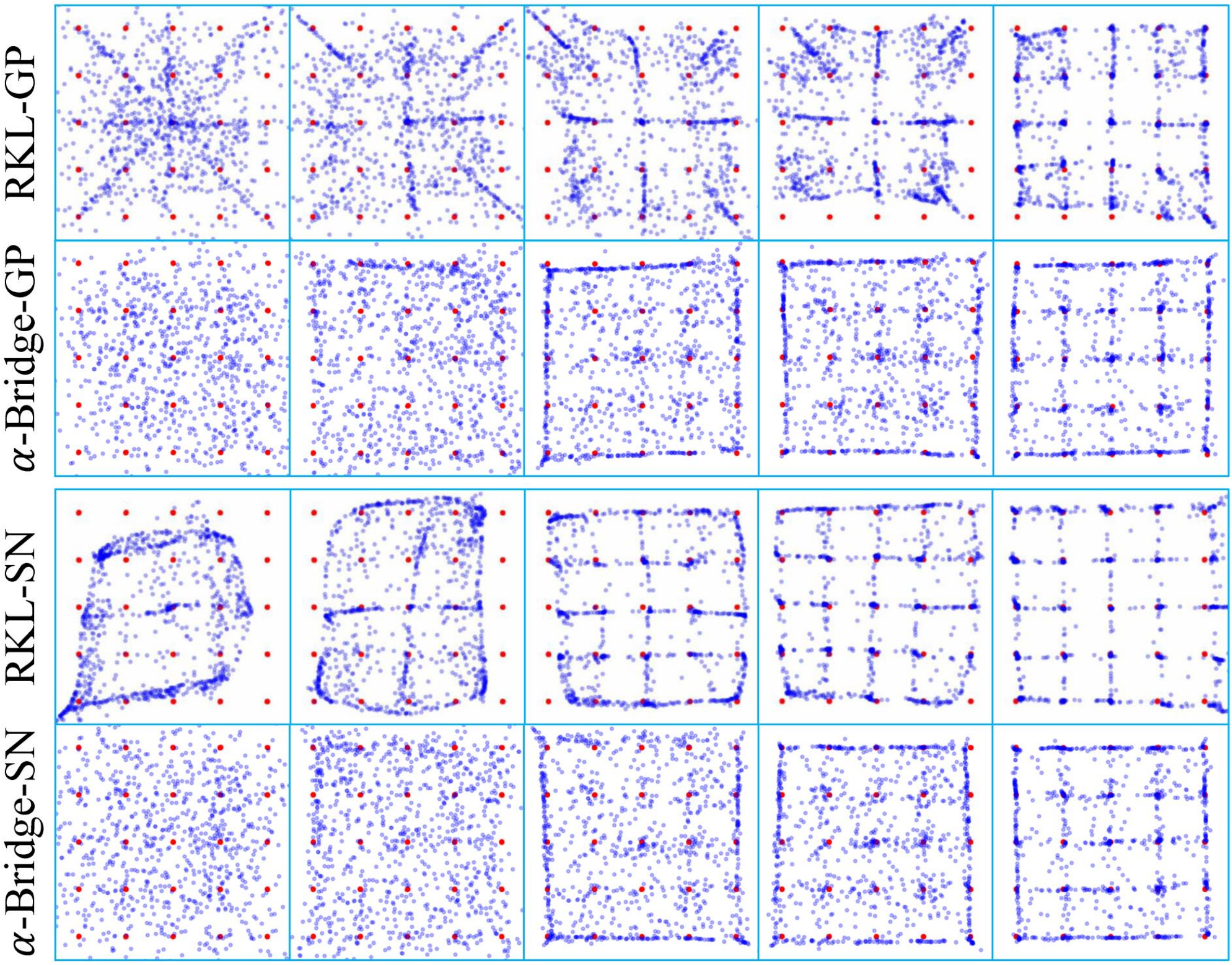}
	\caption{Demonstration on the dynamic evolution of the generated samples (blue) from the compared methods during training. Columns correspond to 1K, 2K, 4K, 6K, and 10K iterations. Real data samples are shown in red.
	}
	\label{fig:Sample_25G}
\end{figure}

\section{Experiments}
\label{sec:Experim}

\begin{figure}[tb]
	\centering
	\subcaptionbox{$\beta_1=0.1$ \label{fig:25Gs_4IS_0_1}}[.48\linewidth]{
		\includegraphics[height=0.33 \columnwidth]{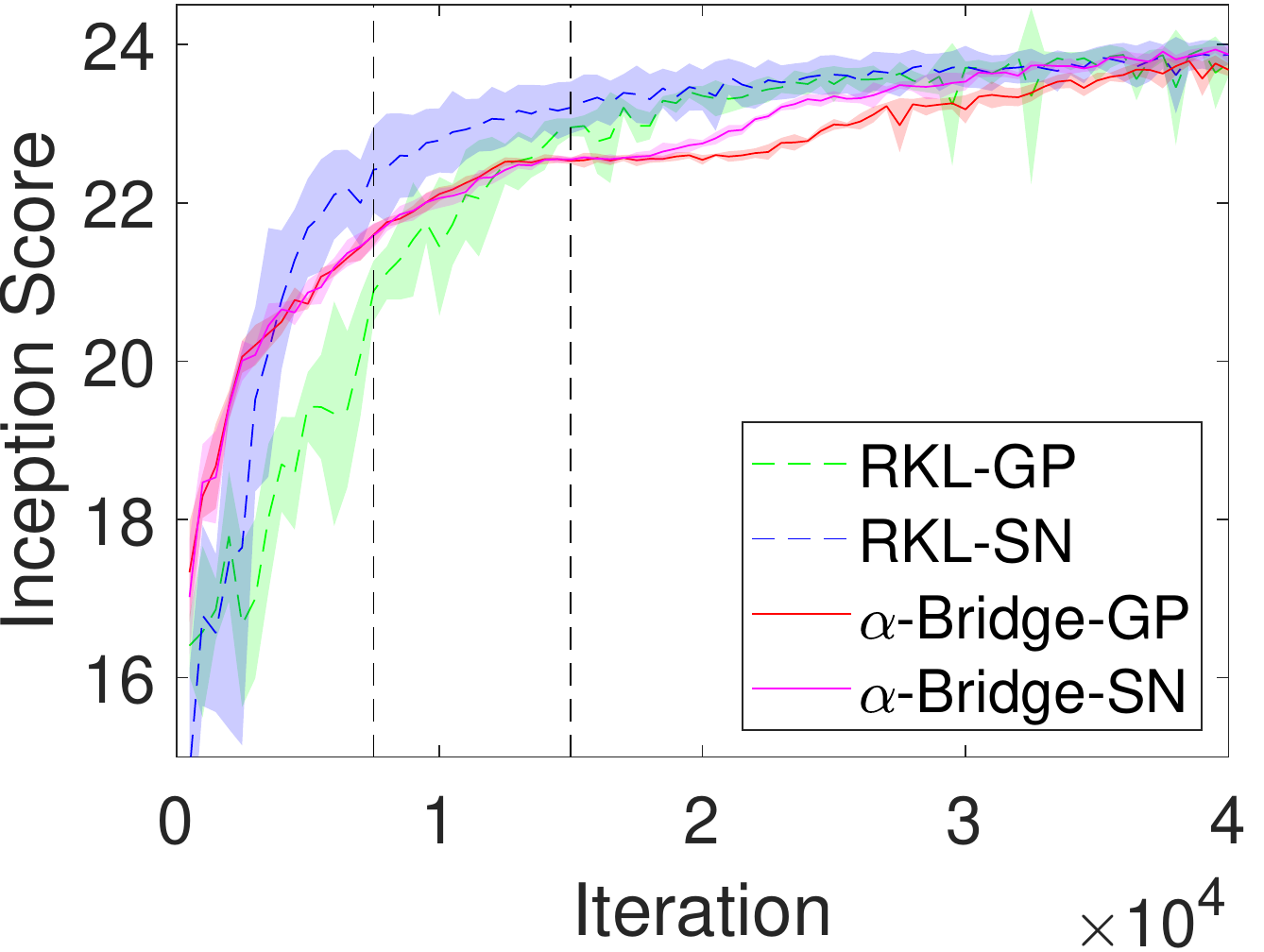}
	}
	\subcaptionbox{$\beta_1=0.1$ \label{fig:25Gs_4LL_0_1}}[.48\linewidth]{
		\includegraphics[height=0.33 \columnwidth]{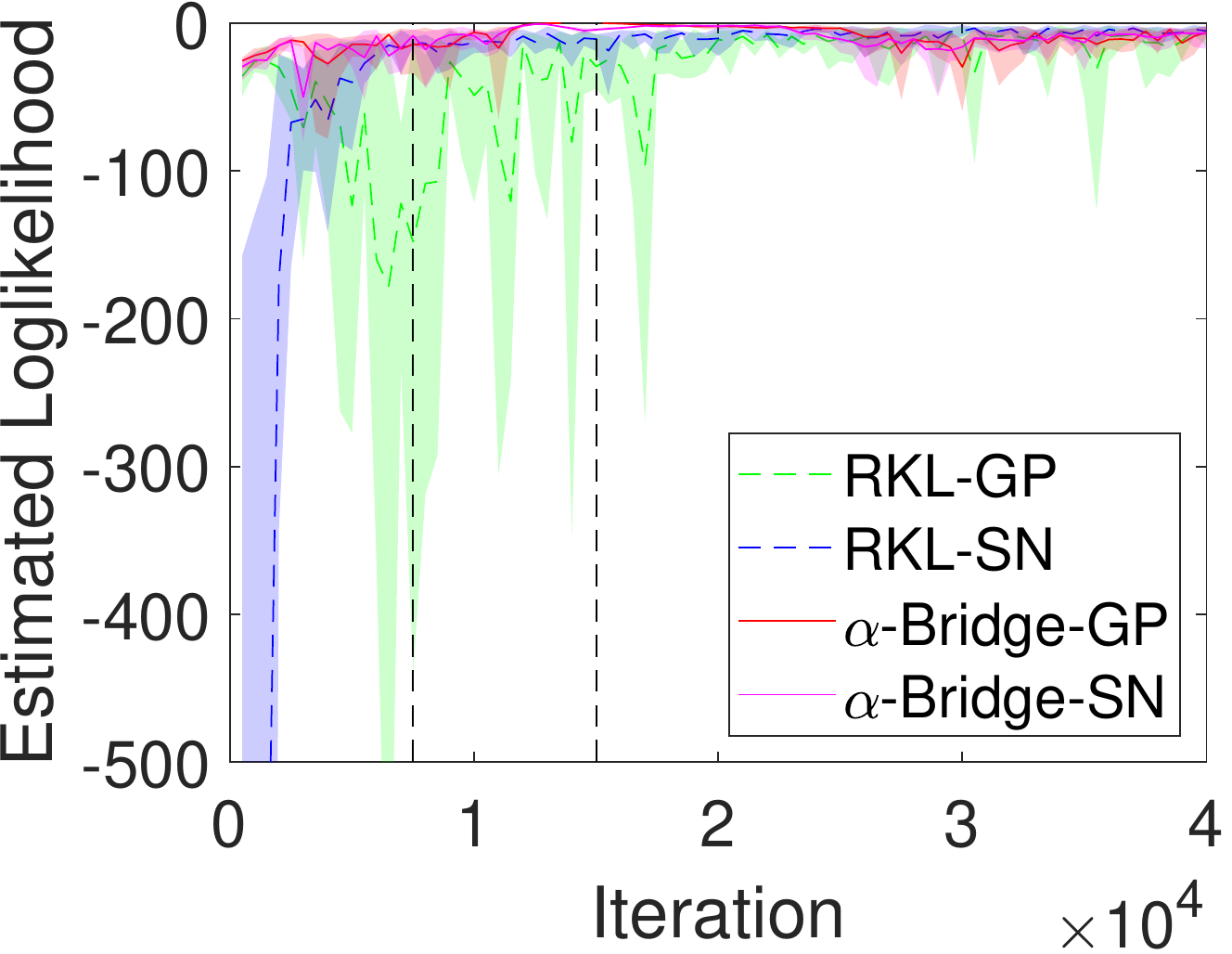}
	}
	\subcaptionbox{$\beta_1=0.5$ \label{fig:25Gs_4IS_0_5}}[.48\linewidth]{
		\includegraphics[height=0.33 \columnwidth]{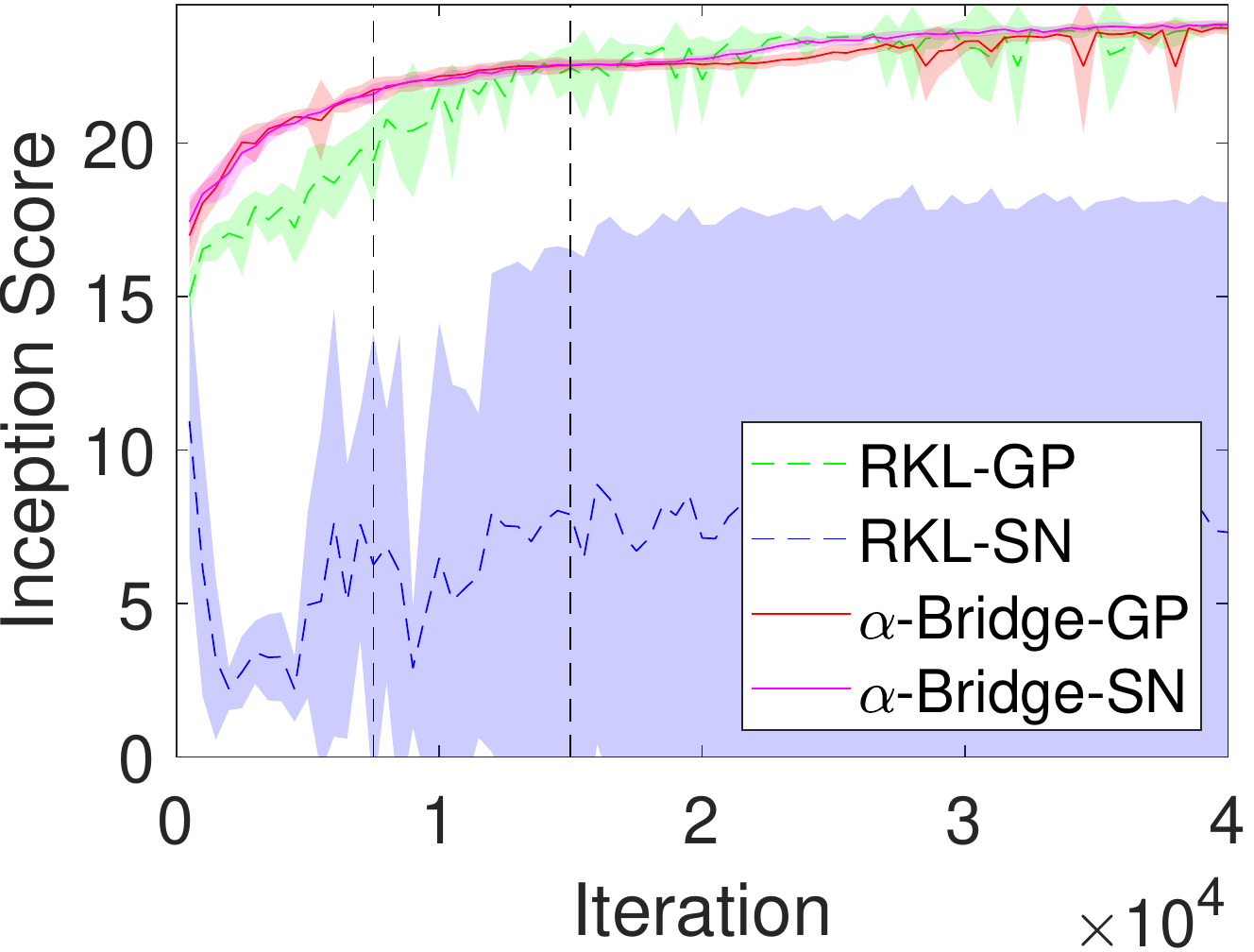}
	}
	\subcaptionbox{$\beta_1=0.5$ \label{fig:25Gs_4LL_0_5}}[.48\linewidth]{
		\includegraphics[height=0.33 \columnwidth]{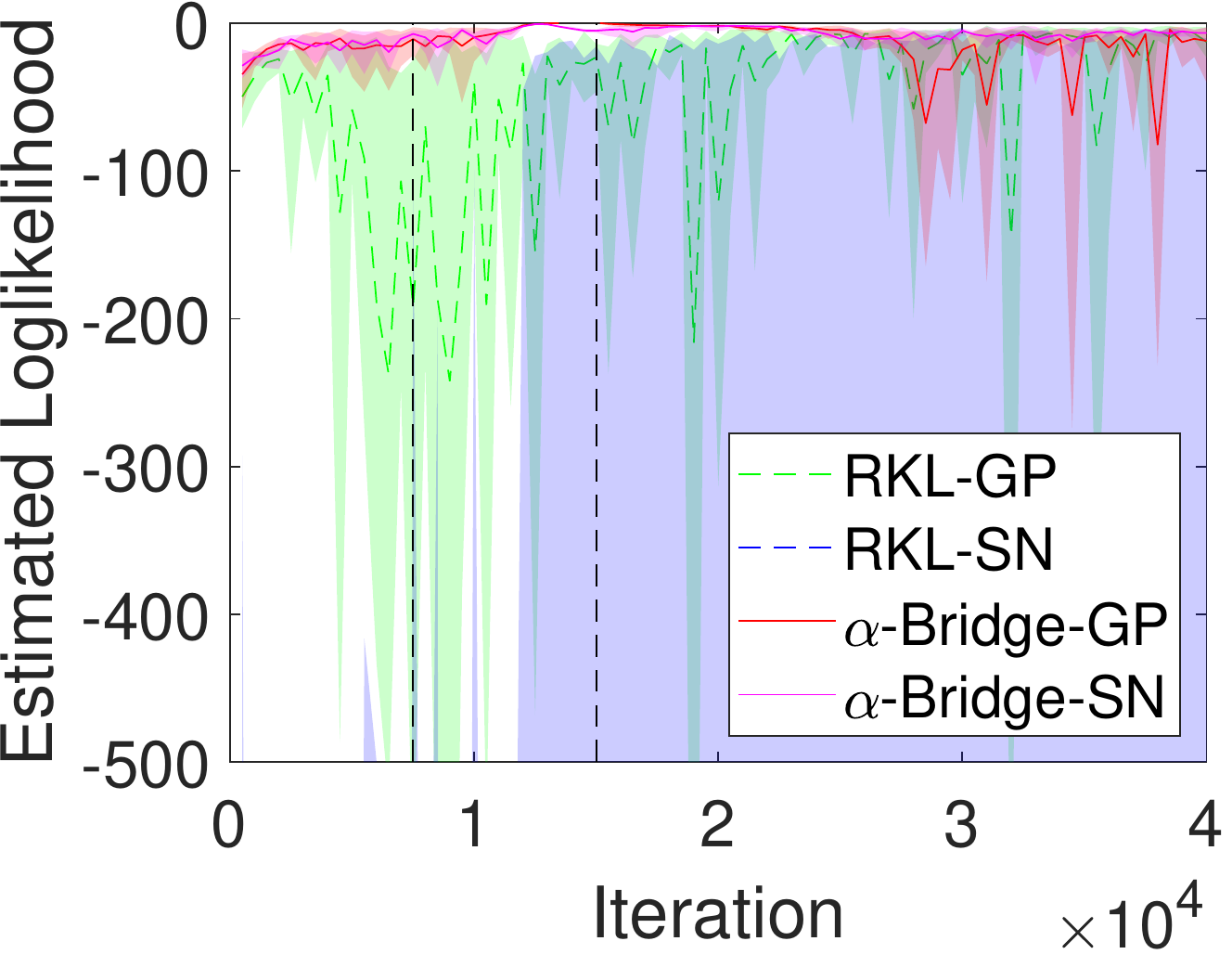}
	}
	\caption{Quantitative performance of the methods along the training process.
		Inception score (a)$/$(c) and estimated log-likelihood (b)$/$(d) when Adam \cite{kingma2014adam} hyperparameter $\beta_1=0.1$ $/$ $\beta_1=0.5$.
		Higher is better for both metrics.
		The curves are calculated over $10$ random trials. 
		Two vertical dashed lines are used to indicate the three steps of the $\alpha$-Bridge.
	}
	\label{fig:25Gs_IS_LL}
\end{figure}

We demonstrate the proposed $\alpha$-Bridge from three perspectives. 
First we show that the $\alpha$-Bridge, dynamically transferring advantages from ML to adversarial learning, exhibits a more stable training with improved robustness to hyperparameters (this is expected because of the aforementioned discussions of control variants interpretation and connections to prior GAN regularization methods).
We then show that the $\alpha$-Bridge is capable of smoothly transferring the information learned during ML learning to adversarial learning, circumventing the forgetting issue shown in Figure \ref{fig:explodes}.
Finally we highlight the versatility of the $\alpha$-Bridge, by showing its capability in transplanting the variational approximation within ML learning into an inference arm for adversarial learning. See 
Appendix \ref{sec:app_exp} 
for the detailed experimental settings.

\subsection{Stability and Robustness}
\label{sec:exp_25Gaussian}

The $25$-Gaussians example from \cite{tao2018chi} is adopted, where the data are generated from a 2D Gaussian mixture model with $25$ components, as shown in Figure \ref{fig:Sample_25G}. For direct comparison, reverse-KL-based GANs are chosen as baselines, with recent techniques to stabilize their training, \ie gradient penalty (GP) \cite{mescheder2018training} and spectral normalization (SN) \cite{miyato2018spectral}. Note it is shown in \cite{GoogleCompareGAN} that most GANs with the same budgets can reach similar performance with enough hyperparameter optimization and random restarts. Thus reverse-KL-based GANs further stabilized by GP/SN are considered as fairly good baselines (named as RKL-GP and RKL-SN, respectively)\footnote{Empirically, RKL-GP and RKL-SN show comparable/better results than WGAN-GP \cite{gulrajani2017improved} on $25$-Gaussians.}.
The inception score (IS) \cite{salimans2016improved} and the log-likelihood estimated with kernel density estimation \cite{parzen1962estimation} are used to measure the plausibility of generated samples and the data-mode-covering level of learned models, respectively.
Baseline methods are carefully tuned with their best settings adopted for fair comparison (see 
Appendix \ref{sec:app_tuningbase_25Gs}).

Figure \ref{fig:25Gs_IS_LL} shows the results of the considered methods. 
It is clear that $\alpha$-Bridge, thanks to its smooth transferring nature, is capable of benefiting from the advantages of ML learning, resulting in more stable training (see Figures \ref{fig:25Gs_4IS_0_1} and \ref{fig:25Gs_4IS_0_5}) while keeping most data modes covered (see Figures \ref{fig:25Gs_4LL_0_1} and \ref{fig:25Gs_4LL_0_5}). 
Comparing Figures \ref{fig:25Gs_4IS_0_1}-\ref{fig:25Gs_4LL_0_1} to Figures \ref{fig:25Gs_4IS_0_5}-\ref{fig:25Gs_4LL_0_5} shows $\alpha$-Bridge is relatively more robust to hyperparameters than baseline methods (see 
Appendix \ref{sec:app_tuningbase_25Gs} 
for more details). 
Figure \ref{fig:Sample_25G} shows one training curve of the compared methods, highlighting $\alpha$-Bridge's ability to benefit from the advantages of ML learning.
To address the concern of how $\alpha$-Bridge performs on real datasets, we conduct another experiment on CIFAR$10$ \cite{krizhevsky2009learning}
and observe an improved performance of (IS, FID\cite{heusel2017gans})=$(7.225, 28.083)$ over $(6.558, 33.707)$ of the vanilla DCGAN baseline (see 
Appendix \ref{sec:app_exp_cifar10}
for more results).

\subsection{Smooth Transfer of Information from ML to Adversarial Learning}

\begin{figure}[tb]
	\centering
		\includegraphics[width=\columnwidth]{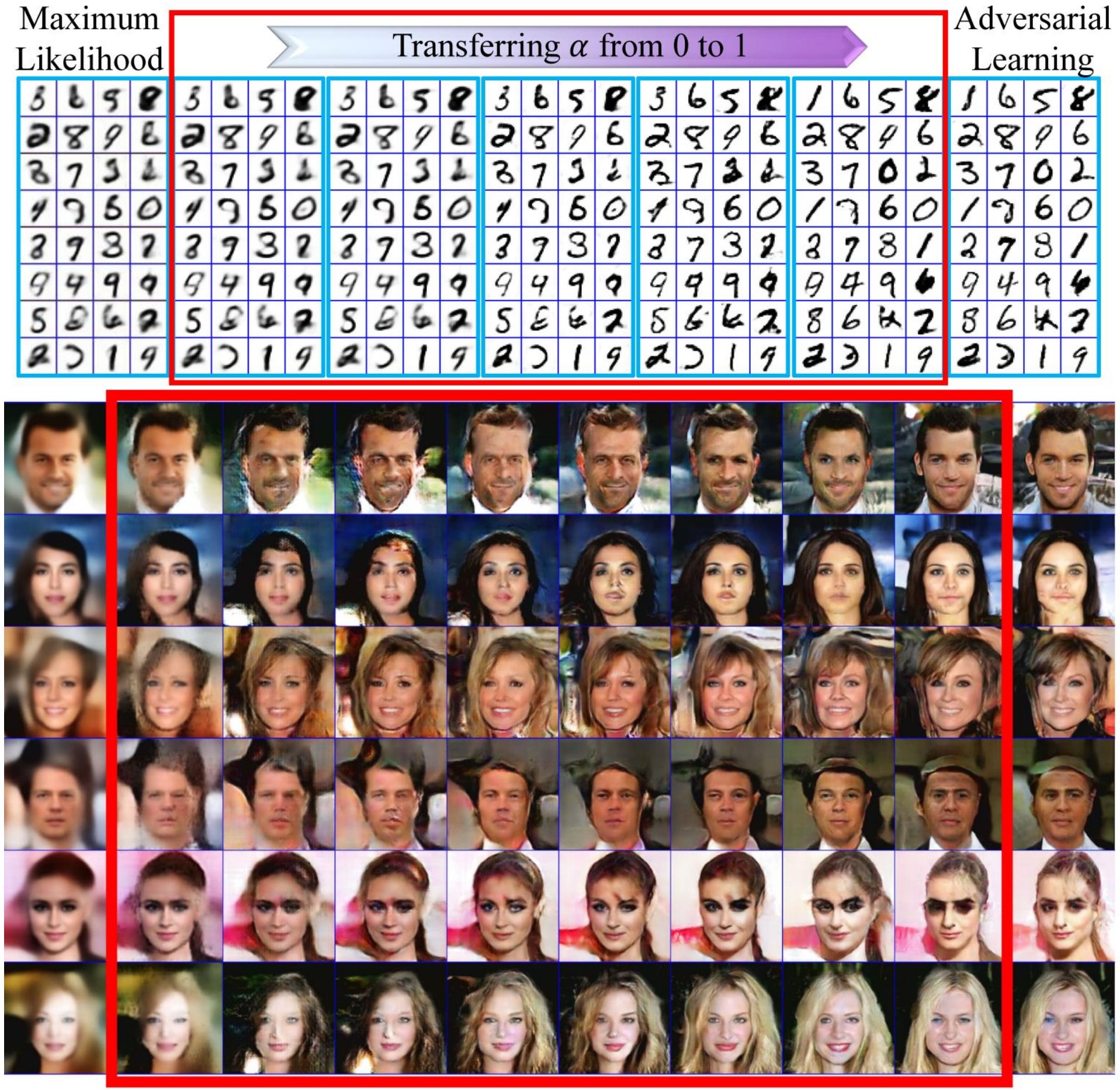}
		\caption{Random samples generated along the training process of the $\alpha$-bridge on MNIST (top) and CelebA (bottom). 
			Note most information is transferred from ML to adversarial learning, such as the class, rotation, and style of generated digits, and the basic tone, gender, expression, pose of the head, hair style of generated faces.
		}
		\label{fig:Sample_MNIST_CelebA}
\end{figure}

Besides inheriting the advantages of ML learning, another advantage of the $\alpha$-Bridge is a smooth transfer of the information learned during ML to adversarial learning.
For an explicit demonstration, we run $\alpha$-Bridge on the MNIST \cite{lecun1998gradient} and CelebA \cite{liu2015deep} datasets, and present the generated samples along the training process, as shown in Figure \ref{fig:Sample_MNIST_CelebA}. 
ML learning, \ie Step $I$ of Algorithm \ref{alg:Alpha-divergence}, provides fairly good initialization on both datasets; thanks to the zero-avoiding nature of ML learning, one might anticipate an initialization covering all data modes, similar to the phenomena observed in Figure \ref{fig:Sample_25G}.
When it comes to the transferring Step $I\!I$, $\alpha$-Bridge smoothly inherits what's learned in ML learning at the beginning, and then gradually adds more detailed information (such as sharper edges on MNIST digits and clearer background on CelebA faces) to generate increasingly realistic images.
After the transferring Step $I\!I$, one observes generated images exhibiting the features from adversarial learning, whose image quality is further refined by the adversarial Step $I\!I\!I$ of Algorithm \ref{alg:Alpha-divergence}. 
By reviewing the whole process shown in Figure \ref{fig:Sample_MNIST_CelebA}, one observes that $\alpha$-Bridge is capable of smoothly transferring most information from ML to adversarial learning, in contrast to the forgetting shown in Figure \ref{fig:explodes}.

\subsection{Transplanting ML Variational Posterior into Inference Arm for Adversarial Learning}

\begin{figure}[tb]
	\centering
		\includegraphics[width= \columnwidth]{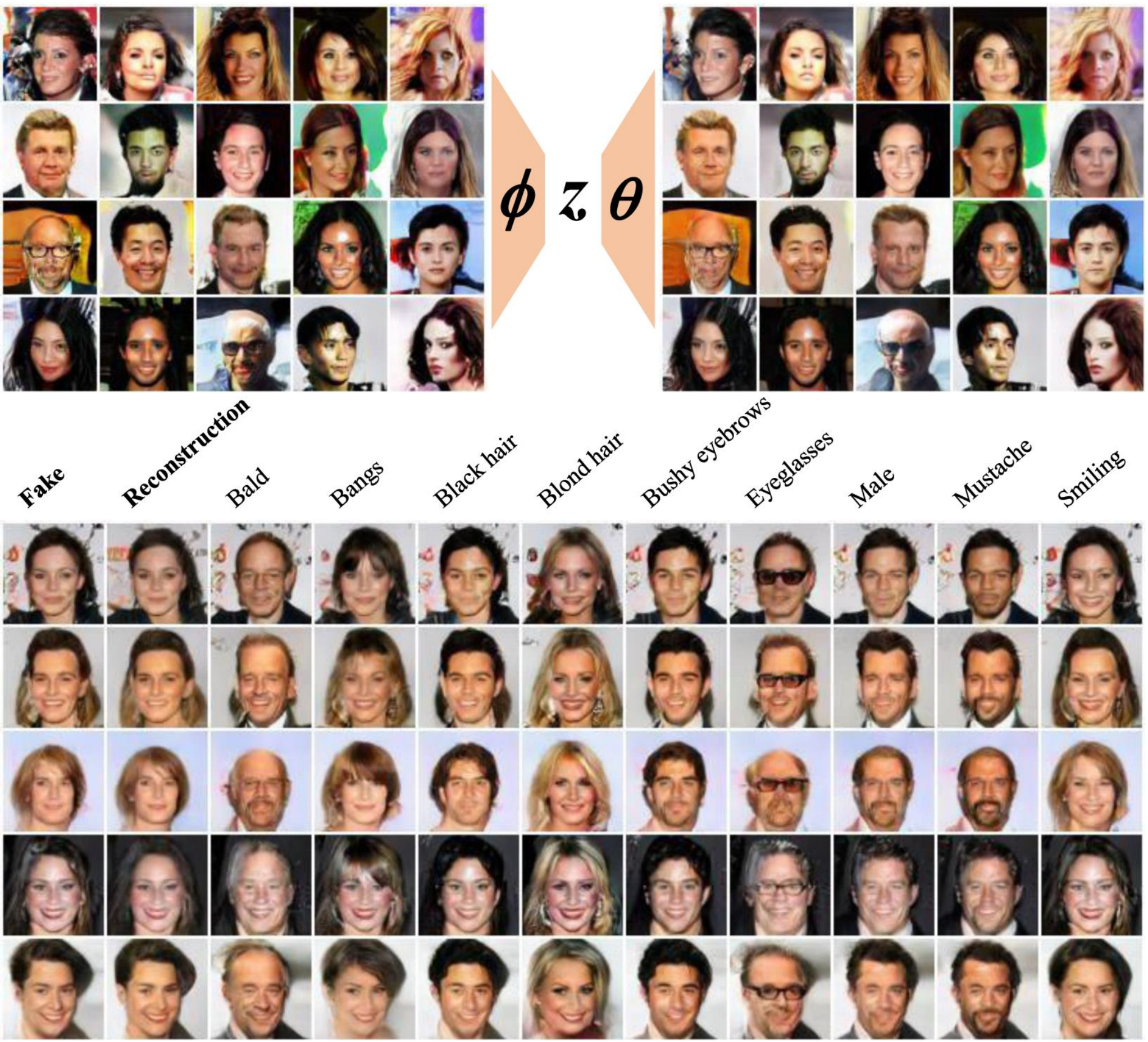}
		\caption{Using the inference arm transplanted by $\alpha$-Bridge to reconstruct (top) and manipulate (bottom) GAN generated images.
			$\phiv$ and $\thetav$ denote the inference arm $q_{\phiv}(\zv|\xv)$ and the generator $G_{\thetav}(\zv)$, respectively. 
		}
		\label{fig:attribute_inference_arm}
\end{figure}

In addition to inheriting the advantages and information from ML learning, we find that the smooth dynamical training of Algorithm \ref{alg:Alpha-divergence} also enables $\alpha$-Bridge to transplant the variational approximation within ML learning into an inference arm for adversarial learning. Such a capacity is  appealing because it enables exploiting the generative power of GANs for various practical applications. 
See 
Appendix \ref{sec:app_TransInferArm} 
for technical details.

To verify the effectiveness of the transplanted inference arm, Figure \ref{fig:attribute_inference_arm} (top) shows the encoder-decoder reconstruction for the generated fake images. It is apparent that the reconstructions are fairly good, confirming the effectiveness of the inference arm.
One can also exploit that arm for manipulation of GAN generated images, as shown in Figure \ref{fig:attribute_inference_arm} (bottom). Detailed implementations for reconstruction and manipulation are given in 
Appendix \ref{sec:app_TransInferArm}. 
It is clear that with this inference arm, one can modify the semantic concepts of the generated images like bangs, hair, gender, \etc Such capacity is valuable for transferring the generative power of GANs to various down-steaming tasks.

\section{Conclusions}

Motivated by the fact that maximum likelihood (ML) and adversarial learning have complementary characteristics, we have proposed a novel $\alpha$-Bridge, constituted via the $\alpha$-divergence, to unify their advantages in a principled manner. 
Our $\alpha$-Bridge has as its foundation newly recognized twin gradients of the $\alpha$-divergence, one of which utilizes the gradient information from the ML (forward KL) perspective, and the other from the adversarial learning (reverse KL) perspective.
We also have revealed two generalizations of $\alpha$-Bridge
that closely resemble CycleGAN \cite{zhu2017unpaired} and ALICE \cite{li2017alice}.

\subsubsection{Acknowledgments.} The research was supported by part by DARPA, DOE, NIH, NSF and ONR. The Titan Xp GPU used was donated by the NVIDIA Corporation.

\fontsize{9.0pt}{10.0pt}\selectfont
\bibliography{ReferencesCong}
\bibliographystyle{aaai}

\normalsize
\newpage
\appendix

\begin{center}
	\large{\textbf{Appendix of\\
			Bridging Maximum Likelihood and Adversarial Learning via $\alpha$-Divergence}}
	\\
	 \vspace{0.5 cm}
		{\small 
			\textbf{
				Miaoyun Zhao
				\qquad
				Yulai Cong
				\qquad
				Lawrence Carin}
			\\
			Department of Electrical and Computer Engineering, Duke University
		}
\end{center}

\normalsize

\section{Derivations of the Twin Gradients of $\alpha$-Divergence}
\label{sec:derive_twin_grad}

The $\alpha$-divergence between the model distribution $p_{\thetav}(\xv)$ and the underlying data distribution $q(\xv)$ is defined as  
\begin{equation}
    \Dc_{\alpha} [p_{\thetav}(\xv) \| q(\xv)]
    = \frac{1}{\alpha (1 - \alpha)} \Big[
    1 - \int p_{\thetav}(\xv)^{\alpha} q(\xv)^{1-\alpha} d\xv \Big].
\end{equation}
The corresponding gradient wrt $\thetav$ is
\begin{equation}
\begin{aligned}
    & \nabla_{\thetav} \Dc_{\alpha} [p_{\thetav}(\xv) \| q(\xv)]
    \\
    & 
    = \frac{1}{1 - \alpha} \Big[
    - \int p_{\thetav}(\xv)^{\alpha - 1} q(\xv)^{1-\alpha} \nabla_{\thetav} p_{\thetav}(\xv) d\xv \Big].
\end{aligned}
\end{equation}
By extracting $q(\xv)$ to form an expectation-based expression, we have
\begin{equation}
\begin{aligned}
    \nabla_{\thetav} & \Dc_{\alpha} [p_{\thetav}(\xv) \| q(\xv)]
    \triangleq \nabla_{\thetav} \Dc_{\alpha}^F
    \\
    & = \frac{1}{1 - \alpha} \Ebb_{q(\xv)} \Big[
    - p_{\thetav}(\xv)^{\alpha - 1} q(\xv)^{-\alpha} \nabla_{\thetav} p_{\thetav}(\xv) \Big]
    \\
    & 
    = \frac{1}{1-\alpha} \Ebb_{q(\xv)} \bigg[
    -\Big[ \frac{p_{\thetav}(\xv)}{q(\xv)} \Big]^{\alpha}
    \nabla_{\thetav} \log p_{\thetav}(\xv)
    \bigg].
\end{aligned}
\end{equation}
Clearly, the gradient information $\nabla_{\thetav} \log p_{\thetav}(\xv)$ from maximum likelihood learning (forward KL) is used as a building block.

For the gradient expression related to $p_{\thetav}(\xv)$ modeled in \eqref{eq:p_theta_x_gan} we rely on the stop-gradient operator\footnote{For example, {\tt tf.stop\_gradient} in TensorFlow \cite{tensorflow2015-whitepaper} and {\tt torch.no\_grad} in PyTorch \cite{paszke2017automatic}. 
If the stop-gradient operator $\mathbox{\xv}$ is applied to a function $f(\xv)$, the function value remains the same, \ie $f(\mathbox{\xv}) = f(\xv)$, but its gradient is removed, \ie $\nabla_{\xv} f(\mathbox{\xv}) = \bds 0$.} to simplify our derivations as
\begin{equation}
\resizebox{1\hsize}{!}{$
\begin{aligned}
    & \nabla_{\thetav} \Dc_{\alpha} [p_{\thetav}(\xv) \| q(\xv)]
    \\
    & \triangleq \nabla_{\thetav} \Dc_{\alpha}^R
    \\
    & = \frac{1}{1 - \alpha} \nabla_{\thetav} \Big[
    - \int p_{\thetav}(\xv)
    p_{\mathbox{\thetav}}(\xv)^{\alpha - 1} q(\xv)^{1-\alpha}   d\xv \Big]
    \\
    & 
    = \frac{1}{1 - \alpha} \nabla_{\thetav} 
    \Ebb_{p_{\thetav}(\xv)} \bigg[ -
    \Big[ \frac{q(\xv)}{p_{\mathbox{\thetav}}(\xv)} \Big]^{1-\alpha}
    \bigg]
    \\
    & = \frac{1}{1 - \alpha} \nabla_{\thetav} 
    \Ebb_{p(\zv)} \bigg[ -
    \bigg[
    \Big[ \frac{q(\xv)}{p_{\mathbox{\thetav}}(\xv)} \Big]^{1-\alpha}
    \bigg]_{\xv = G_{\thetav}(\zv)}
    \bigg]
    \\
    & = \frac{1}{1 - \alpha} \Ebb_{p(\zv)} \bigg[ -
    [\nabla_{\thetav} G_{\thetav}(\zv)]
    \bigg[ \nabla_{\xv} 
    \Big[ \frac{q(\xv)}{p_{\mathbox{\thetav}}(\xv)} \Big]^{1-\alpha}
    \bigg]_{\xv = G_{\thetav}(\zv)}
    \bigg]
    \\
    & = \frac{1}{1 - \alpha} \Ebb_{p(\zv) \delta(\xv | G_{\thetav}(\zv))} \bigg[
    -[\nabla_{\thetav} G_{\thetav}(\zv)]
    \bigg[ \nabla_{\xv} 
    \Big[ \frac{q(\xv)}{p_{\mathbox{\thetav}}(\xv)} \Big]^{1-\alpha}
    \bigg]
    \bigg]
    \\
    & = \Ebb_{p(\zv) \delta(\xv | G_{\thetav}(\zv))} \bigg[
    -[\nabla_{\thetav} G_{\thetav}(\zv)]
    \Big[\frac{q(\xv)}{p_{\thetav}(\xv)} \Big]^{1-\alpha}
    \Big[ \nabla_{\xv} \log \frac{q(\xv)}{p_{\mathbox{\thetav}}(\xv)} \Big]
    \bigg]
    \\
    & = \Ebb_{p(\zv) \delta(\xv | G_{\thetav}(\zv))} \bigg[
    [\nabla_{\thetav} G_{\thetav}(\zv)]
    \Big[\frac{q(\xv)}{p_{\thetav}(\xv)} \Big]^{1-\alpha}
    \Big[ \nabla_{\xv} \log \frac{p_{\mathbox{\thetav}}(\xv)}{q(\xv)} \Big]
    \bigg].
\end{aligned}
$}
\end{equation}
Note the $\nabla_{\thetav} \Dc_{\alpha}^R$ derived above is a special case of (and thus can be easily verified by) the recently proposed GO gradient \cite{cong2019go}. 


\section{Derivations for Semi-Implicit $p_{\thetav}(\xv)$}
\label{sec:app_semi_implicit}

Different from the main manuscript, in this section, we consider a semi-implicitly-defined model  
\beq\label{eq:semi_implicit_p_theta_x}
p_{\thetav}(\xv): 
\bali
    \xv & \sim p_{\thetav}(\xv | \zv) = \Nc(\xv | G_{\thetav}(\zv), \Sigmamat_{\thetav}(\zv))
    \\
    \zv & \sim p(\zv),
\eali
\eeq
where often $\Sigmamat_{\thetav}(\zv) = \sigma^2 \Imat$. This is reasonable because in practice, the observation noise (usually modeled as Gaussian) exists everywhere.
For specific applications, one may want to construct $\Sigmamat_{\thetav}(\zv)$ as neural networks.

The only difference between the generative process of the semi-implicit model in \eqref{eq:semi_implicit_p_theta_x} and that of the implicit one in \eqref{eq:p_theta_x_gan} of the main manuscript is that, one shall need an additional add-noise step to generate a fake sample.

Then considering the gradient of interest
\begin{equation}\label{eq:}
\begin{aligned}
    & \nabla_{\thetav} \Dc_{\alpha} [p_{\thetav}(\xv) \| q(\xv)]
    \\
    & 
    = \frac{1}{1 - \alpha} \Big[
    - \int p_{\thetav}(\xv)^{\alpha - 1} q(\xv)^{1-\alpha} \nabla_{\thetav} p_{\thetav}(\xv) d\xv \Big],
\end{aligned}
\end{equation}
we derive the equivalent twin gradients similarly as in the main manuscript as
\begin{equation}\label{eq:}
\begin{aligned}
    & \nabla_{\thetav} \Dc_{\alpha} [p_{\thetav}(\xv) \| q(\xv)]
    \triangleq \nabla_{\thetav} \Dc_{\alpha}^F
    \\
    & 
    = \frac{1}{1-\alpha} \Ebb_{q(\xv)} \bigg[
    -\Big[ \frac{p_{\thetav}(\xv)}{q(\xv)} \Big]^{\alpha}
    \nabla_{\thetav} \log p_{\thetav}(\xv)
    \bigg],
\end{aligned}
\end{equation}
\begin{equation}\label{eq:app_GO_grad_semi_implicit}
\begin{aligned}
    & \nabla_{\thetav} \Dc_{\alpha} [p_{\thetav}(\xv) \| q(\xv)]
    \\
    & \triangleq \nabla_{\thetav} \Dc_{\alpha}^R
    \\
    & 
    = \frac{1}{1 - \alpha} \Big[
    - \int \Big[ \frac{q(\xv)}{p_{\thetav}(\xv)} \Big]^{1-\alpha} \nabla_{\thetav} p_{\thetav}(\xv) d\xv \Big]
    \\
    & = \frac{1}{1 - \alpha} \nabla_{\thetav} 
    \Ebb_{p_{\thetav}(\xv)} \bigg[ -
    \Big[ \frac{q(\xv)}{p_{\mathbox{\thetav}}(\xv)} \Big]^{1-\alpha}
    \bigg]
    \\
    & = \frac{1}{1 - \alpha} \nabla_{\thetav} 
    \Ebb_{p(\zv) p_{\thetav}(\xv|\zv)} \bigg[ -
    \Big[ \frac{q(\xv)}{p_{\mathbox{\thetav}}(\xv)} \Big]^{1-\alpha}
    \bigg]
    \\
    & = \frac{1}{1 - \alpha} \Ebb_{p(\zv)} \left[ 
    \nabla_{\thetav} \Ebb_{p_{\thetav}(\xv|\zv)} 
    \bigg[ -
    \Big[ \frac{q(\xv)}{p_{\mathbox{\thetav}}(\xv)} \Big]^{1-\alpha}
    \bigg] \right]
    \\
    & = \frac{1}{1 - \alpha} \Ebb_{p(\zv) p_{\thetav}(\xv|\zv)} \bigg[ -
    [\Gbb^{p_{\thetav}(\xv|\zv)}_{\thetav}] 
    \bigg[ \nabla_{\xv} 
    \Big[ \frac{q(\xv)}{p_{\mathbox{\thetav}}(\xv)} \Big]^{1-\alpha}
    \bigg]
    \bigg]
    \\
    & = \Ebb_{p(\zv) p_{\thetav}(\xv|\zv)} \bigg[
    [\Gbb^{p_{\thetav}(\xv|\zv)}_{\thetav}] 
    \Big[\frac{q(\xv)}{p_{\thetav}(\xv)} \Big]^{1-\alpha}
    \Big[ \nabla_{\xv} \log \frac{p_{\mathbox{\thetav}}(\xv)}{q(\xv)} \Big]
    \bigg],
\end{aligned}
\end{equation}
where the variable-nabla $\Gbb^{p_{\thetav}(\xv|\zv)}_{\thetav}$, defined in \cite{cong2019go}, can be interpreted as the gradient of random variables $\xv$ wrt $\thetav$ conditional on $\zv$.

Accordingly, we have our chosen gradient
\begin{equation}\label{eq:}
\begin{aligned}
    & \nabla_{\thetav} \Dc_{\alpha} [p_{\thetav}(\xv) \| q(\xv)]
    \\
    & = (1 - \gamma_{\alpha}) \nabla_{\thetav} \Dc_{\alpha}^F
    + \gamma_{\alpha} \nabla_{\thetav} \Dc_{\alpha}^R
    \\
    & = \frac{1 - \gamma_{\alpha}}{1 - {\alpha}} \Ebb_{q(\xv)} \bigg[
    -\Big[ \frac{p_{\thetav}(\xv)}{q(\xv)} \Big]^{\alpha}
    \nabla_{\thetav} \log p_{\thetav}(\xv)
    \bigg] + 
    \\
    & \quad 
    \gamma_{\alpha} \Ebb_{p(\zv) p_{\thetav}(\xv|\zv)} \bigg[
    [\Gbb^{p_{\thetav}(\xv|\zv)}_{\thetav}] 
    \Big[\frac{q(\xv)}{p_{\thetav}(\xv)} \Big]^{1-\alpha}
    \Big[ \nabla_{\xv} \log \frac{p_{\thetav}(\xv)}{q(\xv)} \Big]
    \bigg].
\end{aligned}
\end{equation}

For the practical implementation associated with $\nabla_{\thetav} \log p_{\thetav}(\xv)$, one could simply resort to the ELBO technique of the classic maximum likelihood learning, namely
\beq\label{eq:}
\bali
    \nabla_{\thetav} \log p_{\thetav}(\xv) 
    & = \nabla_{\thetav} \Ebb_{q_{\phiv^{*}}(\zv|\xv)} \Big[
    \log \frac{p_{\thetav}(\xv, \zv)}{q_{\phiv^{*}}(\zv | \xv)}
    \Big]
    \\
    & 
    = \Ebb_{q_{\phiv^{*}}(\zv|\xv)} \big[
    \nabla_{\thetav} \log p_{\thetav}(\xv, \zv)
    \big],
\eali
\eeq
where $p_{\thetav}(\xv, \zv)$ is computable thanks to its semi-implicit definition in \eqref{eq:semi_implicit_p_theta_x}, different from what's in the main manuscript for the challenging implicit definition as in \eqref{eq:p_theta_x_gan}.

For the log-ratio term $\log \frac{p_{\thetav}(\xv)}{q(\xv)}$, it is a common practice in GAN literature to solve
\begin{equation}\label{eq:}
\begin{aligned}
    \max_{\betav} \Ebb_{q(\xv)} [\log \sigma(f_{\betav}(\xv))]
    + \Ebb_{p_{\thetav} (\xv)} [\log (1-\sigma(f_{\betav}(\xv))) ],
\end{aligned}
\end{equation}
for $f_{\betav^{*}}(\xv) = \log \frac{q(\xv)}{p_{\thetav}(\xv)}$. The gradient information $\nabla_{\xv} f_{\betav^{*}}(\xv) = \nabla_{\xv} \log \frac{q(\xv)}{p_{\thetav}(\xv)}$ is then exploited for learning the generative model $p_{\thetav}(\xv)$.

Note in this case, we need not to worry about the gap brought by the implicit model as in the main manuscript. 

\section{Derivations for the Two Generalizations of the $\alpha$-Bridge}
\label{sec:Two_connections_cycle}

We first summarize the related parameterized models used in this section. 
\beq
\bali
    & \tilde p_{\thetav}(\xv): \xv \sim \Nc(\xv | G_{\thetav}(\zv), \sigma^2 \Imat), \zv \sim p(\zv)
    \\
    & q_{\phiv}(\zv): \zv \sim \Nc(\zv | E_{\phiv}(\xv), \sigma^2 \Imat), \xv \sim q(\xv)
\eali
\eeq
Note we use a simplified variational approximation $ q_{\phiv}(\zv|\xv) = \Nc(\zv | E_{\phiv}(\xv), \sigma^2 \Imat)$ (instead of $q_{\phiv}(\zv|\xv) = \Nc(\zv | \muv_{\phiv}(\xv), \sigmav^2 (\xv) \Imat)$) to draw connections to the existing CycleGAN and ALICE. We use $E_{\phiv}(\xv)$ in place of $\muv_{\phiv}(\xv)$ to highlight the physical meaning of ``encoder'' and also to keep consistent with the existing CycleGAN and ALICE.

\textbf{Revealing Connection to CycleGAN with Marginals:}

Start from the practical gradient in \eqref{eq:combine_grad_practical} of the main manuscript, we first backward-derive the corresponding identical objective (whose gradient is the same as \eqref{eq:combine_grad_practical}) to reveal the interesting interpretation in \eqref{eq:cycle_reverseKL} of the main manuscript, namely
\begin{equation}\label{eq:app_cycle_1}
\begin{aligned}
\nabla_{\thetav} & \Dc_{\alpha} [p_{\thetav}(\xv) \| q(\xv)]
    \\
    & 
\approx \nabla_{\thetav} \left[
\bali
& \frac{1 - \gamma_{\alpha}}{1 - \alpha}
\Ebb_{q(\xv) q_{\phiv^{*}}(\zv|\xv)} \Big[ 
\frac{e^{- \alpha f_{\betav^{*}}(\xv)}}{ 2\sigma^2}
\| \xv - G_{\thetav}(\zv) \|_2^2
\Big] 
\\
& + \gamma_{\alpha} \Ebb_{p_{\thetav}(\xv)} \big[
-e^{(1 - \alpha) f_{\betav^{*}}(\mathbox{\xv})}
f_{\betav^{*}}(\xv) 
\big]
\eali
\right],
\end{aligned}
\end{equation}
where $\mathbox{\xv}$ is the stop-gradient operator detailed in Appendix \ref{sec:derive_twin_grad}.
Similarly to the case in the $\xv$ space, one can also derive another formula in the $\zv$ space as 
\begin{equation}\label{eq:app_cycle_2}
\begin{aligned}
    & \nabla_{\phiv} \Dc_{\alpha} [q_{\phiv}(\zv) \| p(\zv)]
    \\
    & 
    \approx \nabla_{\phiv} \left[
    \bali
        & \frac{1 - \gamma_{\alpha}}{1 - \alpha} \Ebb_{p(\zv) p_{\thetav^{*}}(\xv|\zv)} \Big[ 
        \frac{e^{- \alpha g_{\gammav^{*}}(\zv)}}{2\sigma^2}
        \| \zv - E_{\phiv}(\xv) \|_2^2
        \Big] 
        \\
        &  + \gamma_{\alpha} \Ebb_{q_{\phiv}(\zv)} \big[
        -e^{(1 - \alpha) g_{\gammav^{*}}(\mathbox{\zv})}
        g_{\gammav^{*}}(\zv)
        \big]
    \eali
    \right],
\end{aligned}
\end{equation}
where $g_{\gammav}(\zv)$ corresponds to the additional discriminator in the $\zv$ space, which can be similarly solved via
\begin{equation}\label{eq:}
\begin{aligned}
\max_{\gammav} \Ebb_{p(\zv)} [\log \sigma(g_{\gammav}(\zv))]
+ \Ebb_{q_{\phiv} (\zv)} [\log (1-\sigma(g_{\gammav}(\zv))) ].
\end{aligned}
\end{equation}

By combining the above two formula together, we have 
\beq\label{eq:app_cycle_gan}
\bali
    & \Dc_{\alpha} [p_{\thetav}(\xv) \| q(\xv)] + \Dc_{\alpha} [q_{\phiv}(\zv) \| p(\zv)]
    \\
    & 
    \overset{\nabla}{\approx} 
    \left[
    \bali
        & \frac{1 - \gamma_{\alpha}}{1 - \alpha} \Ebb_{q(\xv) q_{\mathbox{\phiv}}(\zv|\xv)} \Big[ 
        \frac{e^{- \alpha f_{\betav^{*}}(\xv)}}{2\sigma^2}
        \| \xv - G_{\thetav}(\zv) \|_2^2
        \Big] 
        \\
        & + \frac{1 - \gamma_{\alpha}}{1 - \alpha} \Ebb_{p(\zv) p_{\mathbox{\thetav}}(\xv|\zv)} \Big[ 
        \frac{e^{- \alpha g_{\gamma^{*}}(\zv)}}{2\sigma^2}
        \| \zv - E_{\phiv}(\xv) \|_2^2
        \Big] 
        \\
        & + \gamma_{\alpha} \Ebb_{p_{\thetav}(\xv)} \Big[
        -e^{(1 - \alpha) f_{\betav^{*}}(\mathbox{\xv})}
        f_{\betav^{*}}(\xv)
        \Big]
        \\
        & + \gamma_{\alpha} \Ebb_{q_{\phiv}(\zv)} \Big[
        -e^{(1 - \alpha) g_{\gamma^{*}}(\mathbox{\zv})}
        g_{\gammav^{*}}(\zv)
        \Big]
    \eali
    \right],
\eali
\eeq
where $\overset{\nabla}{\approx}$ means both sides have approximately equal gradients like those in \eqref{eq:app_cycle_1} and \eqref{eq:app_cycle_2}. 
It is obvious that the right hand side very much resembles the CycleGAN objective \cite{zhu2017unpaired}, with a few differences including
\begin{itemize}

    \item equation \eqref{eq:app_cycle_gan} generalized from our $\alpha$-Bridge has the ratio-related weights within its four terms (with the first two associated with cycle-consistency and the last two related to reverse-KL-based adversarial learning); by contrast, CycleGAN employed user-defined hyperparameters to balance its four terms.

    \item CycleGAN used the original GAN loss \cite{goodfellow2016nips}, while equation \eqref{eq:app_cycle_gan} utilizes the reverse-KL-based GAN objective (the last two terms). In fact, many of the multiple forms of GAN are closely related \cite{nowozin2016f} and they performs similar with the same computational budget and enough hyperparameter optimization \cite{kurach2018gan,GoogleCompareGAN}.
    
    \item equation \eqref{eq:app_cycle_gan} stops parts of the gradients of its first two cycle-consistency terms, while CycleGAN used both gradients wrt $\phiv$ and $\thetav$ from each cycle-consistency term.
    
    
\end{itemize}


\textbf{Revealing Connection to ALICE in the Joint Space:}

Similar to the above derivations, we first take a step back to re-express the chosen gradient in \eqref{eq:combine_grad} of the main manuscript as
\begin{equation}\label{eq:}
\begin{aligned}
    & \nabla_{\thetav} \Dc_{\alpha} [p_{\thetav}(\xv) \| q(\xv)]
    = (1 - \gamma_{\alpha}) \nabla_{\thetav} \Dc_{\alpha}^F
    + \gamma_{\alpha} \nabla_{\thetav} \Dc_{\alpha}^R
    \\
    & = \frac{1 - \gamma_{\alpha}}{1 - \alpha} \Ebb_{q(\xv)} \bigg[
    -\Big[ \frac{p_{\thetav}(\xv)}{q(\xv)} \Big]^{\alpha}
    \nabla_{\thetav} \log p_{\thetav}(\xv)
    \bigg] + 
    \\
    & \quad \gamma_{\alpha} \Ebb_{p(\zv) \delta(\xv | G_{\thetav}(\zv))} \bigg[
    [\nabla_{\thetav} G_{\thetav}(\zv)]
    \Big[ \frac{q(\xv)}{p_{\thetav}(\xv)} \Big]^{1-\alpha}
    \Big[ \nabla_{\xv} \log \frac{p_{\thetav}(\xv)}{q(\xv)} \Big]
    \bigg].
    \\
    & = \nabla_{\thetav} \left[
    \bali
        & \frac{1 - \gamma_{\alpha}}{1 - \alpha} \Ebb_{q(\xv)} \left[ -
        \left[ \frac{p_{\mathbox{\thetav}}(\xv)}{q(\xv)} \right]^{\alpha}
        \log p_{\thetav}(\xv)
        \right] + 
        \\
        & \gamma_{\alpha}\Ebb_{p_{\thetav}(\xv)} \left[
        \mathbox{\left[ \frac{q(\xv)}{p_{{\thetav}}(\xv)} \right]^{1-\alpha}}
        \log \frac{p_{\mathbox{\thetav}}(\xv)}{q(\xv)} 
        \right]
    \eali
    \right]
\end{aligned}
\end{equation}
By replacing $p_{\thetav}(\xv)$ and $q(\xv)$ with $p_{\thetav}(\xv, \zv)$ and $q_{\phiv}(\xv,\zv)$, respectively, and using another discriminator $h_{\etav^{*}}(\xv, \zv) = \log \frac{q_{\phiv}(\xv,\zv)}{p_{\thetav}(\xv, \zv)} $ in the joint space, we have
\begin{equation}\label{eq:}
\begin{aligned}
    & \nabla_{\thetav} \Dc_{\alpha} [p_{\thetav}(\xv, \zv) \| q_{\phiv}(\xv,\zv)]]
    \\
    & \approx \nabla_{\thetav} \left[
    \bali
        & \frac{1 - \gamma_{\alpha}}{1 - \alpha} \Ebb_{q_{\phiv}(\xv,\zv)} \big[ -
        e^{- \alpha h_{\etav^{*}}(\xv, \zv)}
        \log p_{\thetav}(\xv, \zv)
        \big] + 
        \\
        & \gamma_{\alpha} \Ebb_{p_{\thetav}(\xv, \zv)} \big[-
        e^{(1 - \alpha) h_{\etav^{*}}(\mathbox{\xv}, \zv)}
        h_{\etav^{*}}(\xv, \zv) 
        \big]
    \eali
    \right],
\end{aligned}
\end{equation}
By mirroring the above equation, we have
\begin{equation}\label{eq:}
\begin{aligned}
    & \nabla_{\phiv} \Dc_{\alpha} [q_{\phiv}(\xv,\zv) \| p_{\thetav}(\xv, \zv)]]
    \\
    & \approx \nabla_{\phiv} \left[
    \bali
        & \frac{1 - \gamma_{\alpha}}{1 - \alpha} \Ebb_{p_{\thetav}(\xv, \zv)} \big[ -
        e^{ \alpha h_{\etav^{*}}(\xv, \zv)}
        \log q_{\phiv}(\xv,\zv)
        \big] + 
        \\
        & \gamma_{\alpha} \Ebb_{q_{\phiv}(\xv,\zv)} \big[
        e^{-(1 - \alpha) h_{\etav^{*}}(\xv, \mathbox{\zv})}
        h_{\etav^{*}}(\xv, \zv)
        \big]
    \eali
    \right].
\end{aligned}
\end{equation}

By combining the above two equation together mimicking \eqref{eq:app_cycle_gan}, we have
\beq\label{eq:app_alice_gan}
\bali
    & \Dc_{\alpha} [p_{\thetav}(\xv, \zv) \| q_{\mathbox{\phiv}}(\xv,\zv)]] + \Dc_{\alpha} [q_{\phiv}(\xv,\zv) \| p_{\mathbox{\thetav}}(\xv, \zv)]]
    \\
    & \qquad\qquad
    \overset{\nabla}{\approx} 
    \left[
    \bali
        & \frac{1 - \gamma_{\alpha}}{1 - \alpha} \Ebb_{q_{\mathbox{\phiv}}(\xv,\zv)} \big[ -
        e^{- \alpha h_{\etav^{*}}(\xv, \zv)}
        \log p_{\thetav}(\xv| \zv)
        \big] 
        \\
        & + \frac{1 - \gamma_{\alpha}}{1 - \alpha} \Ebb_{p_{\mathbox{\thetav}}(\xv, \zv)} \Big[ 
        -e^{ \alpha h_{\etav^{*}}(\xv, \zv)}
        \log q_{\phiv}(\zv | \xv) \Big]
        \\
        & + \gamma_{\alpha} \Ebb_{p_{\thetav}(\xv, \zv)} \big[-
        e^{(1 - \alpha) h_{\etav^{*}}(\mathbox{\xv}, \zv)}
        h_{\etav^{*}}(\xv, \zv) 
        \big]
        \\
        & + \gamma_{\alpha} \Ebb_{q_{\phiv}(\xv,\zv)} \big[
        e^{-(1 - \alpha) h_{\etav^{*}}(\xv, \mathbox{\zv})}
        h_{\etav^{*}}(\xv, \zv)
        \big]
    \eali
    \right].
\eali
\eeq
Accordingly, we have the right hand side resembling ALICE \cite{li2017alice}, as stated in the main manuscript.
Despite the ratio-related weights, the first two terms are closely related to the cycle-consistency, which was exploited in ALICE to develop a lower bound for conditional entropy.
The last two terms are the reverse-KL-based GAN losses, which are expected to have a similar performance to the original GAN loss used in ALICE \cite{kurach2018gan,GoogleCompareGAN}. 
Similar to \eqref{eq:app_cycle_gan}, parts of the gradient information are removed from the first two cycle-consistency terms.

\section{Adversarial Learning initialized by Maximum Likelihood Learning}
\label{sec:ML_AL_explode}

The details of how we get Figure \ref{fig:explodes} are given below. 
The experimental settings and model architectures from Appendix \ref{sec:app_exp_25Gaussian} are used. 

First, we follow the classic variational inference to train the model $p_{\thetav}(\xv) = \int p_{\thetav}(\xv | \zv) p(\zv) d\zv$ and the variational posterior $q_{\phiv}(\zv|\xv)$ by maximizing the ELBO in \eqref{eq:elbo_ML}. Simultaneously we use the generated fake samples and the true data samples to pretrain the discriminator $\Dc_{\betav}(\xv)$, with the objective in \eqref{eq:GAN_lossd}.
All $p_{\thetav}(\xv)$, $q_{\phiv}(\zv|\xv)$, and $\Dc_{\betav}(\xv)$ are saved for downstreaming experiments (like the transferring Step of the proposed $\alpha$-Bridge).

\begin{figure}[H]
	\begin{center}
		
		\subcaptionbox{Adam hyperparameter $\beta_1=0$ \label{fig:}}{
			\includegraphics[height=0.12 \columnwidth]{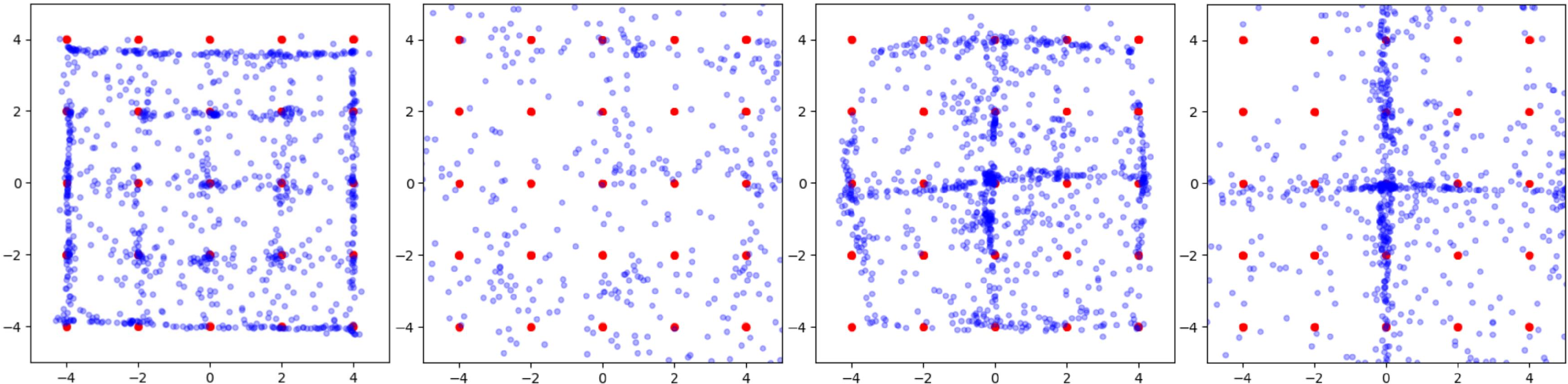}
			\includegraphics[height=0.12 \columnwidth]{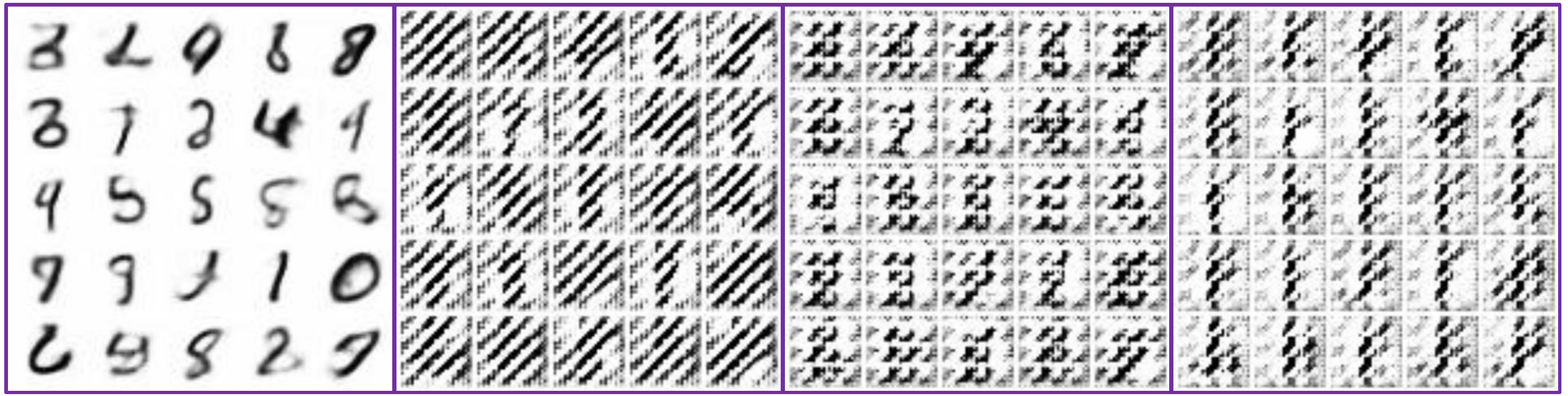}
		}
		\subcaptionbox{Adam hyperparameter $\beta_1=0.5$ \label{fig:}}{
			\includegraphics[height=0.12 \columnwidth]{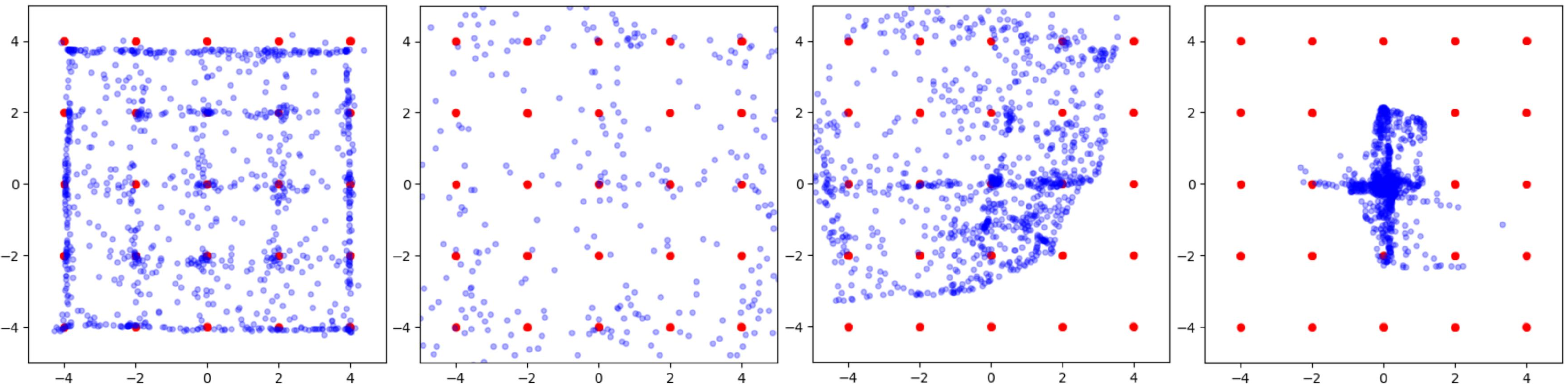}
			\includegraphics[height=0.12 \columnwidth]{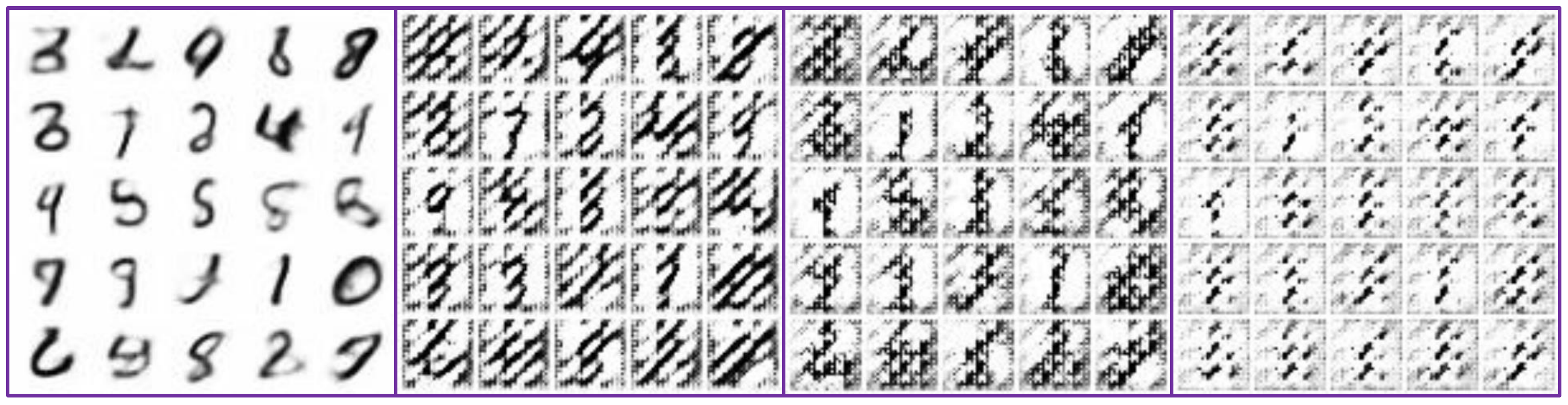}
		}
		\subcaptionbox{Adam hyperparameter $\beta_1=0.9$ \label{fig:}}{
			\includegraphics[height=0.12 \columnwidth]{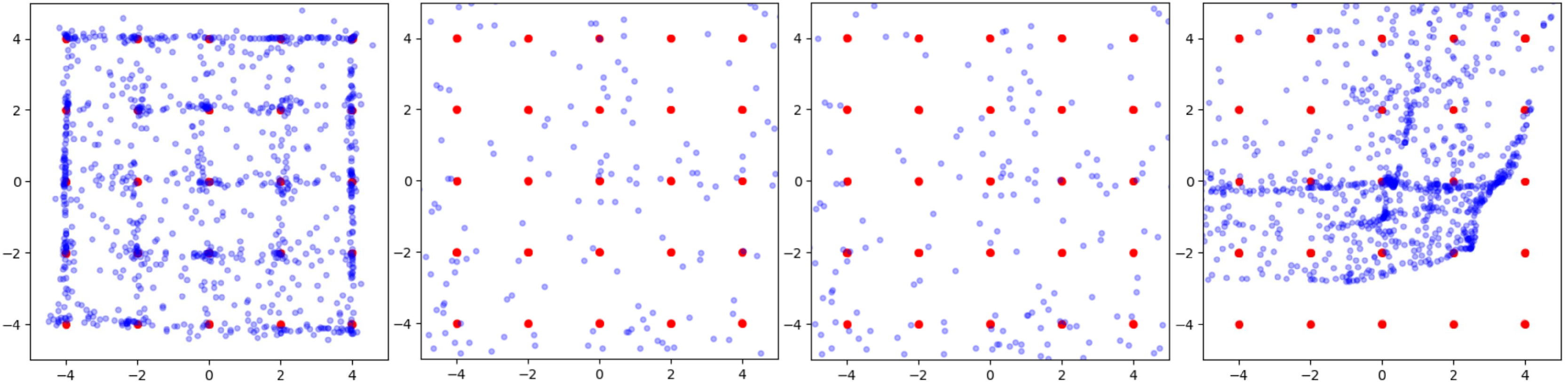}
			\includegraphics[height=0.12 \columnwidth]{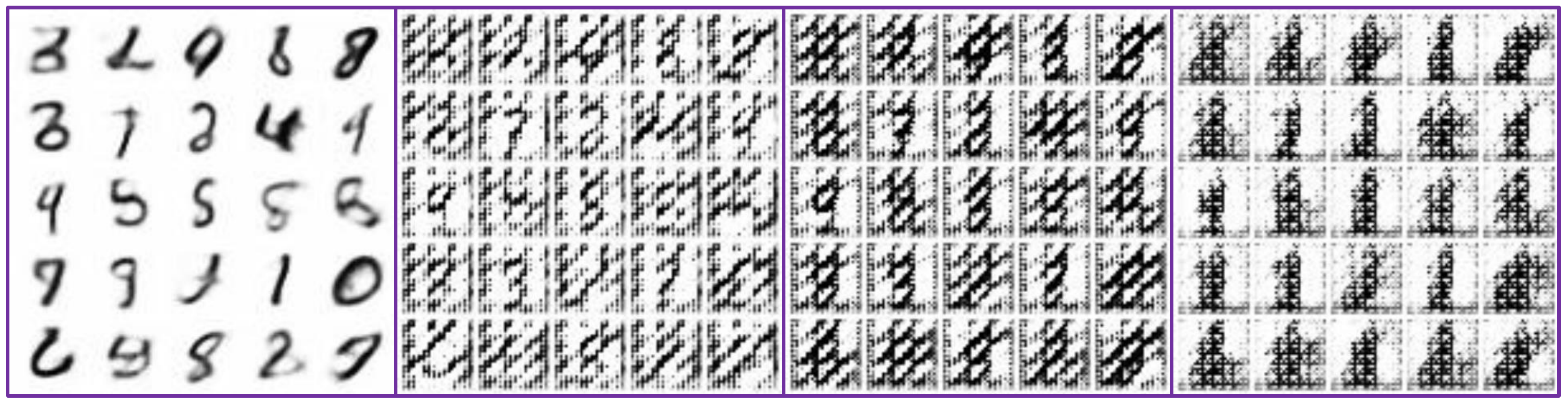}
		}
		\caption{\small 
			Demonstration of adversarial learning forgetting the information learned/initialized by ML learning on 25-Gaussians \cite{tao2018chi} (the left four) and MNIST \cite{lecun1998gradient} (the right four). From left to right are the results of ML initialization, $20$, $40$, and $60$ following iterations of adversarial learning, respectively. 
		}
		\label{fig:app_explodes}
	\end{center}
\end{figure}

Next we load the pretrained $p_{\thetav}(\xv)$ and $\Dc_{\betav}(\xv)$ for initialization and start the reverse-KL-based adversarial learning to train the parameters $\thetav$ and $\betav$ with the reverse KL objective in \eqref{eq:RKL_loss_grad} and the GAN objective in \eqref{eq:GAN_lossd}, respectively.
\textbf{Note the pretrained discriminator $\Dc_{\betav}(\xv)$ is meant for providing great help for adversarial learning. With a randomly initialized discriminator, worse forgetting is empirically observed for adversarial learning. By contrast, the $\alpha$-Bridge can inherit the information from ML even with a randomly initialized discriminator (which is usually the case since in practice one can only expect a pretrained $p_{\thetav}(\xv)$ and $q_{\phiv}(\zv|\xv)$ from the classic variational inference, where there is no place for the discriminator $\Dc_{\betav}(\xv)$).} The ML initialization, together with the following fake samples generated after $20$, $40$, and $60$ iterations of adversarial learning, are collected and demonstrated in Figure \ref{fig:app_explodes}.

\section{Empirical Evaluation of the Intuitive Choices of $\alpha$ and $\gamma_{\alpha}$}
\label{sec:alpha_w_alpha}

The $25$-Gaussians experiment (as shown in Section \ref{sec:exp_25Gaussian} and Appendix \ref{sec:app_exp_25Gaussian}) is adopted for empirical evaluation of different strategies for $\alpha$ and $\gamma_{\alpha}$ (see Table \ref{tab:alpha_w_alpha}).

\begin{table}[H]\centering
	\caption{Tested strategies for $\alpha$ and $\gamma_{\alpha}$.
		$t \in [0^{+}, 1^{-}]$ linearly corresponds to the iteration ${iter} \in [1,2,\cdots, {NumIters}]$; for example, $t = \epsilon + ({iter}-1) \frac{1-2\epsilon}{{NumIters}-1}$. ${NumIters}$ is the total number of iterations and $\epsilon$ is a small positive number.
		$a=c=20, b=d=-10$ in this experiment.
	}\label{tab:alpha_w_alpha}
	\resizebox{0.99\hsize}{!}{
	\begin{tabular}{c|c}\hline\hline
		$\alpha$  	& 	$\gamma_{\alpha}$   \\ \hline\hline
		Linear:	$\alpha(t) = t$
		& Linear: $\gamma_{\alpha} = \alpha$  \\ 
		Sigmoid: $\alpha(t) = \frac{\sigma(at+b) - \sigma(b)}{\sigma(a+b)-\sigma(b)}$
		& Sigmoid: $\gamma_{\alpha} = \frac{\sigma(c\alpha+d) - \sigma(d)}{\sigma(c+d)-\sigma(d)}$ \\ 
		X5:	$16(t-0.5)^5 + 0.5$
		& Cosine: $\gamma_{\alpha} = 0.5 - 0.5\cos(\pi \alpha)$ \\ 
      	& Square: $1 - (1-\alpha)^2$ \\ \hline\hline
	\end{tabular}
    }
\end{table}

\begin{figure}[H]
	\begin{center}
		\subcaptionbox{$\alpha$-Bridge-GP \label{fig:}}{
			\includegraphics[height=0.24 \columnwidth]{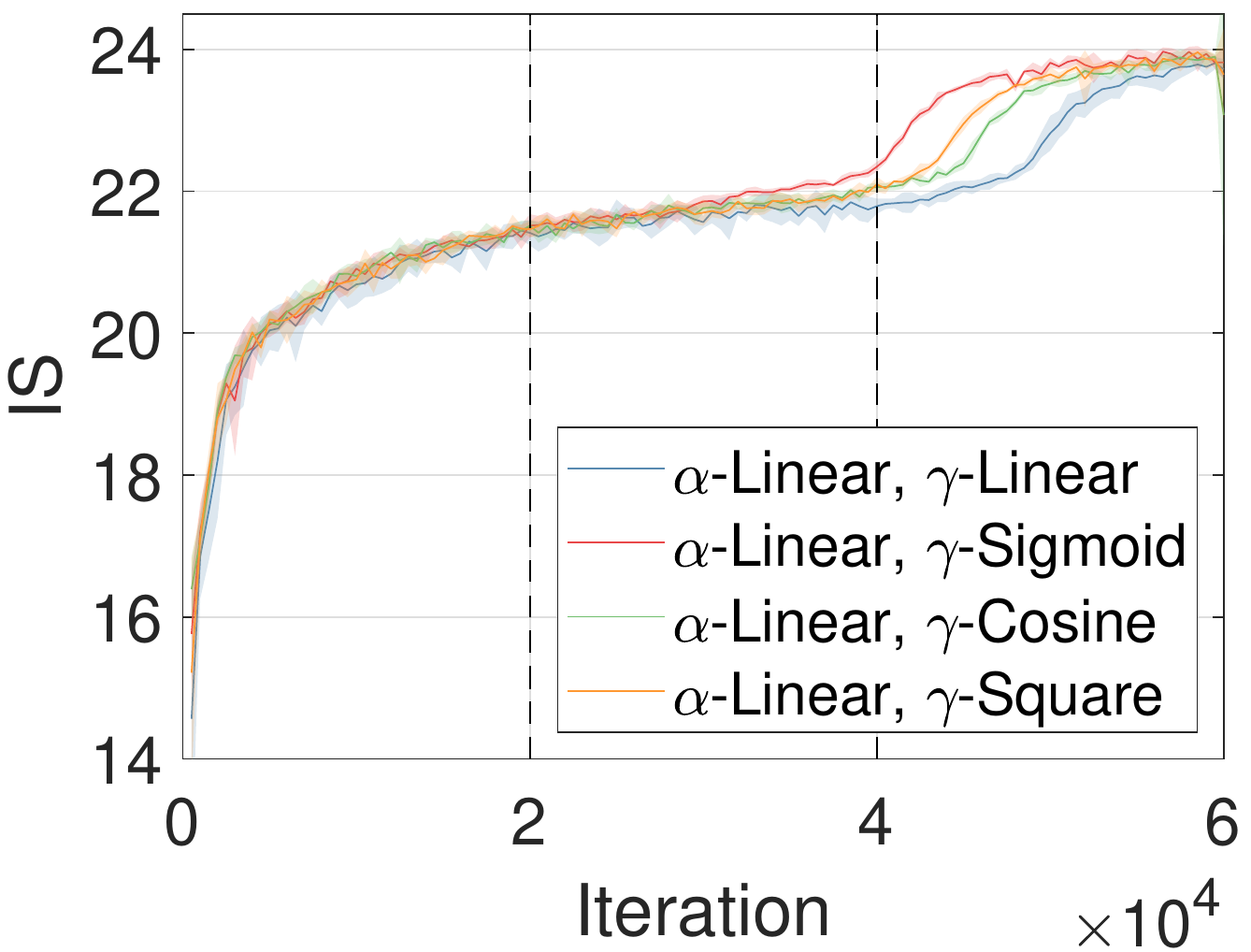}
			\includegraphics[height=0.24 \columnwidth]{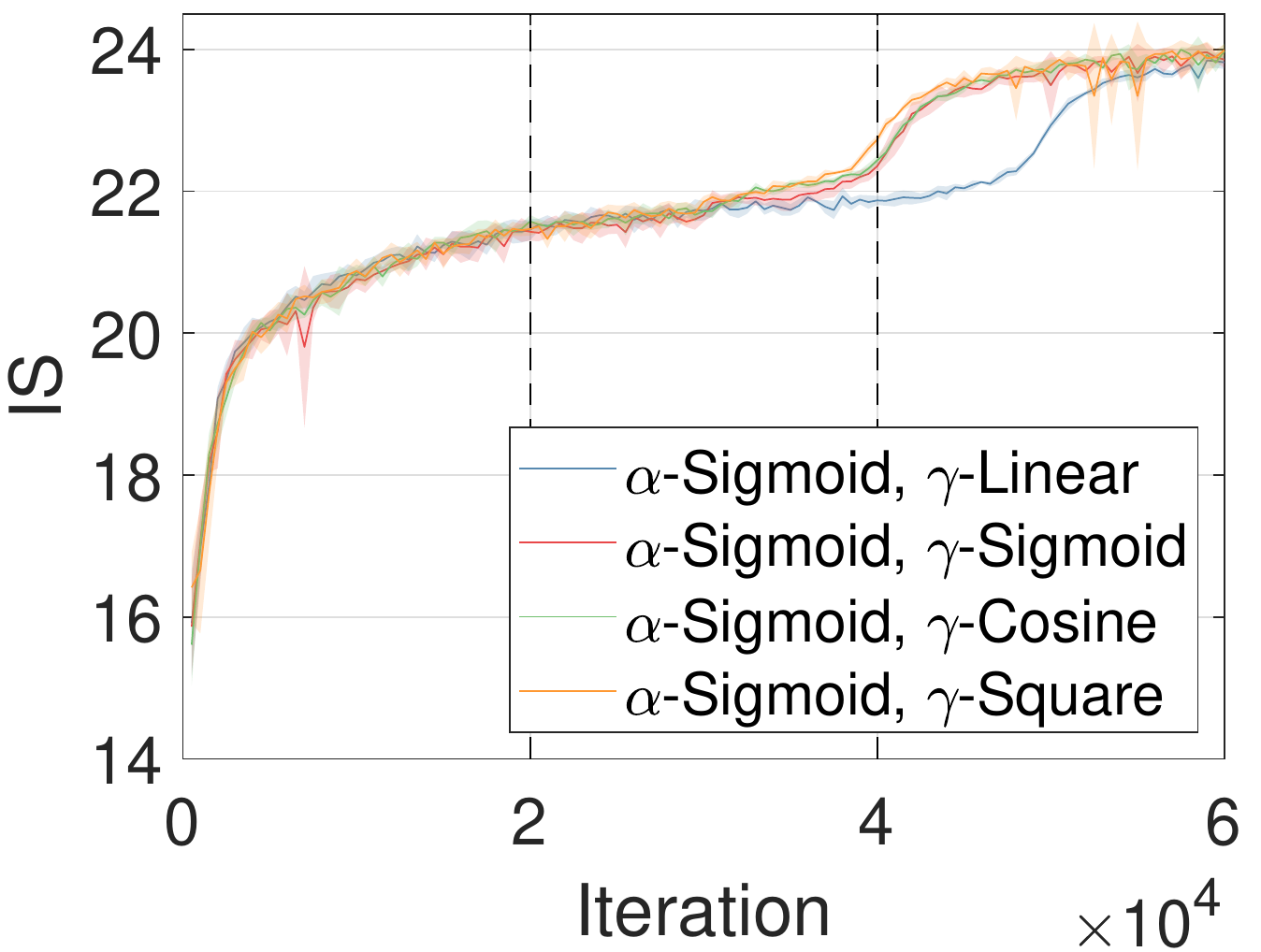}
			\includegraphics[height=0.24 \columnwidth]{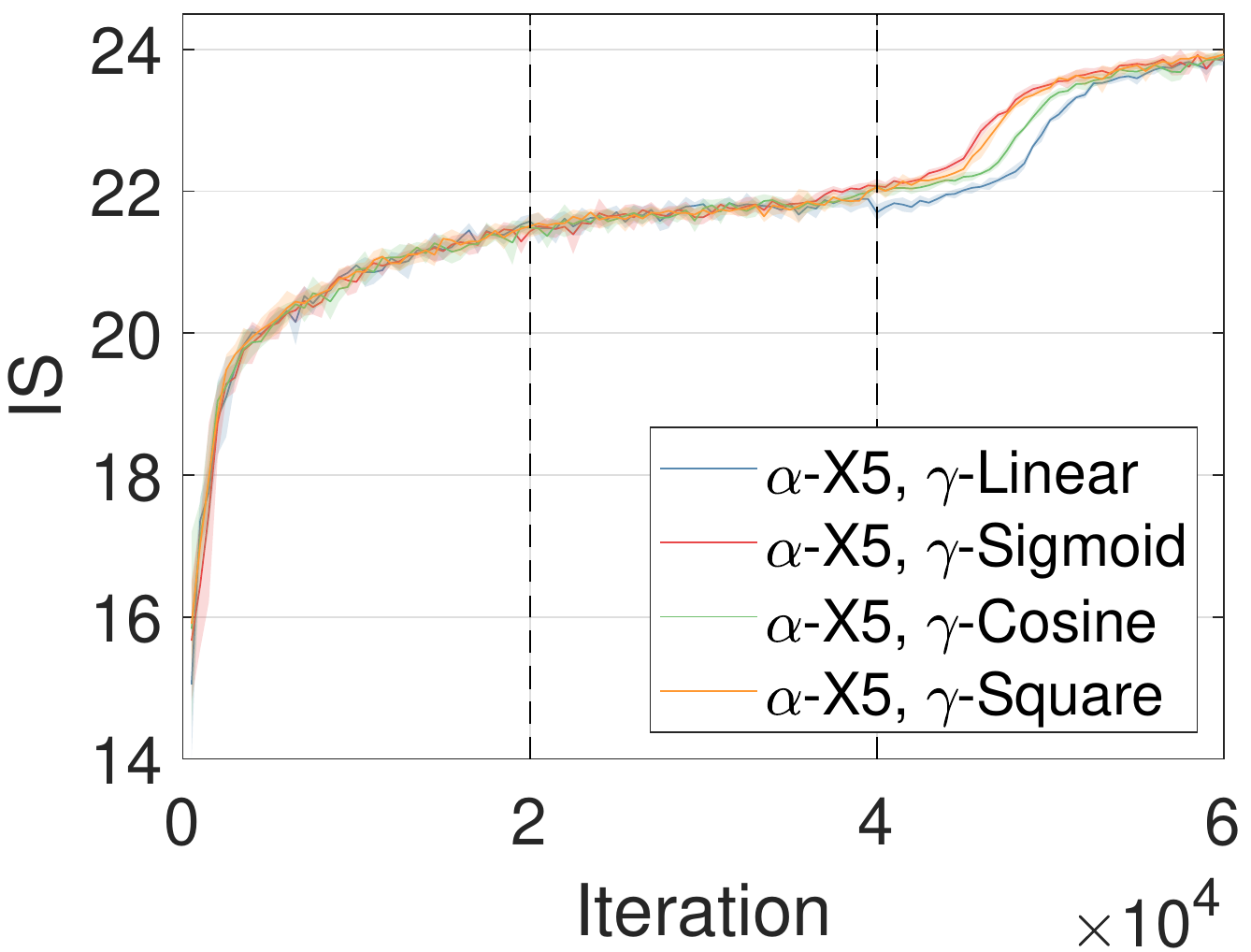}
		}
		\subcaptionbox{$\alpha$-Bridge-SN \label{fig:}}{
			\includegraphics[height=0.24 \columnwidth]{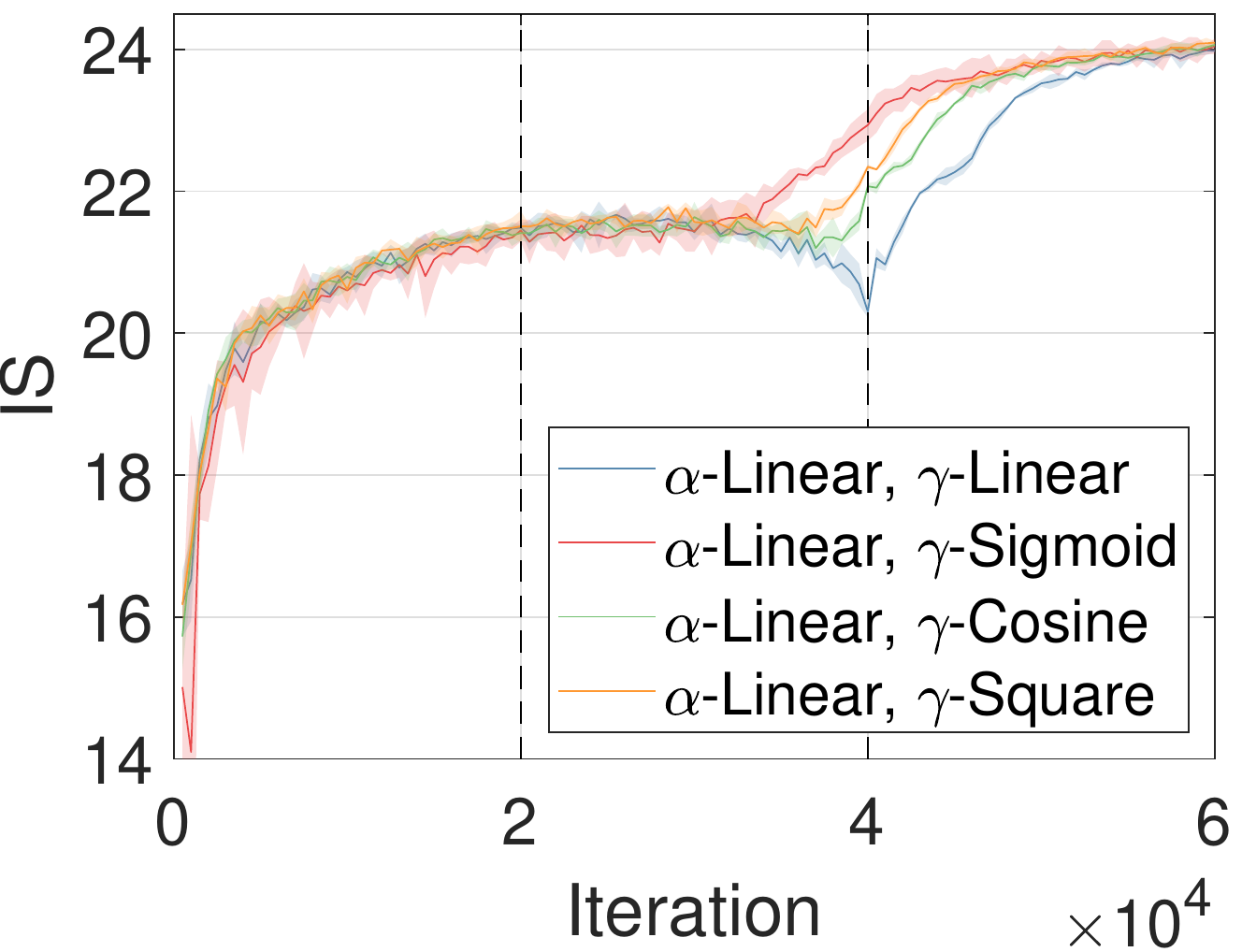}
			\includegraphics[height=0.24 \columnwidth]{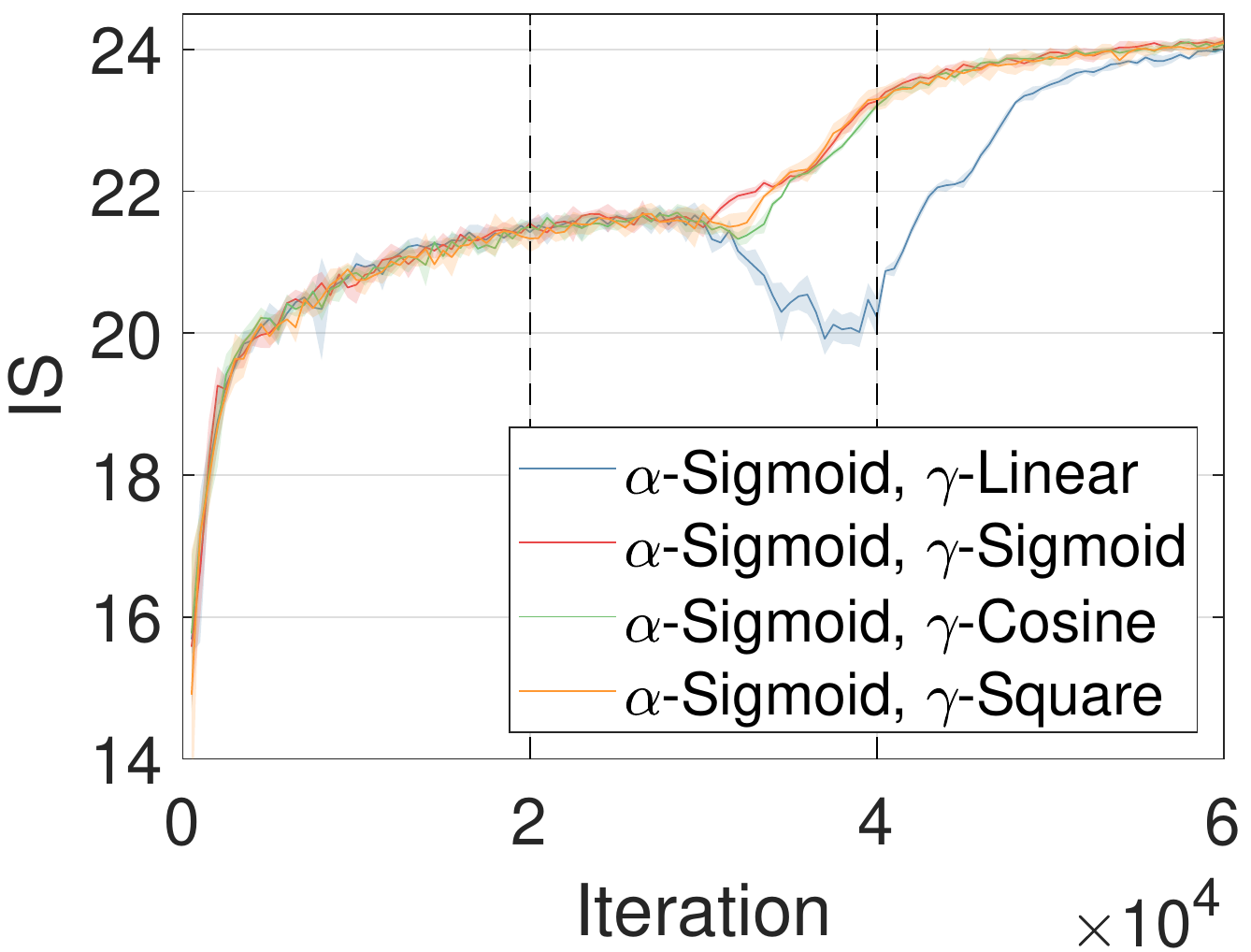}
			\includegraphics[height=0.24 \columnwidth]{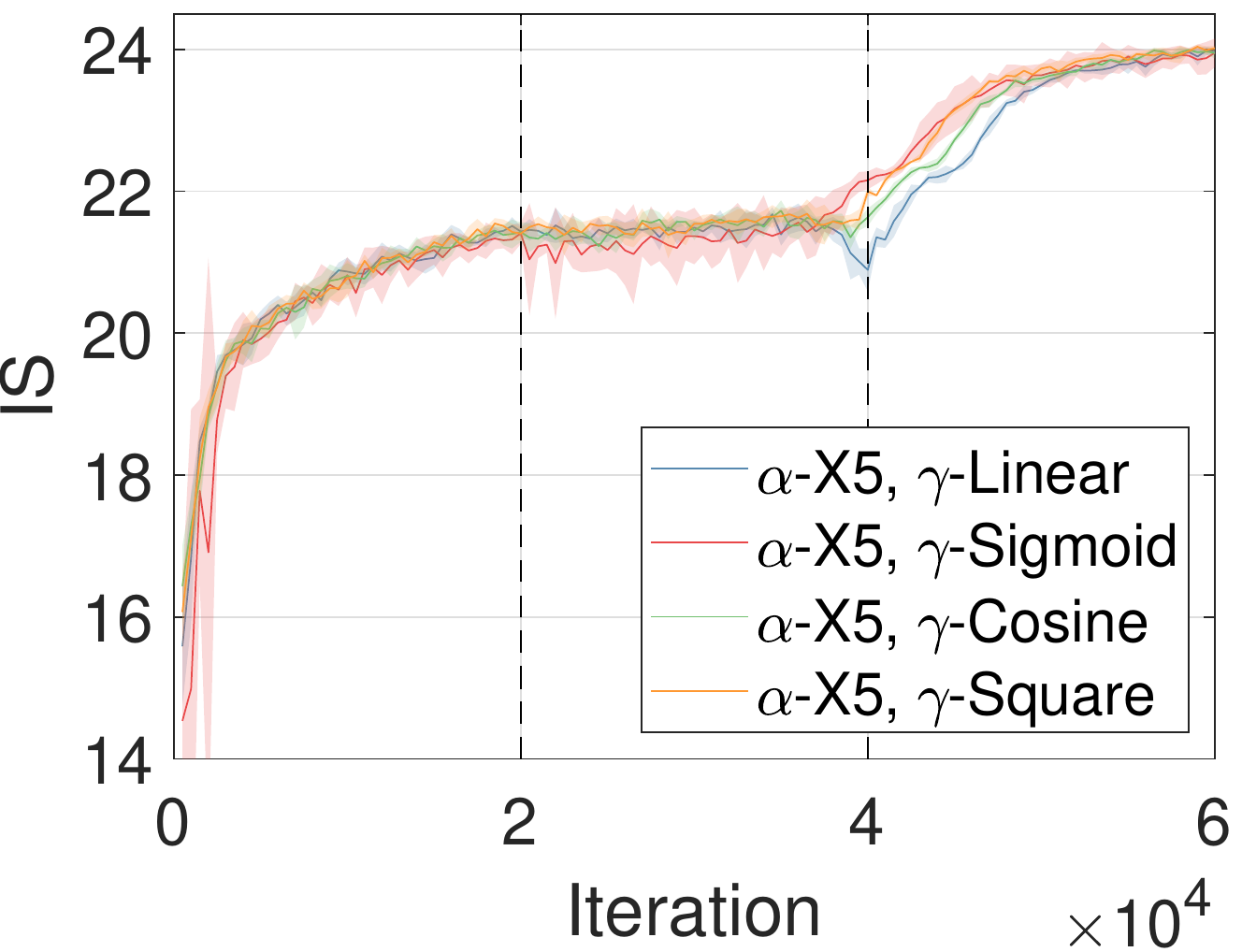}
		}
		\caption{\small Empirical evaluation of different strategies for $\alpha$ and $\gamma_{\alpha}$. The top$/$bottom row shows the results from $\alpha$-Bridge-GP$/$$\alpha$-Bridge-SN. Curves are calculated based on $10$ random trials. Two vertical dashed lines are used to indicate the three steps of the $\alpha$-Bridge.
		}
		\label{fig:alpha_w_alpha}
	\end{center}
\end{figure}

Figure \ref{fig:alpha_w_alpha} shows the experimental results from different combinations of strategies in Table \ref{tab:alpha_w_alpha}. Curves are calculated based on $10$ random trials. It seems $\alpha$-Sigmoid with $\gamma$-{Sigmoid} and $\alpha$-Linear with $\gamma$-{Sigmoid} work better in this experiment.

\section{The Performance of $\alpha$-Bridge Compared with The Other potential Combination Methods}
\label{sec:vs_FKLpRKL}
Here we discusses and compare $\alpha$-Bridge with three other potential approaches to combine adversarial and ML learning. These methods are, 
(i) FR, derectly add up the Forwad KL loss and Reverse KL loss;
(ii) FRw, weighting average the Forward and Reverse KL loss as $L = (1. - \gamma) L_{FKL} + \gamma L_{RKL}$ with $\gamma$ varying from $0$ to $1$;
(iii) FRw3, replace the second step of our method as FRw.

In the experiments we try to use a parameterized model $p_{\theta}(\xv)$ to match the real data distribution $q(\xv)$. Here $q(\xv)$ is a 1-D Gaussian mixture data with three components, each component has equal probability. The experiments are divided into two different cases: (i) the parameterized model $p_{\theta}(\xv)$ is expressive enough (have three components), and (ii) $p_{\theta}(\xv)$ has limited power (only have two components). For the first case, we choose three different ground truth distributions for $q(\xv)$ as shown in the first three rows of Table \ref{tab:vs_FKLpRKL_enough}. For each distribution, ten trials are conducted and the number of times that $p_{\theta}(\xv)$ perfectly matches $q(\xv)$ are recorded in the table. 
The results for the second case are shown in the last row of Table \ref{tab:vs_FKLpRKL_enough}. And we consider $p_{\theta}(\xv)$ matches two modes of $q(\xv)$ as successful case.

\begin{figure}[H]
	\begin{center}
		\includegraphics[width=0.8\columnwidth]{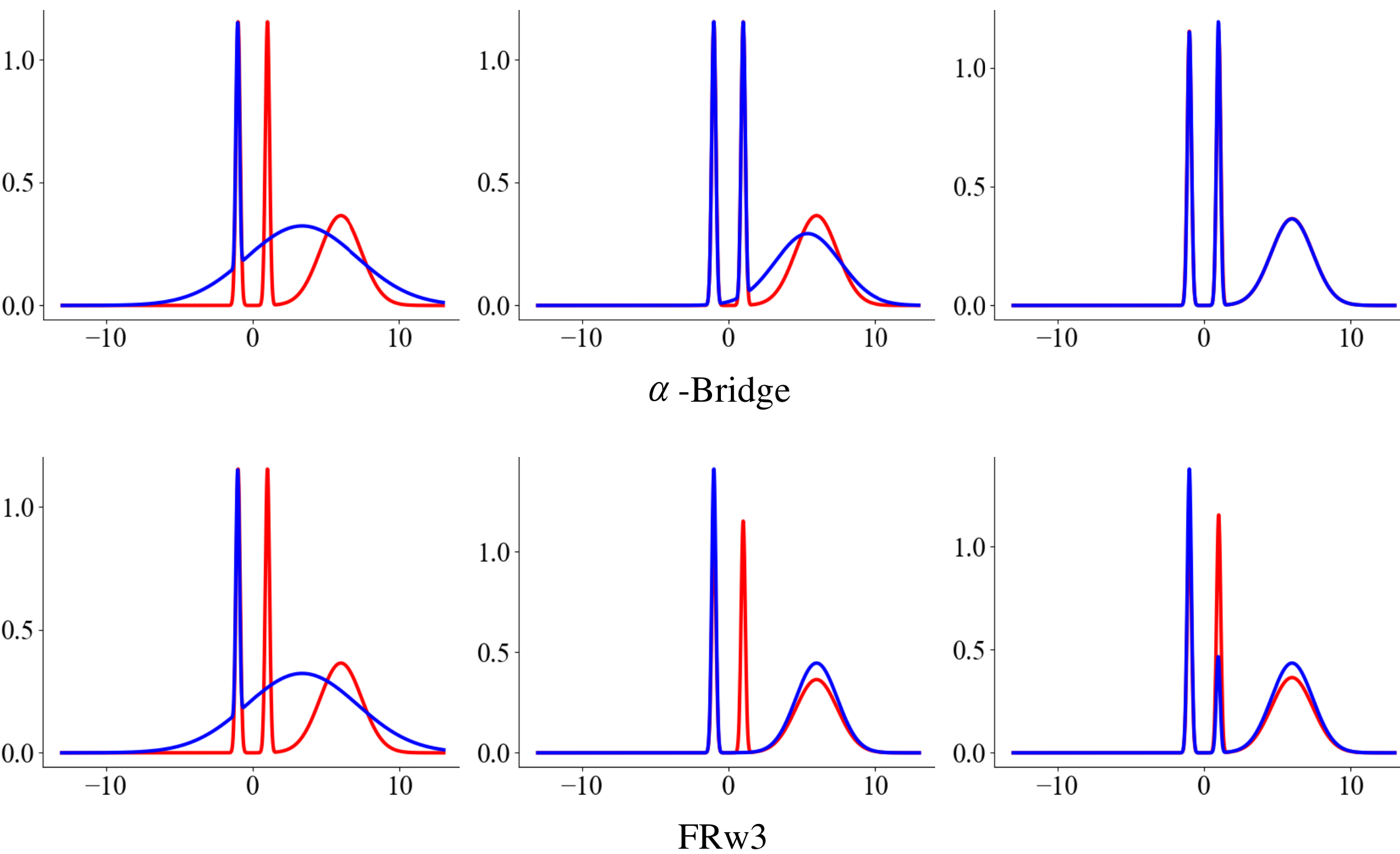}
		\caption{\small 
			Results on 1-D Gaussian mixture data. The snapshot of the distribution $p_{\theta}(\xv)$ (blue curve) during training process for $\alpha$-Bridge (the first row) and FRw3 (the second row). The ground truth distribution $q(\xv)$ is in red. Apparently, FRw3 has forget the middle mode during training, and randomly catch it by luck, while $\alpha$-Bridge doesn't forget the mode in hand and match them one by one. Because $\alpha$-Bridge has its object smoothly varying from FKL to RKL, the optimization process is more stable than the other method. 
		}
		\label{fig:vs_FKLpRKL_3c}
	\end{center}
\end{figure}

When $p_{\theta}(\xv)$ is expressive enough, $\alpha$-Bridge is comparable with the best of the other method.
Obviously, when the parameterized model has limited power, which is often the case in real application, $\alpha$-Bridge shows great advantages over the other combination methods. 
We assume the reason is that,  $\alpha$-Bridge has its object smoothly varying from FKL to RKL, thus the optimization process is more stable than the other method. Take FRw3 as an example, we show the snapshots of $p_{\theta}(\xv)$ during the optimization process in Figure \ref{fig:vs_FKLpRKL_3c}. Apparently, FRw3 has forget the middle mode during training, and randomly catch it by luck, while $\alpha$-Bridge doesn't forget the mode in hand and match them one by one.

\begin{table}[H]\centering
	\caption{counts of the trials that $p_{\theta}(\xv)$ matches $q_(\xv)$ for different combination methods.
	}\label{tab:vs_FKLpRKL_enough}
	\resizebox{0.95\hsize}{!}{
	\begin{tabular}{c|c|c|c|c}\hline\hline
		$q(\xv)$  	                                            &FR& FRw& FRw3&$\alpha$-Bridge \\ \hline\hline
		$\mu_{1},\mu_{2}, \mu_{3}= (6.0, -1.0, 1.0)$  & \multirow{2}{*}{9} & \multirow{2}{*}{4} & \multirow{2}{*}{5} & \multirow{2}{*}{10}     \\
		$\sigma_{1}, \sigma_{2}, \sigma_{3} =(1.0, 0.1, 0.1)$
		&     &                  &             &            \\ \hline   
		$\mu_{1},\mu_{2}, \mu_{3}=(6.0, 0.0, -2.0)$ & \multirow{2}{*}{8}  & \multirow{2}{*}{8} & \multirow{2}{*}{10}  &  \multirow{2}{*}{10} \\   
		$\sigma_{1}, \sigma_{2}, \sigma_{3} =(1.0, 0.1, 0.2)$   &      &                   &               &              \\ \hline  
		$\mu_{1},\mu_{2}, \mu_{3}= (5.0, -1.0 -3.0)$ & \multirow{2}{*}{4}  & \multirow{2}{*}{8} & \multirow{2}{*}{8}  &  \multirow{2}{*}{8} \\  
		$\sigma_{1}, \sigma_{2}, \sigma_{3} =(1.0, 0.2, 0.2)$   &      &                   &                &               \\ \hline   
		$\mu_{1},\mu_{2}, \mu_{3}=(6.0, -1.0, 1.0)$ & \multirow{2}{*}{0}  & \multirow{2}{*}{0} & \multirow{2}{*}{1}  &  \multirow{2}{*}{10} \\     
		$\sigma_{1}, \sigma_{2}, \sigma_{3} =(0.3, 0.3, 0.3)$   &      &                  &                &              \\ \hline 
	\end{tabular}
	}
\end{table}

%

\section{Experimental Settings}
\label{sec:app_exp}

The detailed settings for the experiments are given below.

\subsection{$25$-Gaussians}\label{sec:app_exp_25Gaussian}

The data are collected from a 2D Gaussian mixture model with $25$ components uniformly distributed on the $5 \times 5$ grid, as shown in Figure \ref{fig:Sample_25G} of the main manuscript.
The covariance matrix for each component is set to $0.0002\Imat_{2\times2}$. 
The centroids distance along the grid is set to $2$, such that the overall data distribution exhibits severely separated modes, making it a relatively hard task for GAN.

The latent code $\zv$ with a dimension $d_{\zv}=2$ is sampled from the standard Gaussian. 
The generator $G_{\thetav}(\zv)$ is constructed with {\tt Linear} layers; the corresponding network architecture is shown in Table \ref{tab:network_25G}. 
The discriminator shares a similar architecture, with the two differences of the last layer having an output size of $1$ and different {\tt BN} settings for different regularizations.
When using the spectral normalization (SN) \cite{miyato2018spectral} for regularization, we replace the {\tt BN} layers with {\tt SN} layers for the discriminator.
When using the gradient penalty (GP) \cite{mescheder2018training}, we remove the {\tt BN} layers. 
For the variational inference arm used by our $\alpha$-Bridge $q_{\phiv}(\zv|\xv)=\Nc(\muv_{\phiv}(\xv), \sigmav^2_{\phiv}(\xv)\Imat)$, we choose a similar network architecture as in Table \ref{tab:network_25G}, with the difference of replacing the last {\tt Linear} layer with two sub-networks (shown in Table \ref{tab:sub_q}) to model $\muv_{\phiv}(\xv)$ and $\log\sigmav_{\phiv}(\xv)$, respectively.

\begin{table}[H]
	\centering
	\caption{The network architecture used on $25$-Gaussians. \emph{lReLU} for leaky ReLU with slope $0.2$.
	}\label{tab:network_25G}
	\begin{tabular}{lcl}\hline\hline
		Layer               & kernel      & output        \\ \hline
		Linear, lReLU   & $-$         & 400          \\ \hline
		Linear, lReLU   & $-$         & 400          \\ \hline
		Linear, lReLU   & $-$         & 400          \\ \hline
		Linear, lReLU   & $-$         & 400          \\ \hline
		Linear              & $-$         & $2$ (generator) or          \\ 
		              &        &  $1$ (discriminator)          \\ \hline\hline
	\end{tabular}
\end{table}

\begin{table}[H]\centering
	\caption{The sub-network architecture to model $\muv_{\phiv}(\xv)$ and $\log\sigmav_{\phiv}(\xv)$ for the variational inference arm $q_{\phiv}(\zv|\xv)$ on $25$-Gaussians.}\label{tab:sub_q}
	\begin{tabular}{lcc}\hline\hline
		Layer               & kernel      & output        \\ \hline
		Linear              & $-$         &  $d_{\zv}=2$        \\ \hline\hline
	\end{tabular}
\end{table}

The training iterations for Steps $I$, $I\!I$, and $I\!I\!I$ of our $\alpha$-Bridge (see Algorithm \ref{alg:Alpha-divergence}) are set to $7.5$K, $7.5$K, and $25$K, respectively.
All the models are trained using Adam with the batch size $50$, the learning rate $lr=0.0002$, and $\beta_2=0.999$.

We use the inception score (IS) \cite{salimans2016improved} and the log-likelihood estimated with kernel density estimation \cite{parzen1962estimation} as quantitative metrics, to measure the plausibility of generated samples and the data-mode-covering level of learned models, respectively.
Following \cite{li2017alice}, we compute $\text{IS}=\Ebb_{p_{\thetav}(\xv)} \big[ \Dc_{\KL} [p(y)\|p(y|\xv)] \big]$ with the help of a classifier $p(y|\xv)$ pre-trained on $10,000$ data samples, where $p_{\thetav}(\xv)$ represents the generative model and $p(y)$ is the uniform prior of label $y$.
The classifier $p(y|\xv)$ is a $5$-layer neural networks with {\tt Linear} layers and {\tt lReLU} activation, pre-trained to yield $100\%$ classification accuracy on the training data.
Note this kind of IS evaluation is not related to the inception model trained on ImageNet. 
To estimate the log-likelihood, we use $5$K data samples and choose a Gaussian kernel with the same variance of the data, \ie $\sigma^2 = 0.0002$. The hyperparameter $\gamma$ of the gradient penalty (GP) is set to $\gamma=10$ by referring to the original paper \cite{mescheder2018training}. 
$10$ random trials are ran to get the curves shown in Figure \ref{fig:25Gs_IS_LL} of the main manuscript.

\subsection{Tuning Baseline Methods on $25$-Gaussians}
\label{sec:app_tuningbase_25Gs}

\begin{figure}[H]
	\begin{center}
		\subcaptionbox{RKL-GP \label{fig:}}{
			\includegraphics[height=0.35 \columnwidth]{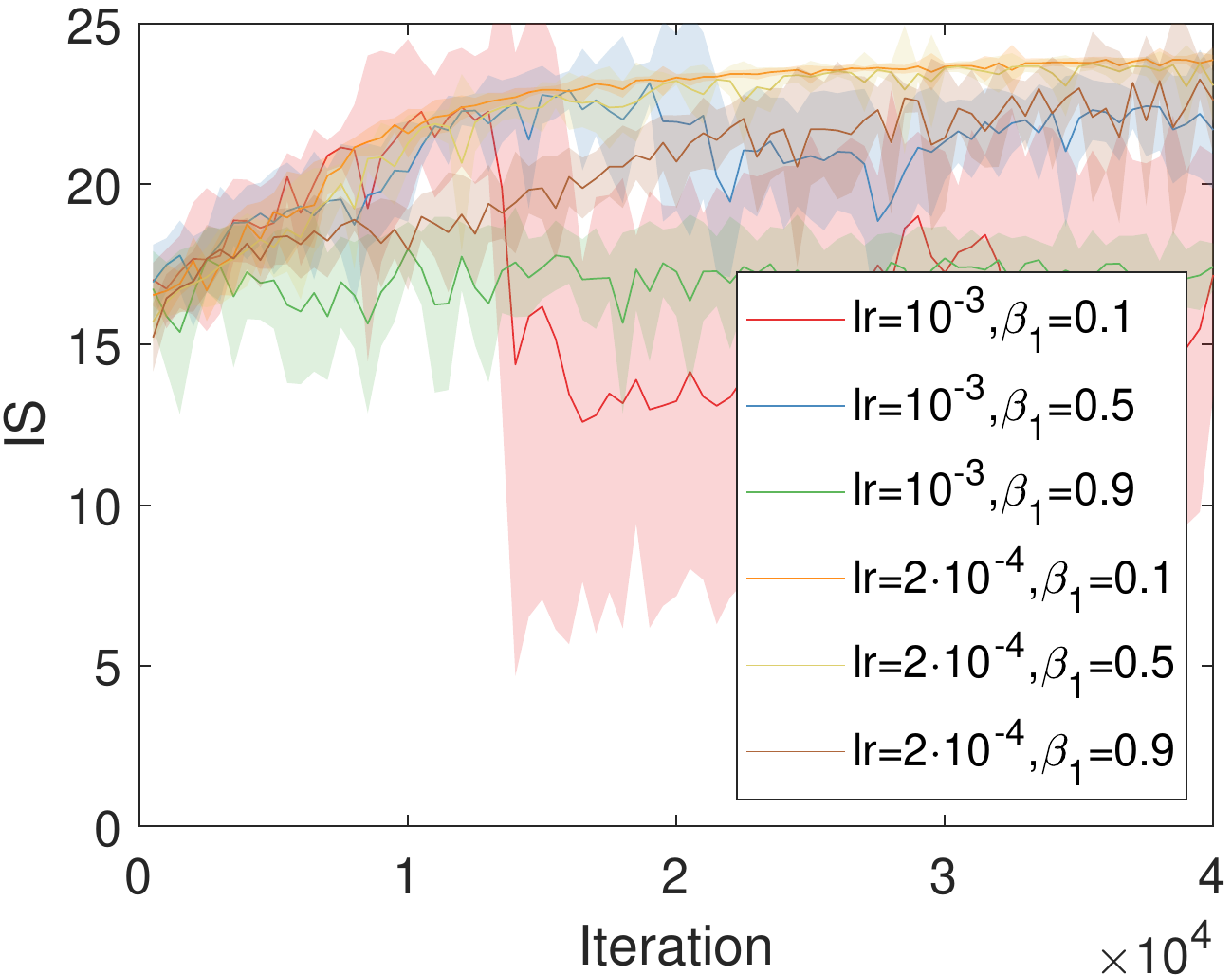}
		}
		\subcaptionbox{RKL-SN \label{fig:}}{
			\includegraphics[height=0.35 \columnwidth]{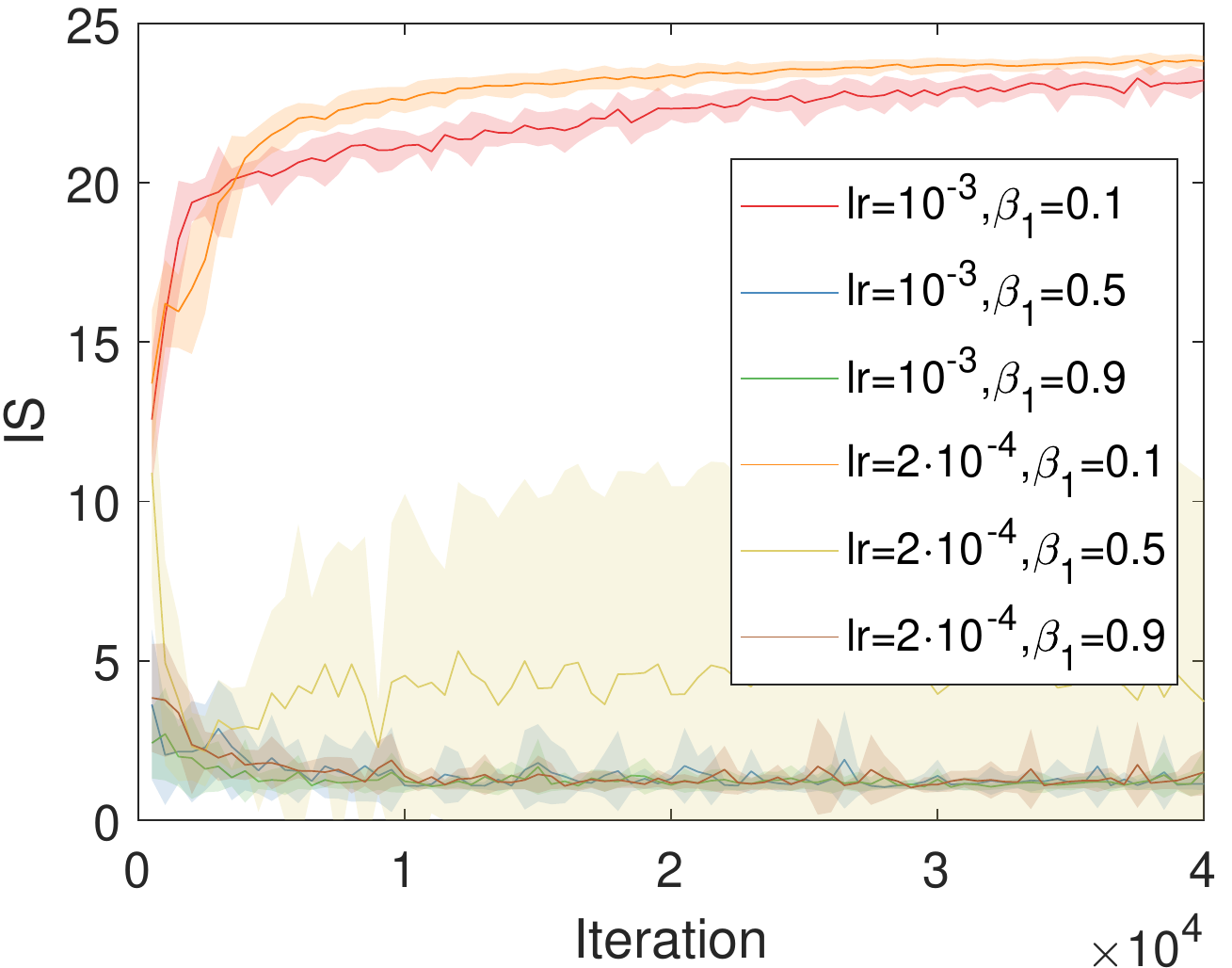}
		}
		\caption{\small Empirical evaluation of different hyperparameter settings for the baseline methods, \ie the reverse-KL-based adversarial learning regularized by GP (RKL-GP) (a) and that regularized by SN (RKL-SN) (b).
		}
		\label{fig:base_hypara}
	\end{center}
\end{figure}

To get competitive baselines, we test several hyperparameter settings and choose among them the best overall settings to build a fair comparison environment. The experimental results are given in Figure \ref{fig:base_hypara}. 
It seems the settings $lr=2\cdot10^{-4}, \beta_1=0.1$ work the best overall among the tested settings for the baseline methods. Accordingly, it is chosen.

\begin{figure}[H]
	\begin{center}
		\subcaptionbox{\label{fig:4compare_4robust_GP}}{
			\includegraphics[height=0.35 \columnwidth]{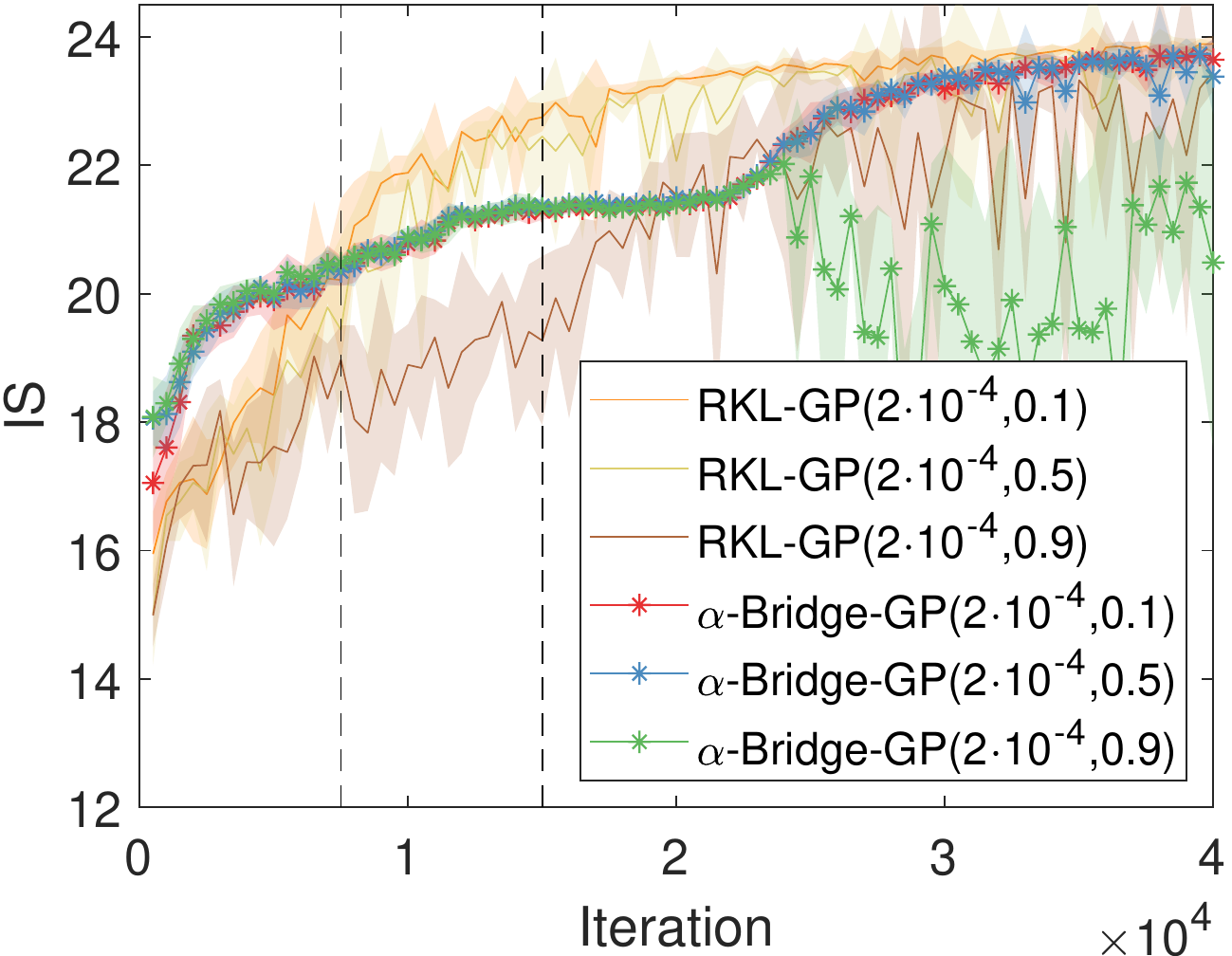}
		}
		\subcaptionbox{\label{fig:4compare_4robust_SN}}{
			\includegraphics[height=0.35 \columnwidth]{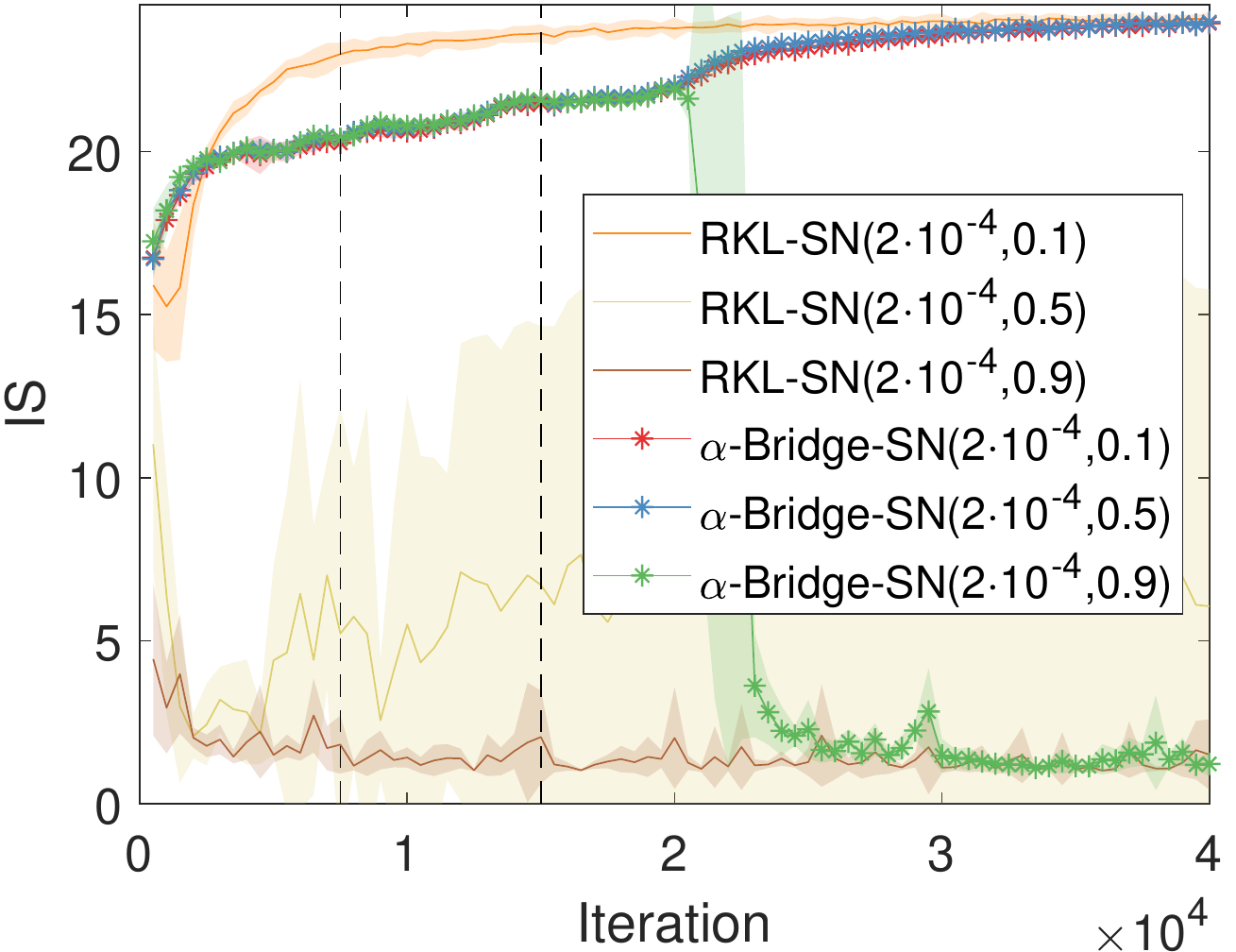}
		}
		\caption{\small Comparing the $\alpha$-Bridge methods with the baseline methods on different hyperparameter settings. The notation following methods denotes $(\textit{learning rate}, \beta_1)$. Two vertical dashed lines are used to indicate the three steps of the $\alpha$-Bridge.
		}
		\label{fig:4compare_4robust}
	\end{center}
\end{figure}

For parts of the hyperparameter settings, we compare our $\alpha$-Bridge methods to the baseline methods and present the corresponding results in Figure \ref{fig:4compare_4robust}. It is clear that our methods are more robust to the tested hyperparameter settings than the baseline methods, especially during the first two steps. Unfortunately for some situations where the baseline methods fail, the $\alpha$-Bridge methods may also fail in Step $I\!I\!I$ sometimes; after all, Step $I\!I\!I$ of the $\alpha$-Bridge \emph{is} the reverse-KL-based adversarial learning. But we do observe the $\alpha$-Bridge works when the corresponding baseline fails, such as for the settings $(2\cdot10^{-4}, 0.5)$ in Figure \ref{fig:4compare_4robust}.

\subsection{MNIST}\label{sec:app_exp_mnist}

For the experiments on MNIST, we use the DCGAN \cite{radford2015unsupervised} architectures, as shown in Tables \ref{tab:network_MNIST_G} and \ref{tab:network_MNIST_D}, for the generator and discriminator, respectively.
The variational inference arm $q_{\phiv}(\zv|\xv)=\Nc(\muv_{\phiv}(\xv), \sigmav^2_{\phiv}(\xv)\Imat)$ has a similar network architecture as that of the discriminator, with the difference of replacing the last {\tt Linear} layer with two shallow sub-networks to model $\muv_{\phiv}(\xv)$ and $\log\sigmav_{\phiv}(\xv)$, respectively.
The prior of the latent code $\zv$ is the standard Gaussian. $\zv$ has a dimension $d_{\zv}=64$. 
We train each step (see Algorithm \ref{alg:Alpha-divergence}) of our $\alpha$-Bridge for $5, 5, 40$ epochs, respectively, on the MNIST training data.   
The batch size is set to $100$.
We use GP to regularize the discriminator in this experiment. 
The GP hyperparameter $\gamma$ is set to $\gamma=10$.

\begin{table}[H]\centering
	\caption{The generator architecture on MNIST. 
		\emph{BN} stands for batch normalization \cite{ioffe2015batch}. 
		The kernel size is described in format
		$h_{filter} \times w_{filter} \times stride$. The output shape is $channels \times h \times w$.}\label{tab:network_MNIST_G}
	\begin{tabular}{lcl} \hline\hline
		Layer              & Kernel             & Output \\ \hline
		Linear, BN, lReLU  & $-$                & $256\times4\times4$  \\ \hline
		Deconv, BN, lReLU  & $4\times4\times2$  & $128\times7\times7$  \\ \hline
		Deconv, BN, lReLU  & $5\times5\times2$  & $64\times14\times14$ \\ \hline
		Deconv, BN, lReLU  & $5\times5\times2$  & $32\times28\times28$ \\ \hline
		Deconv, Tanh       & $5\times5\times1$  & $1\times28\times28$ \\ \hline\hline
	\end{tabular}
\end{table}

\begin{table}[H]\centering
	\caption{The discriminator architecture on MNIST.}\label{tab:network_MNIST_D}
	\begin{tabular}{lcl}\hline\hline
		Layer              & Kernel               & Output               \\ \hline
		Conv, lReLU        & $3\times3\times2$ & $32\times14\times14$ \\ \hline
		Conv, lReLU & $3\times3\times2$ & $64\times7\times7$ \\ \hline
		Conv, lReLU & $3\times3\times2$ & $128\times4\times4$ \\ \hline
		Linear      & $-$               & $1$\\ \hline\hline
	\end{tabular}
\end{table}

\begin{table}[H]\centering
	\caption{The sub-network architecture to model $\muv_{\phiv}(\xv)$ and $\log \sigmav_{\phiv}(\xv)$ for the variational inference arm $q_{\phiv}(\zv|\xv)$ on MNIST.}\label{tab:sub_q_mnist}
	\begin{tabular}{lcc}\hline\hline
		Layer               & kernel      & output        \\ \hline
		Linear, lReLU       & $-$         & 1000          \\ \hline
		Linear              & $-$         &  $d_{\zv}=64$        \\ \hline\hline
	\end{tabular}
\end{table}


\subsection{CIFAR$10$}\label{sec:app_exp_cifar10}

\begin{table}[H]\centering
	\caption{The generator architecture on CIFAR$10$.}\label{tab:network_CIFAR10_G}
	\begin{tabular}{lcl}\hline\hline
		Layer             & Kernel             & Output                  \\ \hline
		Linear, BN, lReLU & $-$                & $128\times4  \times 4$ \\ \hline
		Deconv, BN, lReLU & $4\times4\times2$  & $256 \times8  \times 8$ \\ \hline
		Deconv, BN, lReLU & $4\times4\times2$  & $128 \times16 \times16$ \\ \hline
		Deconv, BN, lReLU & $4\times4\times2$  & $64   \times32 \times32$ \\ \hline
		Deconv, Tanh      & $3\times3\times1$  & $3   \times32 \times32$ \\ \hline\hline
	\end{tabular}
\end{table}
\begin{table}[H]\centering
	\caption{The discriminator architecture on CIFAR$10$. }\label{tab:network_CIFAR10_D}
	\begin{tabular}{lcl} \hline\hline
		Layer           & Kernel            & Output                 \\ \hline
		Conv, BN, lReLU & $3\times3\times1$ & $32\times32\times32$  \\ \hline
		Conv, BN, lReLU & $4\times4\times2$ & $64\times16\times16$    \\ \hline
		Conv, BN, lReLU & $4\times4\times2$ & $128\times8\times8$    \\ \hline
		Conv, BN, lReLU & $4\times4\times2$ & $256\times4\times4$    \\ \hline
		Conv, BN, lReLU & $4\times4\times2$ & $512\times2\times2$  \\ \hline
		Linear          & $-$               & $1$                    \\ \hline\hline
	\end{tabular}
\end{table}
\begin{table}[H]\centering
	\caption{The sub-network architecture to model $\muv_{\phiv}(\xv)$ and $\log\sigmav_{\phiv}(\xv)$ on CIFAR$10$.}\label{tab:sub_q_cifar}
	\begin{tabular}{lcc}\hline\hline
		Layer               & kernel      & output        \\ \hline
		Linear, BN, lReLU   & $-$         & 512          \\ \hline
		Linear              & $-$         & $d_{\zv}=128$        \\ \hline\hline
	\end{tabular}
\end{table}

To address the concern of practical benefits of the proposed $\alpha$-Bridge, we conduct another experiment on CIFAR$10$ \cite{krizhevsky2009learning} for quantitative evaluations on real datasets.
We implement the experiment based on the codebase from \url{https://github.com/pfnet-research/chainer-gan-lib}.
For the vanilla DCGAN baseline, we used the implementation therein to report the results shown in the main manuscript (for fair comparison, we replace the original GAN loss with the reverse KL loss). The used network architectures are shown in Tables \ref{tab:network_CIFAR10_G}-\ref{tab:sub_q_cifar}, respectively.
For our $\alpha$-Bridge, we kept the network architectures and hyperparameters unchanged, only to add another inference arm and modify the loss function following Algorithm \ref{alg:Alpha-divergence} of the main manuscript. The inference arm is constructed similar to the discriminator, with the only difference of replacing the last layer with two sub-networks (see Table \ref{tab:sub_q_cifar}) to output the mean $\muv_{\phiv}(\xv)$ and the log of the standard deviation $\log\sigmav_{\phiv}(\xv)$ of the inference arm, respectively. Roughly speaking, the difference between the baseline method and the $\alpha$-Bridge one is that the baseline method only runs Step $I\!I\!I$ of Algorithm \ref{alg:Alpha-divergence}, while the $\alpha$-Bridge method runs the whole Algorithm \ref{alg:Alpha-divergence}.

\begin{figure}[H]
	\begin{center}
		\subcaptionbox{\label{fig:CIFAR10_IS}}{
			\includegraphics[height=0.35 \columnwidth]{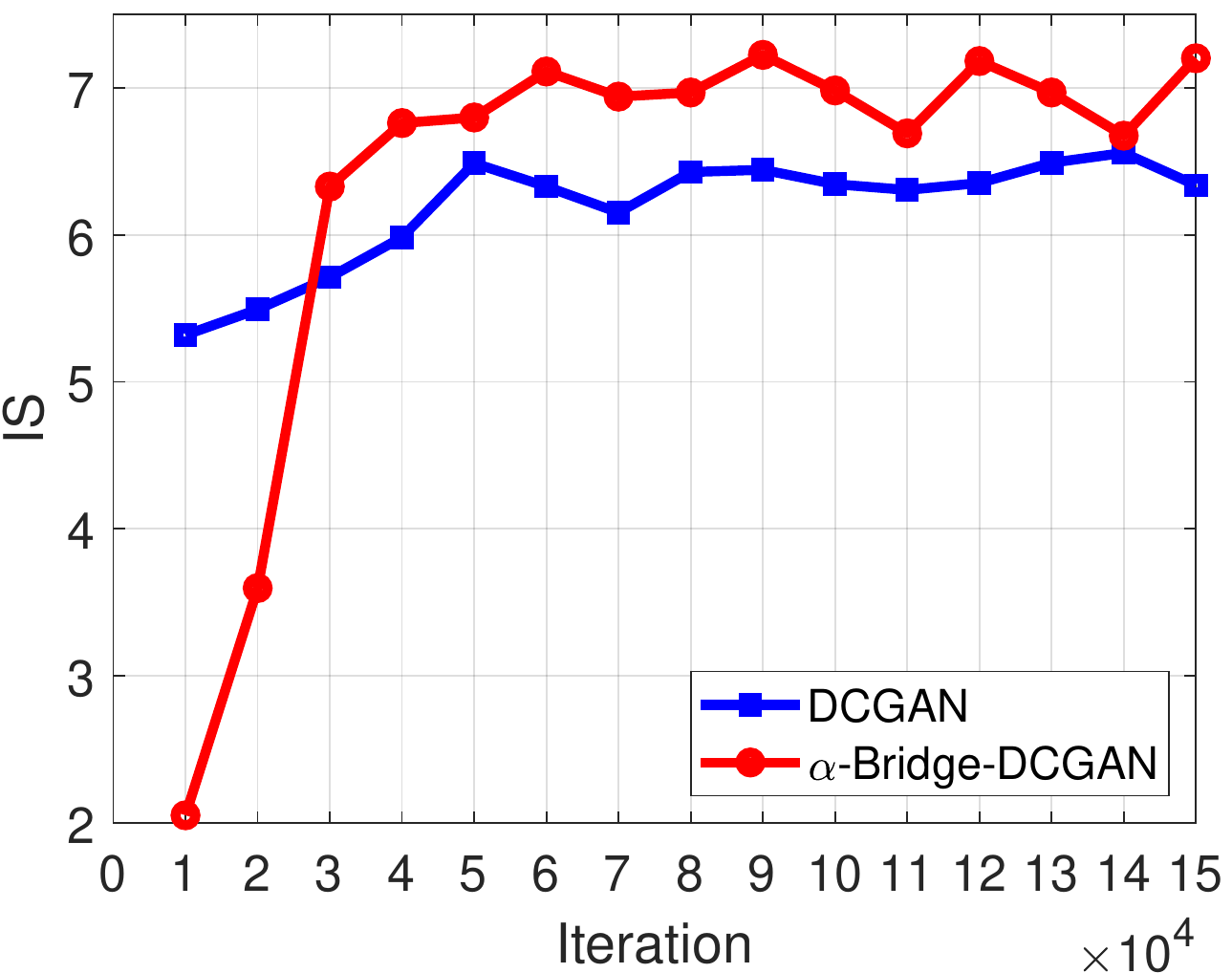}
		}
		\subcaptionbox{\label{fig:CIFAR10_FID}}{
			\includegraphics[height=0.35 \columnwidth]{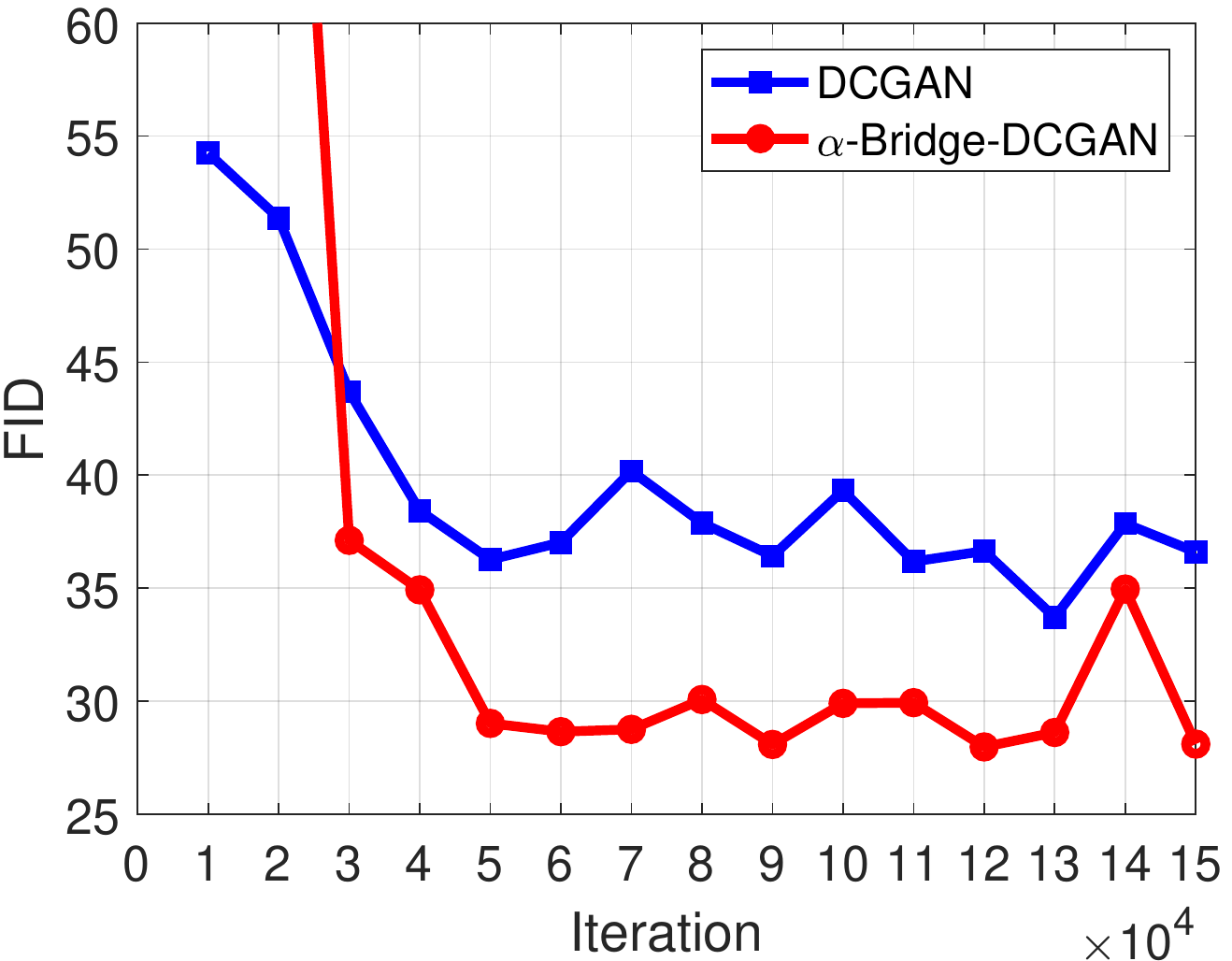}
		}
		\caption{\small Comparison between the $\alpha$-Bridge with the corresponding DCGAN baseline. Steps $I$, $I\!I$, and $I\!I\!I$ of the $\alpha$-Bridge contains 10K, 10K, and 130K iterations, respectively. Higher is better for the IS, while lower is better for the FID.
		}
		\label{fig:CIFAR10_IS_FID}
	\end{center}
\end{figure}

\begin{figure}[H]
	\begin{center}
		\subcaptionbox{DCGAN \label{fig:}}{
			\includegraphics[height=0.35 \columnwidth]{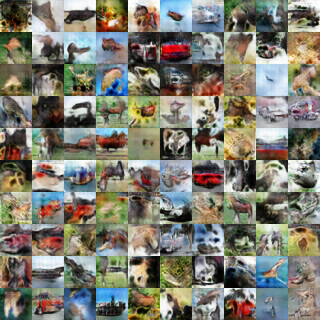}
		}
		\subcaptionbox{$\alpha$-Bridge-DCGAN \label{fig:CIFAR10_gen_sample_beginning_alpha}}{
			\includegraphics[height=0.35 \columnwidth]{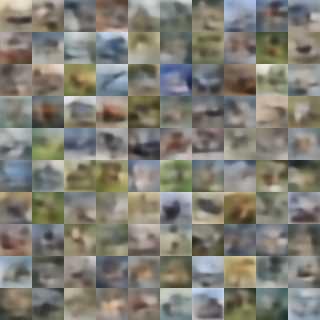}
		}
		\caption{\small Generated samples at $10$K iterations from DCGAN (a) and $\alpha$-Bridge-DCGAN (b). The $\alpha$-Bridge-DCGAN focuses on covering data modes in its Step $I$, leading to worse IS and FID in this stage.
		}
		\label{fig:CIFAR10_gen_sample_beginning}
	\end{center}
\end{figure}

Figure \ref{fig:CIFAR10_IS_FID} shows the IS and FID curves of the compared methods as a function of training iterations. It is apparent that with the $\alpha$-Bridge to benefit from ML learning, one observes improved performance both in IS and FID. Note in the beginning, since the $\alpha$-Bridge is running Step $I$ of Algorithm \ref{alg:Alpha-divergence} (the ML initialization), it focuses on the general information to cover the data modes as shown in Figure \ref{fig:CIFAR10_gen_sample_beginning_alpha}; accordingly, a worse IS and FID are observed. This is expected since after all both IS and FID are proposed for GANs to measure the image quality instead the data-mode-covering-level. However, we do observe that the $\alpha$-Bridge does be able to benefit from the mode-covering of ML learning to get a better final performance. 

More comparison results with other GAN models on CIFAR10 based on the chainer-gan-lib codebase are summarized in Table \ref{tab:fid_is_cifar}, which highlights the priority of the presented $\alpha$-Bridge. Note for fair comparisons, we use the exact experimental settings from that codebase, without adding GP or SN. 
We attribute the success of our $\alpha$-Bridge to two main reasons. ($i$) Among the three advantages that $\alpha$-Bridge transfers from MLE to adversarial learning (see Table 1), the mode covering property is expected to mitigate the mode dropping of adversarial learning, and accordingly benefits a better final performance, like a better IS/FID. ($ii$) Transferring from MLE could efficiently initialize a better manifold for adversarial learning to start with. 

\begin{table}[H] \centering
	\caption{The FID and IS of compared methods on CIFAR10 based on the chainer-gan-lib codebase \url{https://github.com/pfnet-research/chainer-gan-lib}.}\label{tab:fid_is_cifar}
	\begin{tabular}{ccc}\hline\hline
		Methods         & FID    & IS  \\ \hline
		BEGAN           & $84.0$ & $5.4$ \\ 
		DRAGAN          & $31.5$ & $7.1$ \\ 
		WGAN-GP         & $28.2$ & $6.8$ \\ 
		RKL-DCGAN       & $33.7$ & $6.6$ \\ \hline
		$\alpha$-Bridge & $\textbf{28.1}$ & $\textbf{7.2}$ \\ \hline\hline
	\end{tabular}
\end{table}

\subsection{CelebA}\label{sec:app_exp_celeba}

\begin{table}[H] \centering
	\caption{The generator architecture on CelebA with image size $64 \times 64$.}\label{tab:network_CelebA_G_64}
	\begin{tabular}{lcl}\hline\hline
		Layer              & Kernel           & Output                \\ \hline
		Linear, BN, lReLU  & $-$              & $1024\times4\times4$  \\ \hline
		Deconv, BN, lReLU & $4\times4\times2$ & $512\times8\times8$ \\ \hline
		Deconv, BN, lReLU & $4\times4\times2$ & $256\times16\times16$ \\ \hline
		Deconv, BN, lReLU & $4\times4\times2$ & $128\times32\times32$ \\ \hline
		Deconv, BN, lReLU & $4\times4\times2$ & $64\times64\times64$  \\ \hline
		Deconv, Tanh      & $4\times4\times2$ & $3\times64\times64$ \\ \hline\hline
	\end{tabular}
\end{table}
\begin{table}[H]\centering
	\caption{The discriminator architecture on CelebA with image size $64 \times 64$.}\label{tab:network_CelebA_D_64}
	\begin{tabular}{lcl}\hline\hline
		Layer              & Kernel           & Output           \\ \hline
		Conv, SN, lReLU  & $3\times3\times1$ & $64\times64\times64$  \\ \hline
		Conv, SN, lReLU  & $3\times3\times2$ & $128\times32\times32$ \\ \hline
		Conv, SN, lReLU  & $3\times3\times2$ & $256\times16\times16$ \\ \hline
		Conv, SN, lReLU  & $3\times3\times2$ & $512\times8\times8$ \\ \hline
		Conv, SN, lReLU  & $3\times3\times2$ & $1024\times4\times4$  \\ \hline
		Linear, SN   & $-$               & $1$                   \\ \hline\hline
	\end{tabular}
\end{table}

\begin{figure}[H]
	\begin{center}
		\includegraphics[width=0.95\columnwidth]{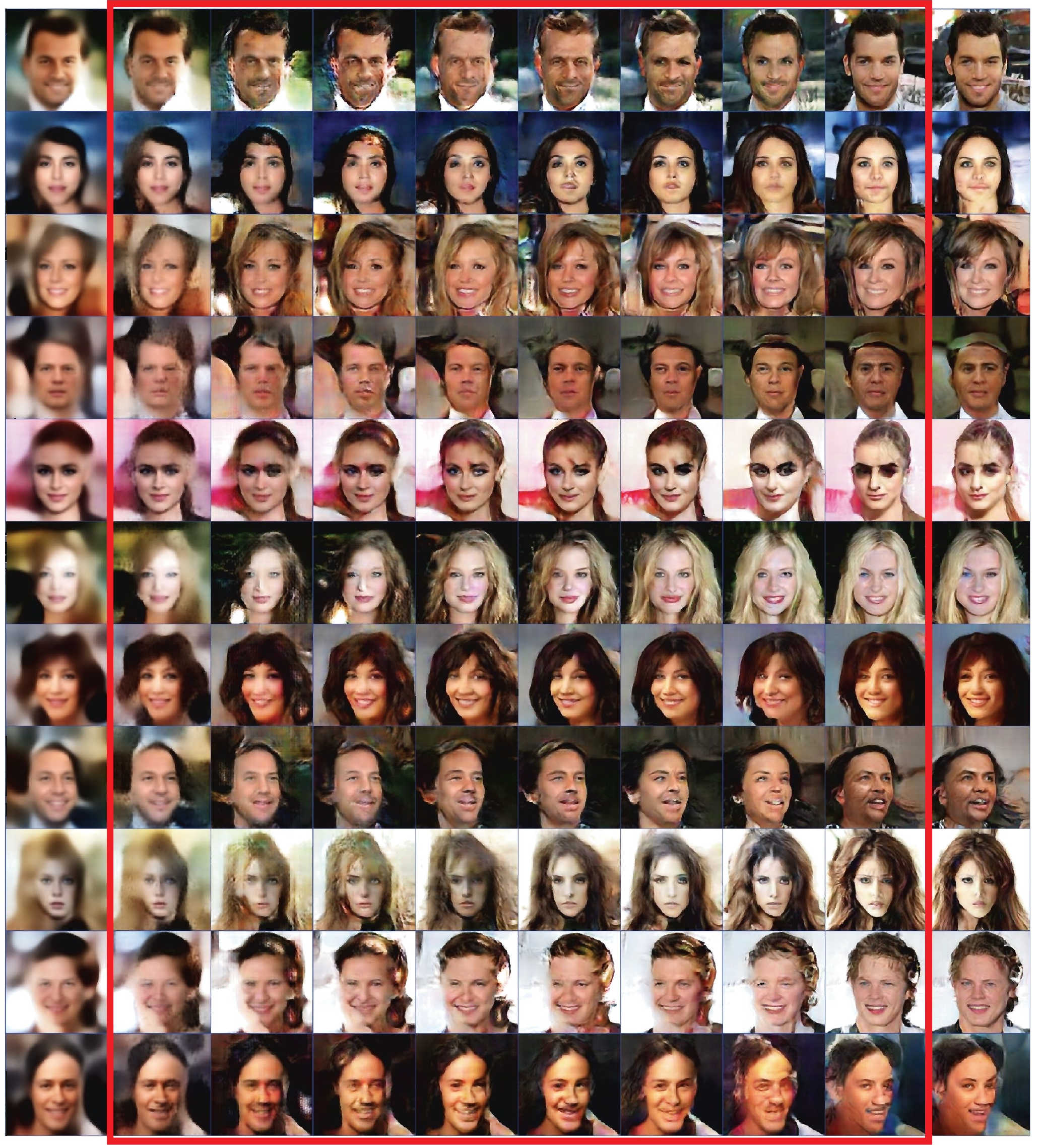}
		\caption{\small Random samples generated by the proposed $\alpha$-bridge on CelebA along the training process of Algorithm \ref{alg:Alpha-divergence} of the main manuscript. Left (Step $I$): maximum likelihood learning for efficient initialization. 
			Middle (inside the red box, Step $I\!I$): transferring from maximum likelihood to adversarial learning. Right (Step $I\!I\!I$): adversarial learning for refinement. 
		}
		\label{fig:app_sample_celeb_big}
	\end{center}
\end{figure}

For the experiments on CelebA, we use the DCGAN \cite{radford2015unsupervised} architectures.
For the reconstruction and manipulation experiments shown in Figure \ref{fig:attribute_inference_arm} of the main manuscript, we use the data images of size $64 \times 64$; accordingly, the network architectures shown in Tables \ref{tab:network_CelebA_G_64}-\ref{tab:network_CelebA_D_64} are used. 
To demonstrate the transfer process more clearly \ie Figure \ref{fig:Sample_MNIST_CelebA}, we used the CelebA images of size $160 \times 160$ and the architectures in Tables \ref{tab:network_CelebA_G_160}-\ref{tab:network_CelebA_D_160}.
The variational inference arm $q_{\phiv}(\zv|\xv)=\Nc(\muv_{\phiv}(\xv), \sigmav^2_{\phiv}(\xv)\Imat)$ has a similar network architecture as that of the discriminator, with the differences of replacing the last {\tt Linear} layer with two shallow sub-networks to model $\muv_{\phiv}(\xv)$ and $\log\sigmav_{\phiv}(\xv)$, respectively.
The prior of the latent code $\zv$ is the standard Gaussian. $\zv$ has a dimension $d_{\zv}=128$. 
We train each step (see Algorithm \ref{alg:Alpha-divergence}) of our $\alpha$-Bridge for $5$, $5$, and $40$ epochs on CelebA.
The batch size is set to $64$ due to the computational budget.
SN is used to regularize the discriminator in this experiment. More examples on the transfer process of our method are shown in Figure \ref{fig:app_sample_celeb_big}.

\begin{table}[H] \centering
	\caption{The generator architecture on CelebA with image size $160 \times 160$.}\label{tab:network_CelebA_G_160}
	\begin{tabular}{lcl}\hline\hline
		Layer              & Kernel           & Output                \\ \hline
		Linear, BN, lReLU  & $-$              & $1024\times5\times5$  \\ \hline
		Deconv, BN, lReLU & $5\times5\times2$ & $512\times10\times10$ \\ \hline
		Deconv, BN, lReLU & $5\times5\times2$ & $256\times20\times20$ \\ \hline
		Deconv, BN, lReLU & $5\times5\times2$ & $128\times40\times40$ \\ \hline
		Deconv, BN, lReLU & $5\times5\times2$ & $64\times80\times80$  \\ \hline
		Deconv, Tanh      & $5\times5\times2$ & $3\times160\times160$ \\ \hline\hline
	\end{tabular}
\end{table}
\begin{table}[H]\centering
	\caption{The discriminator architecture on CelebA with image size $160 \times 160$.}\label{tab:network_CelebA_D_160}
	\begin{tabular}{lcl}\hline\hline
		Layer              & Kernel           & Output           \\ \hline
		Conv, SN, lReLU  & $3\times3\times2$ & $64\times80\times80$  \\ \hline
		Conv, SN, lReLU  & $3\times3\times2$ & $128\times40\times40$ \\ \hline
		Conv, SN, lReLU  & $3\times3\times2$ & $256\times20\times20$ \\ \hline
		Conv, SN, lReLU  & $3\times3\times2$ & $512\times10\times10$ \\ \hline
		Conv, SN, lReLU  & $3\times3\times2$ & $1024\times5\times5$  \\ \hline
		Linear       & $-$               & $1$                   \\ \hline\hline
	\end{tabular}
\end{table}

\subsection{Transplant an Inference Arm for Adversarial Learning}
\label{sec:app_TransInferArm}

We first follow Steps $I$ and $I\!I$ of Algorithm \ref{alg:Alpha-divergence} of the main manuscript to train the generative process 
\beq\label{eq:GAN_generator}
p_{\thetav}(\xv): \xv \sim \delta(\xv | G_{\thetav}(\zv)), \zv \sim p(\zv)
\eeq
and the variational inference arm 
$$
q_{\phiv}(\zv|\xv)=\Nc(\muv_{\phiv}(\xv), \sigmav^2_{\phiv}(\xv)\Imat).
$$
When Step $I\!I\!I$ begins, we train the parameters $\phiv$ along side with the generator and the discriminator. Note instead of maximizing the ELBO in \eqref{eq:elbo_ML} of the main manuscript for training $\phiv$, we solve 
\beq\label{eq:}
\bali
\phiv^{*} = \argmin_{\phiv}  
\Ebb_{q(\xv) \delta(\zv|\muv_{\phiv}(\xv))} \big[
\| \xv - G_{\thetav}(\zv) \|_2^2
\big],
\eali
\eeq
because
\begin{enumerate}
	\item In theory, the posterior $p_{\thetav}(\zv|\xv)$ of the generative process in \eqref{eq:GAN_generator} is a Dirac delta function for common situations where $\xv$ has a higher dimension than that of $\zv$; thus for $q_{\phiv}(\zv|\xv)$ to be identical to $p_{\thetav}(\zv|\xv)$, $\sigmav^2_{\phiv}(\xv)$ must be zero, leading to $q_{\phiv}(\zv|\xv) = \delta(\zv|\muv_{\phiv}(\xv))$;
	
	\item The first reason can be alternatively understood from the perspective of variational inference. To maximize the ELBO \wrt \phiv 
	\beq\label{eq:}
	\bali
	& \text{ELBO}(\phiv) 
	\\
	& = \Ebb_{q(\xv) q_{\phiv}(\zv|\xv)} \left[
	\log p_{\thetav}(\xv | \zv) + \log \frac{p(\zv)}{q_{\phiv}(\zv|\xv)}
	\right]
	\\
	& = \Ebb_{q(\xv) q_{\phiv}(\zv|\xv)} \left[
	\log \lim_{\sigma^2 \to 0} \Nc(\xv | G_{\thetav}(\zv), \sigma^2\Imat ) + \log \frac{p(\zv)}{q_{\phiv}(\zv|\xv)}
	\right]
	\\
	& = \lim_{\sigma^2 \to 0} \Ebb_{q(\xv) q_{\phiv}(\zv|\xv)} \left[
	-\frac{\|\xv - G_{\thetav}(\zv)\|_2^2}{2\sigma^2} + \log \frac{p(\zv)}{q_{\phiv}(\zv|\xv)}
	\right]
	\eali
	\eeq
	for the model in \eqref{eq:GAN_generator} is equivalent to $
	\min_{\phiv} \Ebb_{q(\xv) q_{\phiv}(\zv|\xv)} \left[
	\|\xv - G_{\thetav}(\zv)\|_2^2.
	\right]$
	For situations where $\xv$ has a higher dimension than that of $\zv$, one would expect one and only one $\zv$ for each $\xv$, satisfying $\xv = G_{\thetav}(\zv)$. Accordingly, the optimal $q_{\phiv}(\zv|\xv)$ should be a deterministic function with zero variance $\sigmav^2_{\phiv}(\xv)=0$, \ie a Dirac delta function $q_{\phiv}(\zv|\xv) = \delta(\zv|\muv_{\phiv}(\xv))$.

	
\end{enumerate}

\begin{figure}[H]
	\begin{center}
		\subcaptionbox{fake $\to$ reconstruction \label{fig:}}{
			\includegraphics[width=0.9 \columnwidth]{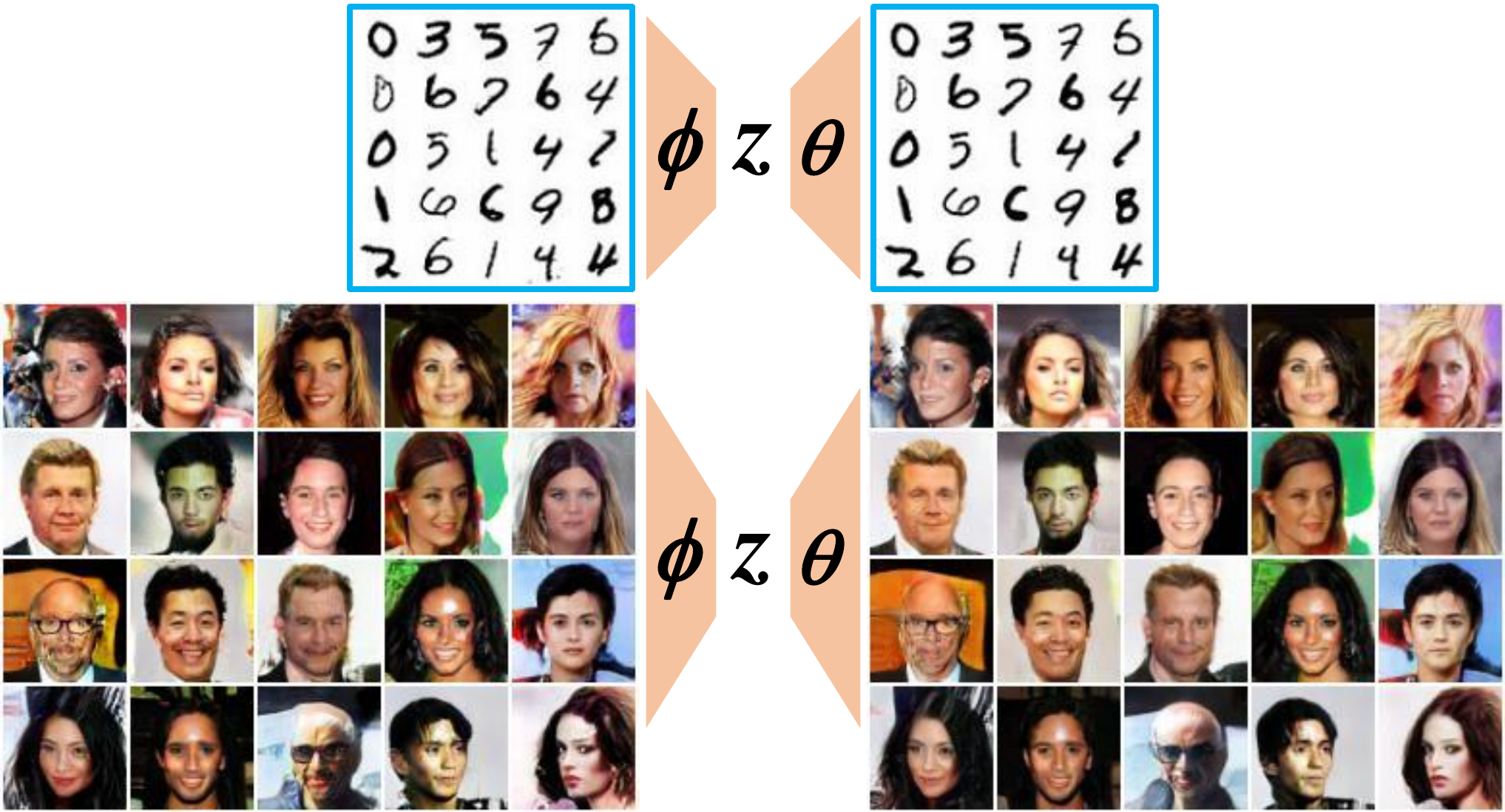}
		}
		\subcaptionbox{real $\to$ reconstruction \label{fig:}}{
			\includegraphics[width=0.9 \columnwidth]{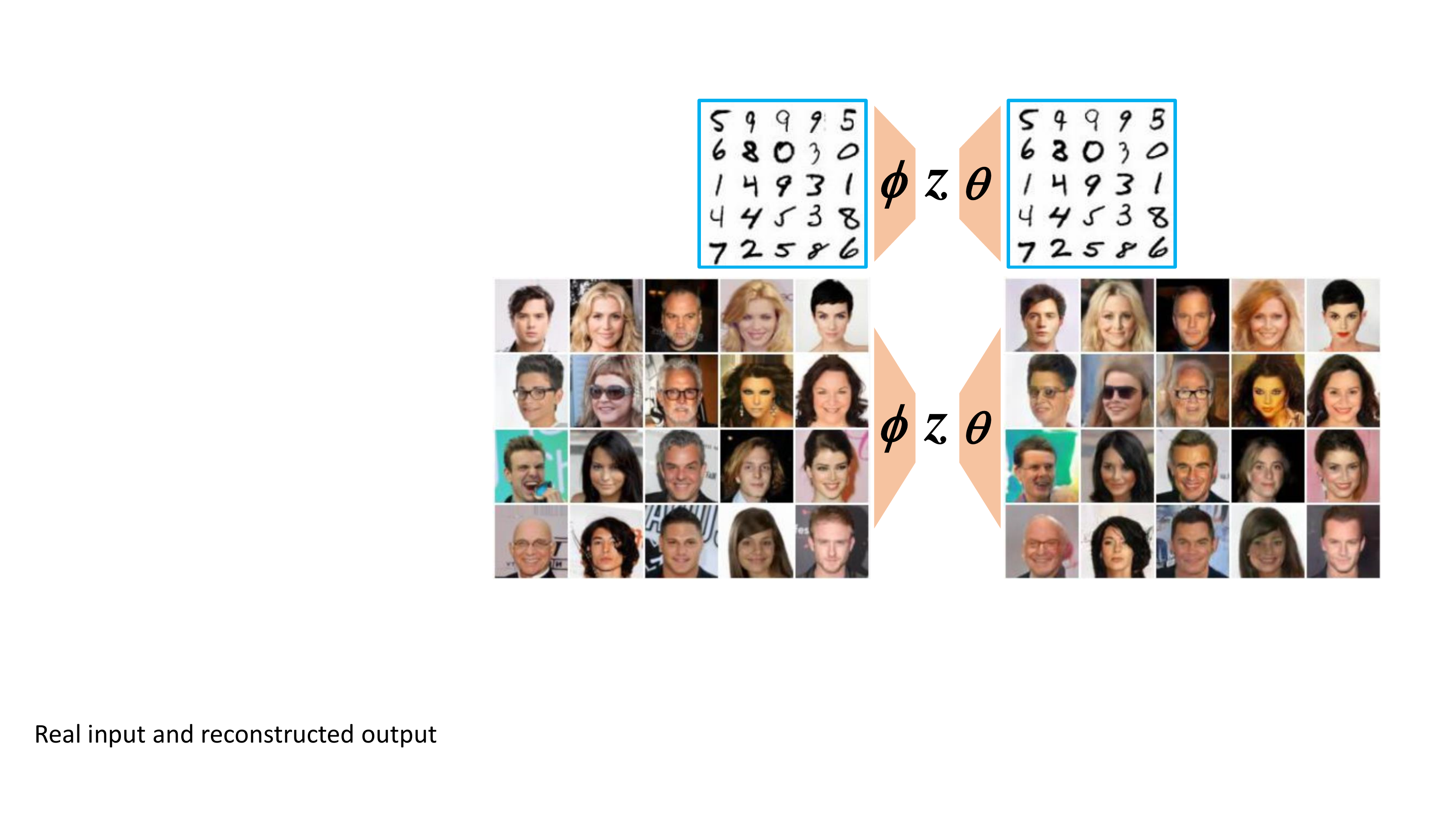}
		}
		\caption{\small Demonstration on the inference arm transplanted with $\alpha$-Bridge. $\phiv$ and $\thetav$ denote the inference arm $q_{\phiv}(\zv|\xv)$ (or $\muv_{\phiv}(\xv)$) and the generator $G_{\thetav}(\zv)$, respectively.  
			(a) Generated fake images and the reconstructions. 
			(b) Real data images and the reconstructions. 
		}
		\label{fig:recon_summary}
	\end{center}
\end{figure}

\begin{figure}[H]
	\begin{center}
		\includegraphics[width=1.0 \columnwidth]{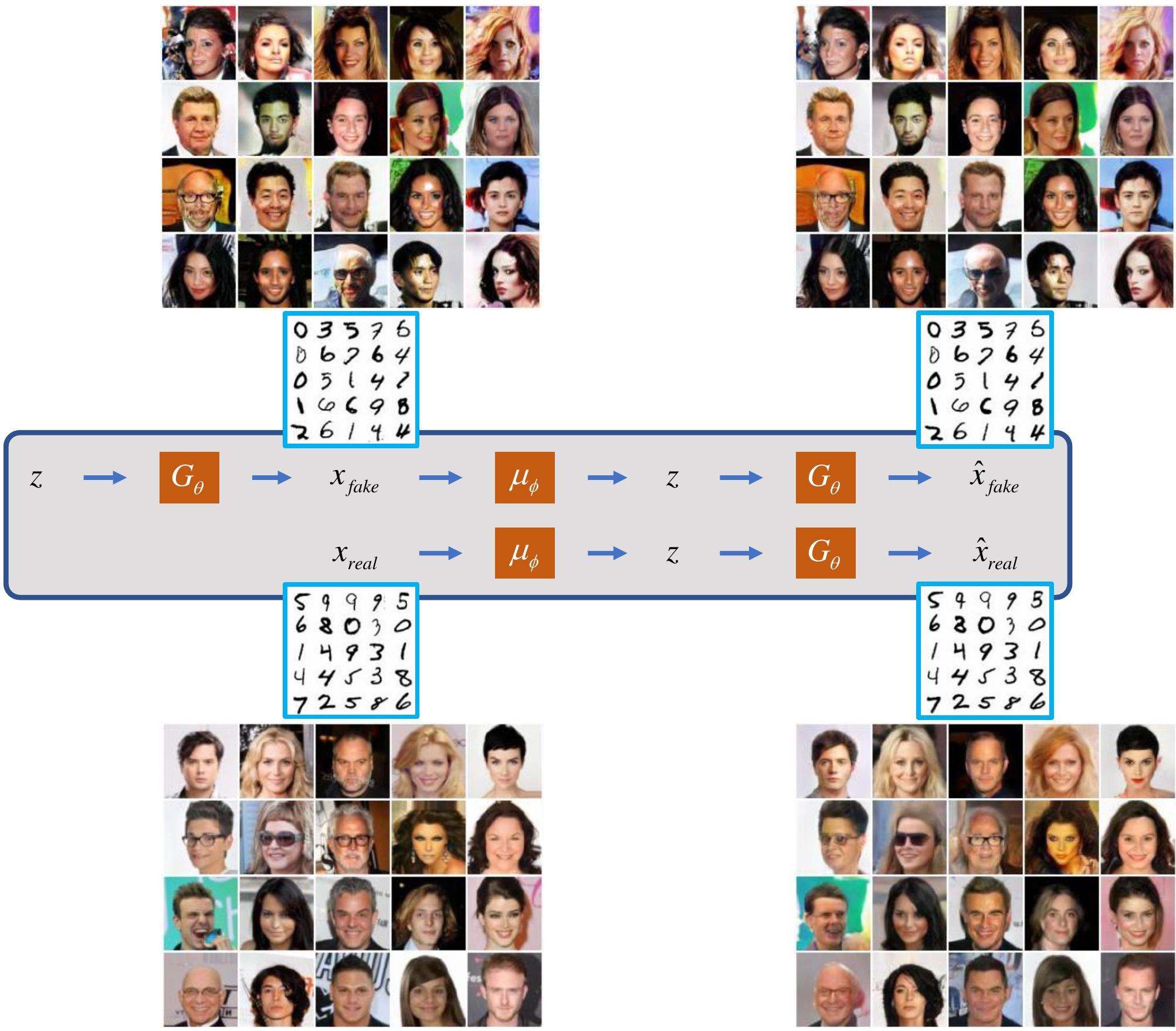}
		\caption{\small Demonstration of using the inference arm transplanted with $\alpha$-Bridge to reconstruct the generated images $\xv_{fake}$ and the data samples $\xv_{real}$. 
		}
		\label{fig:recon_process}
	\end{center}
\end{figure}

\textbf{How we got the reconstruction part of Figure \ref{fig:attribute_inference_arm}?}
Please see Figures \ref{fig:recon_summary} and \ref{fig:recon_process}.

Figures \ref{fig:recon_summary} and \ref{fig:recon_process} clearly demonstrate the effectiveness of the transplanted inference arm. For the generated fake sample, the inference arm is capable of recovering the latent code that almost regenerates the fake sample itself. While for the real data sample, one observes a decent regeneration with high mutual information.

\textbf{How we got the manipulation part of Figure \ref{fig:attribute_inference_arm}?}
By referring to a similar procedure proposed in \cite{larsen2015autoencoding}, we manipulate the generated images via the following two steps:
\begin{enumerate}
	\item \textbf{Extract visual attribute vectors.} We first use the transplanted inference arm $\muv_{\phiv}(\xv)$ (or $q_{\phiv}(\zv|\xv) = \delta(\zv|\muv_{\phiv}(\xv))$) to encode all the training data samples to their latent codes. Then with  the binary attributes of the dataset at hand, for each attribute, we collect the latent codes of the data samples with that attribute and compute their mean; similarly we compute the mean of those without that attribute; the difference between these two means is the visual attribute vector for that attribute. As mentioned in \cite{larsen2015autoencoding}, this simple method will have problems with highly correlated visual attributes. We leave that issue for future research and use this simple method for demonstration in this paper.
	
	\item \textbf{Manipulate fake images.} Generating fake images from the generator is straight-forward. For each generated image $\xv$, we use the inference arm $\muv_{\phiv}(\xv)$ to encode it to its latent code $\zv$, add one visual attribute vector to the code to get the manipulated code $\tilde \zv$, and then regenerate with the generator $G_{\thetav}(\tilde \zv)$ to get the manipulated image $\tilde \xv$. Of course if no visual attribute vector is added, we get the reconstruction. Accordingly, we get the manipulation part of Figure \ref{fig:attribute_inference_arm}.
	
\end{enumerate}

More results on manipulating generated images are given below.

\begin{figure}[H]
\begin{center}
	\includegraphics[width=1.\columnwidth]{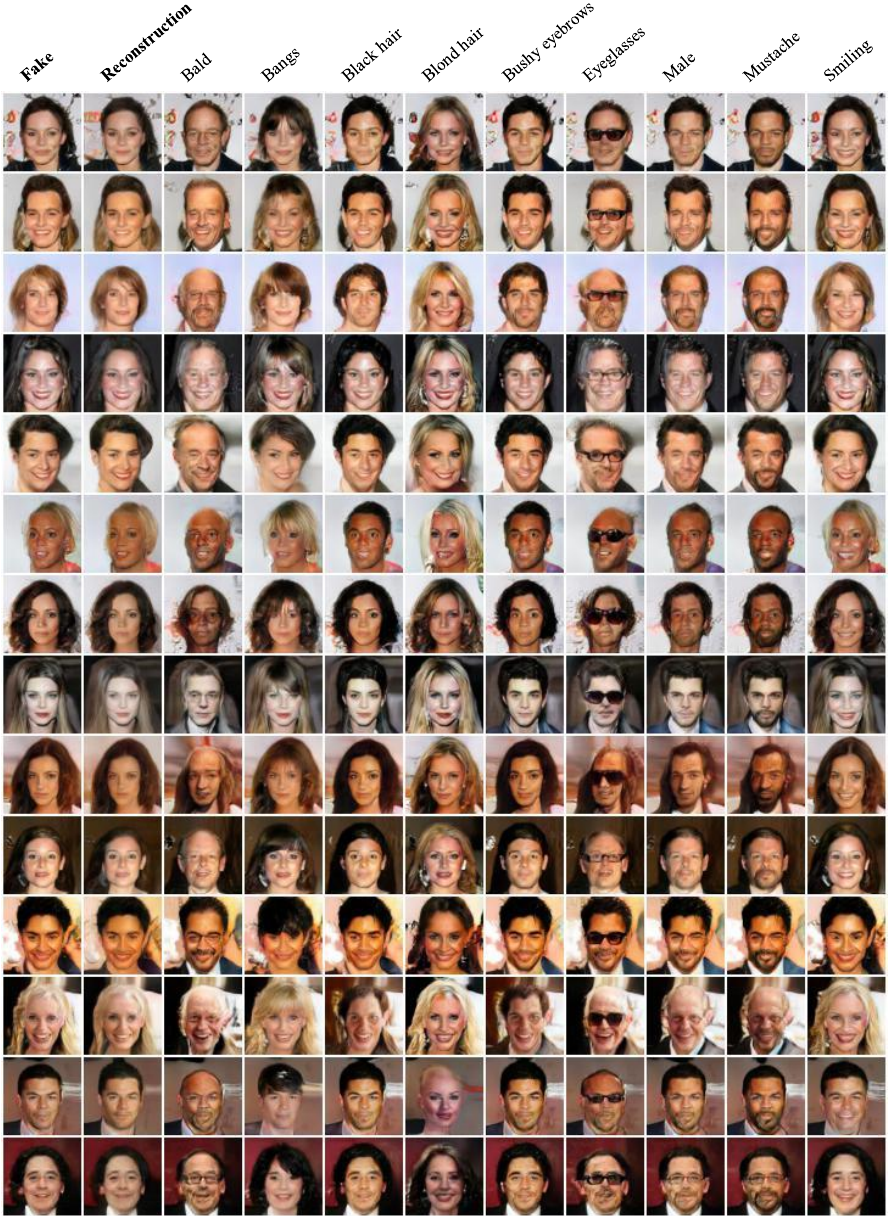}
\end{center}
\end{figure}

\begin{figure}[H]
\begin{center}
	\includegraphics[height=1.7\columnwidth]{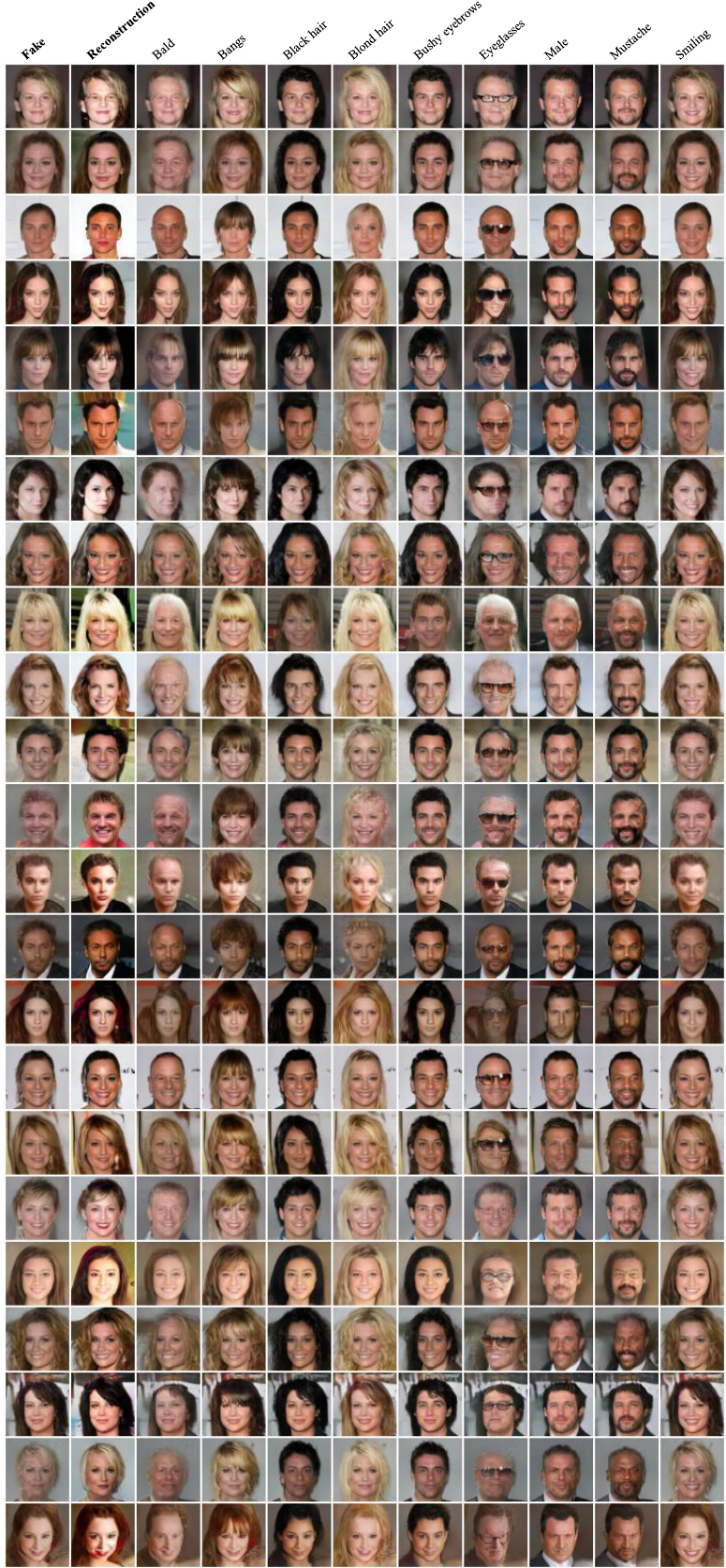}
\end{center}
\end{figure}

\section{Discussions on Other Potential Generalizations of $\alpha$-Bridge}
\label{sec:other_poten_gener}

Here, we provide additional discussions to shed light on other potential generalizations of the proposed $\alpha$-Bridge.

\subsubsection{$\alpha$-Bridge with a specific $\alpha$.}
With a proper prior knowledge on the task, one can choose a specific $\alpha$ to form a proper $\alpha$-divergence objective (which might be problem dependent \cite{hernandez2016black,li2016renyi}).By learning with an unbiased gradient $(1-\rho) \nabla_{\thetav} \Dc_{\alpha}^F + \rho \nabla_{\thetav} \Dc_{\alpha}^R$ with some $\rho$ (which exploits the gradient information from both the forward and reverse KL), one can combine the advantage of the twin gradients and enjoy a gradient estimation with low MC variance.


\subsubsection{Forward and backward transfer of $\alpha$-Bridge.}
Both ML and adversarial learning are developed for matching a model distribution to a data distribution. As mentioned in the main manuscript, $\alpha$-Bridge is proposed to unify the advantage of ML and adversarial learning and enable the smooth transfer from one to the other. Thus, there are two possible directions to transfer: transferring from ML to adversarial learning, or transferring from adversarial learning to ML. How to choose a proper transfer direction is problem dependent.

On the one hand, if one assumes that the model has enough capability and that the optimization is powerful enough to reach the optimum, it ideally does not matter which transfer direction is chosen, as long as the transfer collects all advantages from both sides to match the two distributions. On the other hand, those assumptions might be violated in practice, where we believe both transfer directions make sense for specific applications. In this paper, we focused on transferring from ML to adversarial learning because (i) the task of interest ultimately favors realistic generation (higher priority) more than keeping all data modes (lower priority); (ii) it's highly possible that ML-pretrained models are already available. As for the other transfer direction from adversarial to ML learning, it may be preferred for situations where (i) losing data modes leads to prohibitive regrets, with realistic generation a relatively lower priority and/or (ii) one may already have adversarial-pretrained models. Intuitively, such transfer is expected to preserve tight fitting/realistic generation on most data modes initialized by adversarial learning, with a few of them modified to cover the data modes missed by adversarial learning. Accordingly, one may expect a smaller mass distributed among modes compared to pure ML learning. 

\subsubsection{Is it possible to combing ML and adversarial learning via $f$-GAN?}
Motivated by our paper, one may also consider to develop some other potential ways to bridge MLE and adversarial learning, such as, 

($i$) combining MLE and an FKL f-GAN. As both the combined terms aim at minimizing FKL, ideally they have the same optimal solution that may place mass among modes (thus blurry generation); in practice however, there is a gap between them as f-GAN optimizes a lower bound of FKL and exhibits adversarial characteristics (realistic generation). 
It is shown in \cite{li2019adversarial} that although an FKL f-GAN seems to be learning based on FKL, based on the bound used there, in reality what is done is actually adversarial learning. In fact, each form of the f-GAN (for all valid f functions) actually does similar adversarial learning with different functions acting on the same likelihood ratio (see Sec. 2 of \cite{li2019adversarial} for details). So, there is a potentially large gap between FKL (MLE) and FKL f-GAN. Because of that gap, it may be tricky (if not impossible) to combine MLE and an FKL f-GAN.

($ii$) combining an FKL $f$-GAN and an RKL f-GAN. Similarly, due to the the afore mentioned gap, the mode covering of MLE is not preserved in an FKL f-GAN (adversarial learning). Accordingly, one cannot combine an FKL f-GAN and an RKL f-GAN to bridge MLE and adversarial learning.

\end{document}